\documentclass[12pt,american]{article}
\usepackage{tgtermes}
\usepackage{helvet}
\usepackage[lf]{FiraMono}

\usepackage[T1]{fontenc}
\usepackage[utf8]{inputenc}
\usepackage[a4paper]{geometry}
\usepackage{tabularx}
\usepackage{amsmath}
\usepackage{amssymb}
\usepackage{graphicx}
\usepackage[numbers]{natbib}

\makeatletter

\providecommand{\tabularnewline}{\\}

\usepackage[hidelinks]{hyperref}

\usepackage{amsthm}

\theoremstyle{definition}
\newtheorem{assumption}{Assumption}

\theoremstyle{plain}
\newtheorem{theorem}{Theorem}
\newtheorem{proposition}{Proposition}
\newtheorem{lemma}{Lemma}
\newtheorem{corollary}{Corollary}

\usepackage[nameinlink,noabbrev]{cleveref}

\crefname{assumption}{Assumption}{Assumptions}
\Crefname{assumption}{Assumption}{Assumptions}
\crefname{theorem}{Theorem}{Theorems}
\Crefname{theorem}{Theorem}{Theorems}
\crefname{proposition}{Proposition}{Propositions}
\Crefname{proposition}{Proposition}{Propositions}
\crefname{lemma}{Lemma}{Lemmas}
\Crefname{lemma}{Lemma}{Lemmas}
\crefname{corollary}{Corollary}{Corollaries}
\Crefname{corollary}{Corollary}{Corollaries}

\usepackage{pdfpages}

\ifdefined\showcaptionsetup
 \PassOptionsToPackage{caption=false}{subfig}
\fi
\usepackage{subfig}
\makeatother

\usepackage{babel}
\begin{document}
\title{Learning Bidirectional Causal Interactions with Heteroscedastic Neural
Networks}
\author{Masahiro Tanaka\thanks{Faculty of Economics, Fukuoka University, Fukuoka, Japan. Address:
8-19-1, Nanakuma, Jonan, Fukuoka, Japan 814-0180. E-mail: m.tanaka.tt@fukuoka-u.ac.jp.
This work was supported by JSPS KAKENHI Grant Number 25K21168.}}
\maketitle
\begin{abstract}
Estimating contemporaneous bidirectional interactions from observational
data is difficult because each outcome is endogenous to the other,
while flexible regressions may capture only reduced-form dependence.
This paper proposes SEM-DNN, a heteroscedastic neural simultaneous-equation
estimator that learns reciprocal structural interactions without external
instruments. Identification exploits conditional covariance diagonalization:
when structural shocks have zero conditional means, are conditionally
uncorrelated given predetermined covariates, and exhibit nonproportional
conditional variances, only the true interaction coefficients diagonalize
the conditional residual covariance across the feature space. The
method jointly approximates nonlinear structural mean functions and
feature-dependent variances using a diagonal Gaussian quasi-likelihood
that incorporates the simultaneous-system Jacobian. We establish unique
identification and positive-definite local curvature of the profiled
population criterion and show that, under neural-profile compatibility
conditions, the implemented neural criterion inherits this curvature
despite nonunique network parameterizations. The coefficients admit
a causal interpretation when the structural equations represent autonomous
mechanisms that remain invariant under the relevant interventions.
Monte Carlo experiments with nonlinear, high-dimensional nuisance
functions and non-Gaussian shocks show that SEM-DNN recovers structural
effects more reliably than parametric, kernel-based, and separate-equation
neural alternatives as information increases, although at greater
computational cost. An application to ready-to-eat cereal scanner
data illustrates how the method can study contemporaneous price--sales
feedback and assess identification strength, residual diagonalization,
variance calibration, and optimization sensitivity. The approach may
therefore be useful for bidirectional feedback systems in which credible
instruments are unavailable but informative heteroscedastic variation
is present.\\
\\
Keywords: causal learning, simultaneous equations, heteroscedasticity,
neural networks
\end{abstract}

\section{Introduction}

Many empirical and artificial systems involve contemporaneous bidirectional
interactions. For example, online platforms induce feedback between
users and advertisers, and policy variables are often adjusted in
response to the same outcomes that they affect. In such settings,
the object of interest is not merely the prediction of one outcome
from another, but rather the recovery of the directed structural interactions
governing their joint determination. When the structural equations
are interpreted as autonomous and interventionally invariant mechanisms,
these interactions have a causal interpretation.

This paper studies the problem of learning bidirectional causal effects
between two scalar outcomes from observational data. We consider the
simultaneous structural system 
\begin{align}
y_{1i} & =\gamma_{1}y_{2i}+f_{1}(x_{i})+\varepsilon_{1i},\label{eq:truesys1}\\
y_{2i} & =\gamma_{2}y_{1i}+f_{2}(x_{i})+\varepsilon_{2i},\label{eq:truesys2}
\end{align}
where $f_{1}$ and $f_{2}$ are unknown nonlinear functions of observed
features. The parameters $\gamma_{1}$ and $\gamma_{2}$ are the direct
structural interaction coefficients of primary interest. Under the
maintained structural causal interpretation, $\gamma_{1}$ measures
the direct effect from $y_{2}$ to $y_{1}$, and $\gamma_{2}$ measures
the direct effect from $y_{1}$ to $y_{2}$. The goal is to estimate
$\gamma_{1}$ and $\gamma_{2}$ while allowing flexible, high-dimensional,
and nonlinear nuisance functions. 

The main difficulty is simultaneity. A naive regression of $y_{1}$
on $y_{2}$ and $x$ treats $y_{2}$ as an exogenous covariate, even
though $y_{2}$ is itself determined partly by $y_{1}$. The same
problem arises in the reverse equation. Consequently, flexible single-equation
neural regressions may learn reduced-form predictive dependence without
recovering the structural interaction parameters.

We propose a neural simultaneous-equation estimator that uses conditional
heteroscedasticity as the source of identification. The key idea is
that the correct structural coefficients transform the observed outcomes
into residuals whose conditional covariance matrix is diagonal. At
an incorrect value of $(\gamma_{1},\gamma_{2})$, the transformed
residuals generally mix the two structural shocks, producing conditional
residual correlation. If the two structural shock variances vary nonproportionally
over the feature space, an incorrect transformation cannot diagonalize
the conditional covariance matrix for all covariate values. Thus,
conditional second moments provide information about the two directions
of contemporaneous causality.

To operationalize this idea, we model both the structural mean functions
and the conditional variance functions with neural networks. The conditional
variances are specified as 
\[
\operatorname{Var}(\varepsilon_{ki}\mid x_{i})=g_{k}(x_{i}),\qquad k=1,2,
\]
with
\[
g_{k}(x_{i})=g_{\min}+\psi\{\widetilde{g}_{k}(x_{i})\},\qquad k=1,2.
\]
Here, $\psi(\cdot)$ is the softplus link. The variance networks are
not used merely as uncertainty estimators; rather, they allow the
model to learn the heteroscedastic variation used for identification.
The model is trained by maximizing a diagonal Gaussian quasi-likelihood
for the simultaneous system, including the Jacobian term associated
with the transformation between observed outcomes and structural shocks.
The resulting estimator, which we call the simultaneous-equation deep
neural network estimator (SEM-DNN), jointly learns nonlinear mean
functions, feature-dependent variances, and bidirectional structural
effects.

This paper makes three contributions. First, it develops a joint neural
simultaneous-equation learning framework in which nonlinear representations
of the structural mean functions and feature-dependent conditional
variances are optimized together with the reciprocal interaction coefficients
under a Jacobian-adjusted diagonal Gaussian quasi-likelihood. Unlike
the fixed-feature kernel implementation considered here, the neural
representations of the nuisance functions adapt during structural
estimation. This additional flexibility is accompanied by variance-aware
training stabilization, separate regularization of the mean and variance
networks, and model selection based on the original, unweighted quasi-likelihood.

Second, building on established heteroscedastic identification arguments,
the paper characterizes the unrestricted profiled population criterion
and establishes conditions under which a locally profiled neural population
criterion inherits its curvature despite nonunique network parameterizations.
Identification originates from nonproportional conditional shock variances
rather than neural-network architecture or depth.

Third, through Monte Carlo experiments, parameter-matched and hybrid-network
comparisons, alternative DGPs, and supplementary comparisons with
cyclic causal-learning methods, the paper examines the relationship
between representation flexibility, structural recovery, and estimation
reliability. These experiments document both the potential benefits
and the finite-sample costs of joint neural estimation, rather than
suggesting uniform superiority over more specialized alternatives.

This paper also relates to the broader literature on causal machine
learning \citep{Feuerriegel2024,Kaddour2025}. Neural instrumental-variable
methods \citep{Hartford2017}, such as Deep IV, use flexible function
approximators to estimate causal response functions in the presence
of endogeneity, but require valid instruments satisfying relevance
and exclusion restrictions. SEM-DNN addresses a different identification
setting: it does not require external instruments; instead, it exploits
nonproportional variation in the conditional variances of the structural
shocks. Neural networks approximate the unknown conditional mean and
variance functions; they do not themselves provide causal identification.

The proposed approach is also related to causal discovery in structural
equation models. Classical methods such as LiNGAM \citep{Shimizu2006}
identify linear acyclic structures through non-Gaussian disturbances,
while extensions consider cyclic systems \citep{Lacerda2008} and
observational or interventional equilibrium data \citep{Bongers2021}.
More recent approaches differ in their identifying restrictions. Zhou
et al. \citep{Zhou2022} exploit covariate-dependent causal effects,
whereas SEM-DNN maintains constant interaction coefficients and exploits
heterogeneity in structural shock variances. Dai et al. \citep{Dai2024}
use non-Gaussian independence to learn linear cyclic models. Other
neural approaches, including NODAGS-Flow \citep{Sethuraman2023} and
Bicycle \citep{Rohbeck2024}, incorporate interventional information.
These distinctions matter because SEM-DNN assumes a prespecified reciprocal
two-node structure rather than learning an unrestricted causal graph.
Its identification requires structural shocks that are conditionally
mean-zero and conditionally uncorrelated, together with nonproportional
conditional variances, but requires neither non-Gaussianity nor interventional
observations.

The paper is also related to the econometric literature on simultaneous
equations and identification through heteroscedasticity. Rigobon \citep{Rigobon2003}
introduced identification using regime-dependent variance shifts,
and Milunovich and Yang \citep{Milunovich2018} proposed a quasi-maximum-likelihood
estimator for heteroscedastic simultaneous-equation systems. BACKSHIFT
\citep{Rothenhaeusler2015} similarly exploits second-moment variation
to identify cyclic linear models, but requires covariance changes
across multiple shift-intervention environments. Tanaka \citep{Tanaka2025}
extended heteroscedastic simultaneous-equation estimation to online
estimation and large-scale kernel learning. SEM-DNN does not introduce
a new source of statistical identification. Its methodological contribution
is to integrate adaptive neural representation learning into the heteroscedastic
simultaneous-equation framework, allowing the structural mean functions
and conditional variance functions to be learned jointly with the
reciprocal interaction coefficients. Unlike the parametric and fixed-random-feature
kernel benchmarks considered here, the proposed estimator optimizes
its nonlinear nuisance representations within the structural learning
objective. This flexibility is complemented by a neural-profile curvature
analysis, stabilized joint training, and held-out identification and
model-fit diagnostics. It broadens the available modeling approach
without guaranteeing superior finite-sample structural recovery. The
method also connects to recent work on heteroscedastic neural regression
\citep{Detlefsen2019,Seitzer2022,Stirn2023,WongToi2024}, particularly
the instability and calibration problems associated with joint mean--variance
learning.

These distinctions motivate the supplementary head-to-head comparisons
with Local ISA-LiNG \citep{Dai2024} and BACKSHIFT \citep{Rothenhaeusler2015}
in Appendix~D.13. We construct separate, deliberately simplified
DGPs satisfying the identifying assumptions of SEM-DNN and each competitor.
These designs feature linear feedback and identically zero structural
mean functions, eliminating the need for flexible mean modeling under
the true DGP while SEM-DNN continues to estimate all four nuisance
networks. The experiments therefore assess finite-sample structural
recovery under overlapping identifying assumptions rather than the
methods' full modeling capabilities. The specialized competitors exhibit
accuracy and computational advantages in these settings. The contribution
of SEM-DNN lies instead in its integrated learning formulation, which
allows nonlinear structural mean functions and feature-dependent variances
to be represented adaptively within the simultaneous-equation estimation
objective. The nonlinear simulations and architectural comparisons
elsewhere in the paper examine the benefits and limitations of this
additional modeling flexibility.

To illustrate the empirical use of the proposed framework, we apply
SEM-DNN to the ready-to-eat cereal category in the Dominick’s scanner
database. The application examines contemporaneous feedback between
product movement and package price using historical sales and pricing
information, product attributes, and store characteristics as predetermined
covariates. Stores are separated into training, validation, and diagnostic
partitions so that model selection is conducted independently of the
final diagnostic assessment. The fitted conditional variances exhibit
nonproportional variation, providing heteroscedastic information relevant
to separating the two interaction directions. At the same time, dispersion
across neural-network training seeds and the held-out residual-diagonalization
and variance-calibration diagnostics reveal material empirical uncertainty.
The application is therefore intended to demonstrate the complete
estimation and diagnostic workflow of SEM-DNN, rather than to provide
definitive causal elasticity estimates for cereal demand and pricing.

More broadly, SEM-DNN may be useful in observational systems in which
two outcomes adjust to one another within the same measurement window,
the feature-dependent mean and variance relationships are nonlinear
or high dimensional, and credible external instruments are unavailable.
In contrast to separate flexible regressions, which may reproduce
reduced-form predictive associations, the proposed approach uses conditional
second-moment variation to distinguish the two structural interaction
directions while allowing neural networks to approximate complex nuisance
functions. Its practical value lies not only in producing interaction
estimates but also in providing diagnostics for the strength of the
heteroscedastic identifying variation, remaining conditional residual
dependence, variance calibration, gradient concentration, and sensitivity
to nonconvex optimization. Potential applications include price--quantity
adjustment, interactions between users and online platforms, and coupled
biological, neural, or engineering processes, provided that the covariates
are predetermined and the maintained structural invariance and conditional
shock-orthogonality assumptions are substantively credible. The method
should thus be viewed as a complement to instrumental-variable and
acyclic causal-learning approaches for feedback systems in which identification
through heterogeneous conditional variances is empirically plausible.

The paper proceeds as follows. Section~\ref{sec:Method} introduces
the neural simultaneous-equation model and the quasi-likelihood objective
and discusses identification of bidirectional interactions. Section~\ref{sec:learning-algorithm}
explains the learning algorithm. Section~\ref{sec:simulation-study}
presents Monte Carlo evidence. Section~\ref{sec:empirical-illustration}
presents an empirical illustration, and Section~\ref{sec:conclusion}
concludes. 

\section{Method}

\label{sec:Method}

\subsection{Structural model and causal interpretation}

\label{subsec:causal-learning-model}

We observe $\{y_{1i},y_{2i},z_{i}\}_{i=1}^{n}$, where $y_{1i}$ and
$y_{2i}$ are the two outcomes jointly determined within the focal
outcome window, and $z_{i}$ denotes the raw information available
before that window, as well as time-invariant attributes determined
independently of the within-window outcomes. Our primary statistical
target is the pair of directed structural interactions between $y_{1}$
and $y_{2}$. Under Assumption~\ref{ass:structural-causal-interpretation},
these parameters also represent direct causal interactions.

Let $T_{\mathrm{pre}}$ denote a preprocessing map fitted exclusively
on the training sample, and define the model input by $x_{i}=T_{\mathrm{pre}}(z_{i})$.
We refer to $x_{i}$ as the vector of predetermined covariates. Formally,
each component of $z_{i}$ must satisfy at least one of the following
timing conditions: (i) it is realized strictly before the focal outcome
window; (ii) it is time invariant over the sample and is not constructed
from focal-window outcomes or decisions; or (iii) it is a deterministic
transformation of variables satisfying (i) or (ii), where all data-dependent
transformation parameters are estimated from the training sample only. 

Variables measured or constructed using $y_{1i}$, $y_{2i}$, or other
variables jointly determined within the focal outcome window are excluded
from $z_{i}$. In particular, a variable is not treated as predetermined
merely because it is observed in the data; its construction must not
use contemporaneous outcomes, contemporaneous residuals, or post-outcome
information.

Let $\varepsilon_{i}=(\varepsilon_{1i},\varepsilon_{2i})^{\top}$.
The maintained structural restrictions are 
\[
\operatorname{E}(\varepsilon_{i}\mid x_{i})=0
\]
 and
\[
\operatorname{Var}(\varepsilon_{i}\mid x_{i})=\left(\begin{array}{cc}
g_{10}(x_{i}) & 0\\
0 & g_{20}(x_{i})
\end{array}\right).
\]
Thus, predeterminedness is primarily a timing and construction restriction.
It does not, by itself, establish conditional exogeneity or conditional
orthogonality; these are additional identifying assumptions imposed
on the structural shocks.

The same input vector $x_{i}$ is supplied to the two structural-mean
networks and the two conditional-variance networks. The four networks
may learn different transformations or effectively select different
subsets of $x_{i}$, but they receive the same preprocessed information
set. No outcome-dependent feature selection or preprocessing is performed
separately for the mean and variance networks.

Our goal is to learn directed causal interactions between $y_{1}$
and $y_{2}$ from observational data. We represent the data-generating
mechanism by the two-equation structural model \citep{Milunovich2018}
\begin{align}
y_{1i} & =\gamma_{1}y_{2i}+f_{1}(x_{i};\nu_{1})+\varepsilon_{1i},\label{eq:sys1}\\
y_{2i} & =\gamma_{2}y_{1i}+f_{2}(x_{i};\nu_{2})+\varepsilon_{2i}.\label{eq:sys2}
\end{align}
The parameters $\gamma=(\gamma_{1},\gamma_{2})^{\top}$ are the primary
parameters of interest. They encode the strength and direction of
the within-window structural interaction: $\gamma_{1}$ measures the
effect from $y_{2}$ to $y_{1}$, and $\gamma_{2}$ measures the effect
from $y_{1}$ to $y_{2}$. The functions $f_{1}$ and $f_{2}$ absorb
nonlinear effects of the observed features on the two outcomes. Thus,
the structural parameters $\gamma_{1}$ and $\gamma_{2}$ are estimated
after adjusting for flexible feature-dependent baseline responses.

The model allows bidirectional feedback, and therefore differs from
standard acyclic causal graph learning. The system is instead interpreted
as a two-node structural causal model with simultaneous directed interactions.
Let 
\[
\Gamma(\gamma)=\begin{pmatrix}1 & -\gamma_{1}\\
-\gamma_{2} & 1
\end{pmatrix},\qquad d(\gamma)=\det\Gamma(\gamma)=1-\gamma_{1}\gamma_{2}.
\]
To obtain a well-defined mapping between structural shocks and observed
outcomes, we require the interaction matrix $\Gamma(\gamma)$ to be
nonsingular. In the implementation, we use the bounded reparameterization
$\gamma_{k}=\bar{\gamma}\tanh(\widetilde{\gamma}_{k})$, $k=1,2$,
with a fixed $\bar{\gamma}<1$. This parameterization stabilizes training
and keeps the system uniformly nonsingular.

A key source of information in the model is feature-dependent noise
dispersion. We assume that the structural shocks satisfy 
\[
\operatorname{E}(\varepsilon_{ki}\mid x_{i})=0,\qquad k=1,2,
\]
and model their conditional variances as 
\begin{align}
\operatorname{Var}(\varepsilon_{ki}\mid x_{i}) & =g_{k}(x_{i};\alpha_{k})\nonumber \\
 & =g_{\min}+\psi\{\widetilde{g}_{k}(x_{i};\alpha_{k})\},\label{eq:condval}
\end{align}
for $k=1,2$, where $g_{\min}>0$ is a small fixed lower bound and
\[
\psi(t)=\log\{1+\exp(t)\}
\]
is the softplus link. The lower bound and positive-valued link function
are important in practice because the quasi-log-likelihood introduced
below contains both $\log g_{k}(x_{i};\alpha_{k})$ and $1/g_{k}(x_{i};\alpha_{k})$.
Extremely small fitted variances can therefore produce unstable gradients
and overfitting. The softplus transformation keeps the variance positive
while being less prone to numerical explosion than an exponential
link.

We parameterize the unknown functions using feedforward neural networks.
For each $k=1,2$, the mean network $f_{k}(\cdot;\nu_{k})$ maps $x_{i}$
to a scalar output. It consists of fully connected layers with nonlinear
activation functions, such as the Gaussian error linear unit (GELU),
$\tanh$, sigmoid, and softplus, followed by a linear output layer.
The parameter vector $\nu_{k}$ collects all weights and biases in
this mean network. This specification allows the feature effects in
the structural mean equations to be highly nonlinear while keeping
$\gamma$ as a low-dimensional and interpretable causal target.

Similarly, $\widetilde{g}_{k}(\cdot;\alpha_{k})$ is a scalar-valued
neural network for the unconstrained variance index. The variance
network captures how the magnitude of the unobserved structural shock
changes across the feature space. This conditional heteroscedasticity
is not treated as a nuisance artifact; rather, it provides identifying
information for separating the two directed interactions in the simultaneous
system.

The four neural networks, $f_{1},f_{2},\widetilde{g}_{1},\widetilde{g}_{2}$,
are trained jointly with the structural interaction parameters $\gamma_{1}$
and $\gamma_{2}$. Joint training is necessary because the residuals
\begin{align}
e_{1i} & =y_{1i}-\gamma_{1}y_{2i}-f_{1}(x_{i};\nu_{1}),\label{eq:resid1}\\
e_{2i} & =y_{2i}-\gamma_{2}y_{1i}-f_{2}(x_{i};\nu_{2}),\label{eq:resid2}
\end{align}
depend on both the structural interaction parameters and the neural
mean functions. The variance networks then weight these residuals
through the diagonal Gaussian quasi-likelihood. The resulting estimator
therefore learns nonlinear feature effects, conditional noise levels,
and bidirectional causal interactions in a single optimization problem.
Section~\ref{sec:learning-algorithm} describes the training, regularization,
validation, and checkpoint-selection procedure.

\subsection{Population risk, profiling, and assumptions}

\label{subsec:populationrisketc}

This section defines the population criterion used to learn the bidirectional
interaction parameter $\gamma$ and separates three objects that play
different roles in the identification analysis. The first is the diagonal
Gaussian quasi-likelihood used for training. The second is an unrestricted
function-level profile of this criterion, which isolates the population
information about $\gamma$ after the conditional mean and diagonal
variance functions have been profiled out. The third is the neural-network
parameterization, which is used to approximate these profiled functions
in the implemented estimator.

Let $\gamma_{0}=(\gamma_{10},\gamma_{20})^{\top}$ denote the true
interaction vector. The structural model can be written compactly
as 
\[
\Gamma(\gamma_{0})y_{i}=f_{0}(x_{i})+\varepsilon_{i},
\]
where $y_{i}=(y_{1i},y_{2i})^{\top}$, $\varepsilon_{i}=(\varepsilon_{1i},\varepsilon_{2i})^{\top}$,
and 
\[
f_{0}(x_{i})=\{f_{10}(x_{i}),f_{20}(x_{i})\}^{\top}.
\]
For a candidate parameter value, let 
\[
\eta=(\alpha_{1}^{\top},\alpha_{2}^{\top},\nu_{1}^{\top},\nu_{2}^{\top})^{\top}
\]
collect the neural-network nuisance parameters. The population diagonal
Gaussian quasi-log-likelihood is 
\[
Q(\gamma,\eta)=\operatorname{E}\{\ell_{i}(\gamma,\eta)\},
\]
where 
\begin{align*}
\ell_{i}(\gamma,\eta)= & \log|d(\gamma)|-\log(2\pi)\\
 & -\frac{1}{2}\sum_{k=1}^{2}\left[\log g_{k}(x_{i};\alpha_{k})+\frac{e_{ki}(\gamma,\nu_{k})^{2}}{g_{k}(x_{i};\alpha_{k})}\right].
\end{align*}
The term $\log|d(\gamma)|$ is the Jacobian contribution from the
transformation between observed outcomes and structural shocks. The
word ``quasi'' emphasizes that the diagonal Gaussian likelihood
is used as a training and identification criterion; the structural
shocks need not be Gaussian.

Define the full parameter vector 
\[
\theta=(\gamma^{\top},\alpha_{1}^{\top},\alpha_{2}^{\top},\nu_{1}^{\top},\nu_{2}^{\top})^{\top}.
\]
The sample negative quasi-log-likelihood is 
\[
\mathcal{L}_{n}(\theta)=-\frac{1}{n}\sum_{i=1}^{n}\ell_{i}(\theta).
\]
In the implementation, the estimator minimizes a regularized mini-batch
analogue of $\mathcal{L}_{n}(\theta)$. The regularization terms are
used for finite-sample stabilization and training; they are not part
of the population identifying criterion.

To clarify the source of information about $\gamma$, it is useful
to introduce a function-level profiled version of the criterion. For
any candidate $\gamma$, define the conditional mean of the transformed
outcomes by 
\[
m_{\gamma,k}^{*}(x)=\operatorname{E}\{(\Gamma(\gamma)y_{i})_{k}\mid x_{i}=x\},\qquad k=1,2,
\]
and define the corresponding mean-profiled residual vector by 
\[
r_{\gamma i}=\Gamma(\gamma)y_{i}-m_{\gamma}^{*}(x_{i}),
\]
where 
\[
m_{\gamma}^{*}(x)=\{m_{\gamma,1}^{*}(x),m_{\gamma,2}^{*}(x)\}^{\top}.
\]
Let 
\[
v_{\gamma,k}^{*}(x)=\operatorname{Var}(r_{\gamma,ki}\mid x_{i}=x),\qquad k=1,2.
\]
If the conditional mean functions and diagonal conditional variance
functions are profiled out without neural-network restrictions, the
resulting population criterion is, up to an additive constant independent
of $\gamma$, 
\[
Q^{*}(\gamma)=\log|d(\gamma)|-\frac{1}{2}\operatorname{E}\left[\log v_{\gamma,1}^{*}(x_{i})+\log v_{\gamma,2}^{*}(x_{i})\right].
\]
This unrestricted profiled criterion is not the implemented estimator.
Rather, it is a population benchmark that isolates the identifying
signal carried by the diagonalization of the conditional covariance
matrix of $r_{\gamma i}$. Section~\ref{subsec:identification} shows
that nonproportional variation in the two structural conditional variances
generates strict local concavity of $Q^{*}(\gamma)$ at $\gamma_{0}$.

The neural-network parameterization introduces an additional technical
issue: the nuisance parameter vector $\eta$ is generally not unique.
Different neural-network weights can represent the same conditional
mean and variance functions. Since the structural target is $\gamma$,
this nonuniqueness should not be interpreted as a failure to identify
the directed interaction parameters. We therefore treat nuisance parameter
values that induce the same functions as equivalent. Define the equivalence
relation 
\begin{align*}
\eta\sim\eta'\Longleftrightarrow & f_{k}(\cdot;\nu_{k})=f_{k}(\cdot;\nu_{k}')\ \text{and}\\
 & g_{k}(\cdot;\alpha_{k})=g_{k}(\cdot;\alpha_{k}'),\quad k=1,2
\end{align*}
almost surely on the support of $x_{i}$. Let $\mathcal{E}_{0}$ denote
the equivalence class of nuisance parameters that induce the true
functions $f_{10},f_{20},g_{10},g_{20}$. For any $\eta_{0}\in\mathcal{E}_{0}$,
define 
\[
Q_{0}=Q(\gamma_{0},\eta_{0}),
\]
which is common across all representatives of the equivalence class.

At a regular representative $\eta_{0}$, let $\mathcal{K}_{\eta_{0}}$
denote the tangent space to $\mathcal{E}_{0}$. Directions in $\mathcal{K}_{\eta_{0}}$
correspond to infinitesimal changes in network weights that leave
the represented functions unchanged. Let $\mathcal{T}_{\eta_{0}}=\mathcal{K}_{\eta_{0}}^{\perp}$
be a local transversal slice. Local neural profiling is conducted
on this slice, or equivalently on the quotient space obtained by removing
function-preserving neural reparameterizations. For a small $\delta>0$,
define the local neural profiled population criterion 
\[
Q_{\delta,\eta_{0}}(\gamma)=\sup_{a\in\mathcal{T}_{\eta_{0}},\,\|a\|<\delta}Q(\gamma,\eta_{0}+a).
\]
This criterion profiles only over locally identifiable nuisance directions.
The role of the neural identification argument in Section~\ref{subsec:identification}
is to connect $Q_{\delta,\eta_{0}}(\gamma)$ to the unrestricted profiled
benchmark $Q^{*}(\gamma)$.

\begin{assumption}[Structural causal interpretation]

\label{ass:structural-causal-interpretation}

Equations (\ref{eq:sys1})--(\ref{eq:sys2}) represent autonomous
structural mechanisms. The mechanisms and structural shocks remain
invariant under the interventions considered, the predetermined covariates
$x_{i}$ are not affected by interventions within the focal outcome
window, and the intervened system has a well-defined solution.

\end{assumption}

Under this assumption, $\gamma_{1}$ and $\gamma_{2}$ have the interpretation
of direct contemporaneous causal effects. Assumption~\ref{ass:structural-causal-interpretation}
supplies the causal interpretation of the structural coefficients.
It is conceptually distinct from Assumptions~\ref{ass:structural-primitives}
and \ref{ass:neural-profile-compatibility}, which establish identification
from the observational distribution and compatibility with the neural-network
parameterization, respectively.

The assumptions are divided into structural primitives and neural-profile
conditions to separate the source of identification from the technical
conditions needed to transfer the function-level argument to the neural-network
parameterization.

\begin{assumption}[Structural primitives]

\label{ass:structural-primitives}

The following conditions hold at the population level.
\begin{enumerate}
\item[(S1)] The true interaction parameter $\gamma_{0}$ belongs to a stable
region. There exist a neighborhood $\mathcal{G}$ of $\gamma_{0}$
and a constant $c>0$ such that 
\[
\inf_{\gamma\in\mathcal{G}}|1-\gamma_{1}\gamma_{2}|\geq c.
\]
\item[(S2)] The structural shocks satisfy 
\[
\operatorname{E}(\varepsilon_{i}\mid x_{i})=0,
\]
and 
\[
\operatorname{Var}(\varepsilon_{i}\mid x_{i})=\begin{pmatrix}g_{10}(x_{i}) & 0\\
0 & g_{20}(x_{i})
\end{pmatrix}.
\]
\item[(S3)] The conditional variance functions satisfy 
\[
0<g_{10}(x_{i})<\infty,\qquad0<g_{20}(x_{i})<\infty
\]
almost surely, with
\[
\operatorname{E}\{|\log g_{10}(x_{i})|\}<\infty,\qquad\operatorname{E}\{|\log g_{20}(x_{i})|\}<\infty.
\]
Let
\[
\rho_{0}(x_{i})=\frac{g_{10}(x_{i})}{g_{20}(x_{i})}.
\]
There exist constants $\ensuremath{0<c_{\rho}<C_{\rho}<\infty}$ such
that $c_{\rho}\leq\rho_{0}(x_{i})\leq C_{\rho}$ almost surely, and
$\rho_{0}(x_{i})$ is not almost surely constant. 
\end{enumerate}
\end{assumption}

Assumption~\ref{ass:structural-primitives} contains the primitive
restrictions for observational identification of the structural interaction
coefficients. The diagonal conditional covariance restriction states
that, after conditioning on the predetermined covariates, the two
structural shocks are conditionally uncorrelated. The nonconstant
variance-ratio condition is the source of heteroscedastic identifying
variation. If $\rho_{0}(x_{i})$ were constant almost surely, the
conditional covariance restriction would not generally contain enough
information to separate the two directed interaction coefficients.

The bounded-ratio condition is stronger than necessary. The local
curvature result continues to hold under the weaker moment condition
\[
\operatorname{E}\{\rho_{0}(x_{i})\}<\infty,\qquad\operatorname{E}\{\rho_{0}(x_{i})^{-1}\}<\infty,
\]
together with a local uniform-integrability condition sufficient to
interchange expectation and the second-order expansion.

\begin{assumption}[Neural profile compatibility]

\label{ass:neural-profile-compatibility}

Let $\eta_{0}\in\mathcal{E}_{0}$ be a regular representative.
\begin{enumerate}
\item[(N1)] The neural-network classes correctly represent the true structural
mean and conditional variance functions at $\eta_{0}$. The population
criterion $Q(\gamma,\eta)$ is twice continuously differentiable in
a neighborhood of $(\gamma_{0},\eta_{0})$, after restricting local
nuisance perturbations to the slice $\mathcal{T}_{\eta_{0}}$.
\item[(N2)] For all $h$ in a neighborhood of zero, with $\gamma=\gamma_{0}+h$,
there exists a twice continuously differentiable path $a^{*}(h)\in\mathcal{T}_{\eta_{0}}$,
$a^{*}(0)=0$, such that the neural mean and variance functions induced
by $\eta_{0}+a^{*}(h)$ represent the unrestricted profiled functions
$m_{\gamma,k}^{*}(x)$ and $v_{\gamma,k}^{*}(x)$, $k=1,2$, up to
an error that is $o(\|h\|^{2})$ in the population quasi-likelihood.
Equivalently, 
\[
Q_{\delta,\eta_{0}}(\gamma_{0}+h)=Q^{*}(\gamma_{0}+h)+o(\|h\|^{2}),\qquad h\to0.
\]
\item[(N3)] The equivalence class $\mathcal{E}_{0}$ is locally a $C^{2}$ equivalence
manifold at a regular representative $\eta_{0}$. Let 
\[
J=-\nabla^{2}Q(\gamma_{0},\eta_{0})=\begin{pmatrix}J_{\gamma\gamma} & J_{\gamma\eta}\\
J_{\eta\gamma} & J_{\eta\eta}
\end{pmatrix}.
\]
The nuisance block $J_{\eta\eta}$ is positive semidefinite on the
full nuisance parameter space, with 
\[
\ker(J_{\eta\eta})=\mathcal{K}_{\eta_{0}}.
\]
Equivalently, the induced quadratic form is positive definite on the
quotient slice $\mathcal{T}_{\eta_{0}}=\mathcal{K}_{\eta_{0}}^{\perp}$.
Moreover, 
\[
\ker(J_{\eta\eta})\subseteq\ker(J_{\gamma\eta}).
\]
\end{enumerate}
\end{assumption}

Assumption~\ref{ass:neural-profile-compatibility} is not a primitive
source of causal identification. Rather, it states that the neural
parameterization can locally reproduce the unrestricted profiled mean
and diagonal variance functions, after quotienting out neural-network
reparameterizations that do not change the induced functions. Thus,
the primitive identifying variation comes from Assumption~\ref{ass:structural-primitives},
especially the nonconstant variance ratio $\rho_{0}(x_{i})$, while
Assumption~\ref{ass:neural-profile-compatibility} allows the neural
profiled criterion to inherit the local curvature of the unrestricted
profiled criterion. Section~\ref{subsec:identification} first studies
the unrestricted profiled criterion and then shows how its local curvature
transfers to the neural profiled criterion under Assumptions~\ref{ass:structural-primitives}~and~\ref{ass:neural-profile-compatibility}.

\subsection{Identification of bidirectional interactions}

\label{subsec:identification}

This section explains why the two directed interaction parameters
$\gamma_{1}$ and $\gamma_{2}$ can be learned from observational
data in the proposed heteroscedastic neural simultaneous-equation
model. Identification comes from feature-dependent movement of the
two structural shock variances rather than from exclusion restrictions
or instrumental variables. With nonproportional conditional variances,
only the correct interaction parameter diagonalizes the conditional
covariance matrix of the structural residuals for all covariate values.

The analysis proceeds in two steps. First, we study the unrestricted
profiled population criterion $Q^{*}(\gamma)$ defined in Section~\ref{subsec:populationrisketc}.
This function-level criterion profiles out the conditional mean functions
and the diagonal conditional variance functions without imposing neural-network
restrictions. It therefore isolates the primitive identifying signal.
Second, we show that, under the neural-profile compatibility conditions
in Assumption~\ref{ass:neural-profile-compatibility}, the local
neural profiled criterion $Q_{\delta,\eta_{0}}(\gamma)$ inherits
the same curvature. This separates the statistical source of identification
from the technical nonuniqueness of neural-network weights.

Let 
\[
\Gamma_{0}=\Gamma(\gamma_{0}),\qquad d_{0}=\det\Gamma_{0}=1-\gamma_{10}\gamma_{20}.
\]
For a candidate value $\gamma$, the mean-profiled residual vector
is 
\[
r_{\gamma i}=\Gamma(\gamma)y_{i}-\operatorname{E}\{\Gamma(\gamma)y_{i}\mid x_{i}\}.
\]
Since 
\[
\Gamma(\gamma_{0})y_{i}=f_{0}(x_{i})+\varepsilon_{i}
\]
and $\operatorname{E}(\varepsilon_{i}\mid x_{i})=0$, the profiled
residual can be written as 
\[
r_{\gamma i}=\Gamma(\gamma)\Gamma_{0}^{-1}\varepsilon_{i}.
\]
Moreover, 
\[
\Gamma(\gamma)\Gamma_{0}^{-1}=\frac{1}{d_{0}}\begin{pmatrix}1-\gamma_{1}\gamma_{20} & \gamma_{10}-\gamma_{1}\\
\gamma_{20}-\gamma_{2} & 1-\gamma_{2}\gamma_{10}
\end{pmatrix}.
\]
Thus, for a candidate value $\gamma$, the conditional covariance
between the two residual components is 
\begin{align*}
\operatorname{Cov}(r_{\gamma,1i},r_{\gamma,2i}\mid x_{i})= & \frac{1}{d_{0}^{2}}[(1-\gamma_{1}\gamma_{20})(\gamma_{20}-\gamma_{2})g_{10}(x_{i})\\
 & +(\gamma_{10}-\gamma_{1})(1-\gamma_{2}\gamma_{10})g_{20}(x_{i})].
\end{align*}
At $\gamma=\gamma_{0}$, this covariance is zero almost surely. At
an incorrect value of $\gamma$, the transformation generally mixes
the two structural shocks. The following proposition shows that this
mixing creates local curvature in the unrestricted profiled quasi-likelihood.

\begin{proposition}[Profiled curvature from heteroscedastic diagonalization]

\label{prop:profiled-curvature}

Suppose Assumption~\ref{ass:structural-primitives} holds. Let 
\[
\rho_{0}(x_{i})=\frac{g_{10}(x_{i})}{g_{20}(x_{i})}.
\]
For $h=(h_{1},h_{2})^{\top}$, set $\gamma=\gamma_{0}+h$. Then the
unrestricted profiled population criterion satisfies 
\[
Q^{*}(\gamma_{0}+h)-Q^{*}(\gamma_{0})=
\]
\[
-\frac{1}{2d_{0}^{2}}\operatorname{E}\left[h_{2}^{2}\rho_{0}(x_{i})+2h_{1}h_{2}+h_{1}^{2}\rho_{0}(x_{i})^{-1}\right]+o(\|h\|^{2}).
\]
Equivalently, 
\[
Q^{*}(\gamma_{0}+h)-Q^{*}(\gamma_{0})=-\frac{1}{2}h^{\top}I_{\gamma}^{*}h+o(\|h\|^{2}),
\]
where 
\begin{equation}
I_{\gamma}^{*}=\frac{1}{d_{0}^{2}}\begin{pmatrix}\operatorname{E}\{\rho_{0}(x_{i})^{-1}\} & 1\\
1 & \operatorname{E}\{\rho_{0}(x_{i})\}
\end{pmatrix}.\label{eq:curvature-matrix}
\end{equation}
If $\rho_{0}(x_{i})$ is not almost surely constant, then $I_{\gamma}^{*}\succ0$.

\end{proposition}

The proof is given in Appendix~A.1.

Proposition~\ref{prop:profiled-curvature} gives the local risk geometry
behind the identification argument. The curvature is generated by
the off-diagonal conditional covariance created when an incorrect
$\gamma$ mixes the two structural shocks. To first order, 
\[
\operatorname{Cov}(r_{\gamma,1i},r_{\gamma,2i}\mid x_{i})=
\]
\[
-\frac{1}{d_{0}}\{h_{2}g_{10}(x_{i})+h_{1}g_{20}(x_{i})\}+O(\|h\|^{2}).
\]
The diagonal Gaussian profiled criterion penalizes this conditional
correlation through 
\[
\log\left\{ \frac{v_{\gamma,1}^{*}(x_{i})v_{\gamma,2}^{*}(x_{i})}{\det\operatorname{Var}(r_{\gamma i}\mid x_{i})}\right\} .
\]
Consequently, the local loss in the profiled criterion is proportional
to 
\[
\operatorname{E}\left[\frac{\{h_{2}g_{10}(x_{i})+h_{1}g_{20}(x_{i})\}^{2}}{g_{10}(x_{i})g_{20}(x_{i})}\right].
\]
This expression equals zero only if $h=0$, provided that the variance
ratio $\rho_{0}(x_{i})$ is not almost surely constant. 

The next lemma gives the corresponding global diagonalization statement
within the stable parameter region.

\begin{lemma}[Unique conditional diagonalization]

\label{lem:unique-cond-diag}

Assume that, conditional on $x_{i}$, the structural shocks satisfy
\[
\operatorname{E}(\varepsilon_{i}\mid x_{i})=0,\qquad\operatorname{Var}(\varepsilon_{i}\mid x_{i})=\operatorname{diag}\{g_{10}(x_{i}),g_{20}(x_{i})\}.
\]
Assume also that, for every $\gamma$ in the stability region, the
profiled nuisance space contains 
\[
\operatorname{E}\{\Gamma(\gamma)y_{i}\mid x_{i}\}
\]
and the diagonal conditional second moments 
\[
\operatorname{Var}(r_{\gamma,ki}\mid x_{i}),\qquad k=1,2.
\]
Suppose that the candidate parameter space satisfies the compact stability
restriction 
\[
|\gamma_{k}|\leq\bar{\gamma}<1,\qquad k=1,2,
\]
with 
\[
|\gamma_{k0}|\leq\bar{\gamma}<1,\qquad k=1,2.
\]
Suppose also that $g_{10}(x_{i})>0$ and $g_{20}(x_{i})>0$ almost
surely, and that $g_{10}(x_{i})/g_{20}(x_{i})$ is not almost surely
constant. Then the only candidate $\gamma$ in the stable parameter
region for which 
\[
\operatorname{Cov}(r_{\gamma,1i},r_{\gamma,2i}\mid x_{i})=0
\]
almost surely is $\gamma=\gamma_{0}$. Consequently, $\gamma_{0}$
is the unique maximizer of the unrestricted profiled population criterion
$Q^{*}(\gamma)$ in the stable parameter region.

\end{lemma}

The proof is given in Appendix~A.2.

Lemma~\ref{lem:unique-cond-diag} formalizes the algebraic identification
mechanism. At the true interaction parameter, the transformed residuals
are the structural shocks and have diagonal conditional covariance.
Variation in $g_{10}(x_{i})/g_{20}(x_{i})$ therefore rules out any
incorrect stable value of $\gamma$ that diagonalizes the conditional
residual covariance matrix almost surely.

The compact stability restriction rules out reciprocal algebraic solutions
that can arise from diagonalizing a two-dimensional covariance matrix
but do not correspond to the stable structural feedback system of
interest. This restriction is also consistent with the implementation,
where the bounded reparameterization 
\[
\gamma_{k}=\bar{\gamma}\tanh(\widetilde{\gamma}_{k}),\qquad k=1,2,
\]
keeps the fitted system away from the singular-feedback boundary.

We now transfer the function-level identification argument to the
neural-network parameterization. Because neural-network weights are
generally nonunique, the result is stated on the quotient slice $\mathcal{T}_{\eta_{0}}$
introduced in Section~\ref{subsec:populationrisketc}. The theorem
below shows that the neural profiled criterion inherits the same local
curvature as the unrestricted profiled criterion. 

\begin{theorem}[Local neural identification inherited from profiled curvature]

\label{thm:local-identification}

Suppose Assumptions~\ref{ass:structural-primitives} and \ref{ass:neural-profile-compatibility}
hold at a regular representative $\eta_{0}\in\mathcal{E}_{0}$. Then,
for sufficiently small $\delta>0$, the local neural profiled population
criterion 
\[
Q_{\delta,\eta_{0}}(\gamma)=\sup_{a\in\mathcal{T}_{\eta_{0}},\,\|a\|<\delta}Q(\gamma,\eta_{0}+a)
\]
satisfies, for $h\to0$, 
\[
Q_{\delta,\eta_{0}}(\gamma_{0}+h)-Q_{\delta,\eta_{0}}(\gamma_{0})=-\frac{1}{2}h^{\top}I_{\gamma}^{*}h+o(\|h\|^{2}),
\]
where 
\[
I_{\gamma}^{*}=\frac{1}{d_{0}^{2}}\begin{pmatrix}\operatorname{E}\{\rho_{0}(x_{i})^{-1}\} & 1\\
1 & \operatorname{E}\{\rho_{0}(x_{i})\}
\end{pmatrix}.
\]
If $\rho_{0}(x_{i})$ is not almost surely constant, then $I_{\gamma}^{*}\succ0$.
Hence $\gamma_{0}$ is a strict local maximizer of $Q_{\delta,\eta_{0}}(\gamma)$,
and the directed interaction parameter is locally identified on the
quotient slice. The population first-order conditions hold on the
slice:
\[
\nabla_{\gamma}Q(\gamma_{0},\eta_{0})=0,\qquad\nabla_{\eta}Q(\gamma_{0},\eta_{0})[a]=0\quad\text{for all }a\in\mathcal{T}_{\eta_{0}}.
\]
Equivalently, in local neural-network coordinates, the same profiled
curvature is represented by the Schur complement 
\[
I_{\gamma}=J_{\gamma\gamma}-J_{\gamma\eta}J_{\eta\eta}^{+}J_{\eta\gamma},
\]
where $J=-\nabla^{2}Q(\gamma_{0},\eta_{0})$ and $J_{\eta\eta}^{+}$
denotes the Moore--Penrose inverse. Under Assumption~\ref{ass:neural-profile-compatibility},
the quotient-slice curvature satisfies $I_{\gamma}=I_{\gamma}^{*}$.
If, in addition, the full nuisance parameterization has no competing
local nuisance maximizer outside the quotient-slice neighborhood,
then $\gamma_{0}$ is locally identified in the full neural-network
parameterization.

\end{theorem}

The proof is given in Appendix~A.3.

Assumption~\ref{ass:neural-profile-compatibility} ensures that the
neural mean and variance networks can locally reproduce the profiled
conditional mean and diagonal variance functions, after removing function-preserving
reparameterizations of the network weights.

This separation between identification and approximation is important
for the machine-learning formulation of the method. The neural networks
are not assumed to identify $\gamma$ by themselves. A sufficiently
flexible single-equation neural predictor can learn the reduced-form
association between $y_{1}$, $y_{2}$, and $x$, but such predictive
association does not distinguish simultaneous causal feedback from
residual dependence. Identification comes from requiring that, after
the candidate interaction parameter is used to transform the outcomes,
the remaining residual components can be represented with diagonal
conditional variances. When the two structural variances vary nonproportionally
over the feature space, this diagonalization requirement pins down
the two directions of interaction.

The variance networks therefore play an identifying, rather than merely
predictive, role. They do not merely estimate uncertainty around the
fitted structural means. They approximate the feature-dependent variance
ratio through which the maintained conditional-moment restrictions
generate curvature in the profiled quasi-likelihood. This is also
why the empirical implementation uses diagnostics for the fitted variance
ratio and for the local profiled loss surface around $\widehat{\gamma}$.
If the fitted variance ratio is nearly constant, or if the local profile
is nearly flat, the data contain weak heteroscedastic information
for separating the two directed effects.

These results are identification statements within the maintained
simultaneous structural model. They show that the observational conditional
second moments uniquely determine the stable structural interaction
coefficients when the variance ratio is nonconstant. Their causal
interpretation additionally depends on the structural-autonomy and
intervention-invariance conditions in Assumption~\ref{ass:structural-causal-interpretation}.

\begin{corollary}[Conditional causal interpretation]

\label{cor:cond-causal-interpret}

Under Assumption \ref{ass:structural-causal-interpretation}, the
locally identified structural interaction coefficients in Theorem~\ref{thm:local-identification}
are the direct causal interaction coefficients of the simultaneous
structural system.

\end{corollary}

\section{Learning algorithm}

\label{sec:learning-algorithm}

\subsection{Training challenges}

\label{subsec:training-challenges}

The identification results in Section~\ref{sec:Method} describe
the population source of identification for bidirectional interactions.
In finite samples, however, the proposed model must be trained as
a heteroscedastic neural regression system. This creates a nontrivial
optimization problem. The model contains neural networks for both
the conditional mean functions $f_{1},f_{2}$ and the conditional
variance functions $g_{1},g_{2}$. Consequently, variation in the
observed outcomes can be explained in two different ways: by changing
the fitted structural mean, or by changing the fitted conditional
variance. This mean--variance allocation problem is intrinsic to
likelihood-based heteroscedastic neural learning.

Recent work on heteroscedastic neural regression has shown that maximum-likelihood
training of mean and variance networks can produce undesirable behavior
\citep{Detlefsen2019,Immer2023b,Seitzer2022,Stirn2023}. In particular,
a flexible variance network may compensate for errors in the mean
network, thereby degrading the mean-function fit. Conversely, an overly
flexible mean network may absorb variation that should be represented
as conditional noise. The resulting variance estimates may be miscalibrated,
and the likelihood can favor solutions that are numerically favorable
but statistically misleading. These issues are especially relevant
in the present setting because the variance functions do more than
estimate uncertainty; they also provide identifying variation for
the causal interaction parameters.

The difficulty is amplified by simultaneity. In a standard heteroscedastic
regression problem, the target is usually a conditional mean or a
predictive distribution. In the present model, the main targets are
the structural interaction parameters $\gamma_{1}$ and $\gamma_{2}$.
The residuals used in the likelihood are (\ref{eq:resid1}) and (\ref{eq:resid2}).
Thus, errors in the mean networks, variance networks, or interaction
parameters all affect the same likelihood terms. A naive neural predictor
can easily learn the association between $y_{1}$ and $y_{2}$, but
such an association does not by itself distinguish structural feedback
from reduced-form dependence. The role of the heteroscedastic likelihood
is to distinguish structural feedback from reduced-form dependence
by searching for interaction parameters that make the structural residuals
conditionally diagonal. This separation can be fragile when the nuisance
networks overfit.

Another practical issue is variance collapse. Since the diagonal Gaussian
quasi-likelihood contains the terms $\log g_{k}(x_{i};\alpha_{k})$
and $e_{ki}^{2}/g_{k}(x_{i};\alpha_{k})$, very small fitted variances
can dominate the objective. If a variance network assigns an artificially
small variance to observations with small residuals, the likelihood
may improve without producing a reliable representation of the conditional
noise process. This behavior can destabilize gradient-based optimization
and contaminate the estimation of $\gamma$. For this reason, the
variance functions are parameterized using a softplus link and a positive
lower bound, as described in Section~\ref{subsec:causal-learning-model}.

The model is also sensitive to regularization. The regularization
strength on the mean networks controls how much outcome variation
is absorbed by $f_{1}$ and $f_{2}$, whereas the regularization strength
on the variance networks controls how flexible the heteroscedastic
identifying signal is allowed to be. If the variance networks are
too restricted, the model may fail to exploit the nonproportional
variation in the conditional variances required for identification.
If they are too flexible, the fitted variances may overfit idiosyncratic
residual patterns. Similarly, overly flexible mean networks can reduce
the residual variation from which heteroscedastic identification is
obtained. Hence the regularization parameters directly affect the
practical balance between nuisance-function learning and structural-parameter
estimation.

These considerations motivate the training strategy developed below.
Section~\ref{subsec:training-algorithm} introduces a stabilized
penalized quasi-likelihood objective, bounded parameterization of
the interaction coefficients, validation-based hyperparameter selection,
and early stopping. Section~\ref{subsec:diagnostics} then describes
diagnostics designed to check whether the fitted model contains enough
heteroscedastic variation to support identification and whether the
selected objective provides informative local evidence about the interaction
parameters.

\subsection{Training algorithm and stabilization}

\label{subsec:training-algorithm}

The model is trained using stabilized stochastic update rules based
on the negative diagonal Gaussian quasi-log-likelihood. Since the
mean and variance functions are represented by flexible neural networks,
the training problem involves a high-dimensional nuisance component
in addition to the low-dimensional structural interaction parameter.
We therefore use explicit regularization, a bounded reparameterization
of the interaction coefficients, validation-based early stopping,
and $\beta$-weighted negative-log-likelihood ($\beta$-NLL) stabilization
designed to reduce inverse-variance gradient amplification.

For notational brevity, write $g_{ki}(\alpha_{k})=g_{k}(x_{i};\alpha_{k})$. 

Ignoring the additive constant $\log(2\pi)$, define the per-observation
negative diagonal Gaussian quasi-log-likelihood contribution by 
\begin{align*}
q_{i}(\theta) & =-\ell_{i}(\theta)-\log(2\pi)\\
 & =-\log\left|1-\gamma_{1}\gamma_{2}\right|\\
 & \quad+\frac{1}{2}\sum_{k=1}^{2}\left\{ \log g_{ki}(\alpha_{k})+\frac{e_{ki}(\gamma,\nu_{k})^{2}}{g_{ki}(\alpha_{k})}\right\} .
\end{align*}
For a mini-batch $\mathcal{B}$, the original SEM-DNN training objective
is

\begin{equation}
\mathcal{L}_{\mathcal{B}}^{(0)}(\theta;\phi)=\frac{1}{|\mathcal{B}|}\sum_{i\in\mathcal{B}}q_{i}(\theta)+\mathcal{R}(\theta;\phi).\label{eq:penalizedqll}
\end{equation}
The regularization term is
\[
\mathcal{R}(\theta;\phi)=\sum_{k=1}^{2}\left\{ \phi_{\nu k}\varsigma(\nu_{k})+\phi_{\alpha k}\varsigma(\alpha_{k})\right\} ,
\]
where
\[
\phi=\left(\phi_{\nu1},\phi_{\nu2},\phi_{\alpha1},\phi_{\alpha2}\right)^{\top}.
\]
The penalty function is
\[
\varsigma(\vartheta)=\Vert\vartheta\odot\iota_{\vartheta}\Vert_{2}^{2},
\]
where $\iota_{\vartheta}$ is a mask vector selecting the weight parameters
and $\odot$ denotes elementwise multiplication. Bias parameters are
not penalized. The structural coefficients $\gamma_{1}$ and $\gamma_{2}$
are left unpenalized because they are the low-dimensional target parameters;
penalizing them would change the target criterion and introduce direct
shrinkage bias.

The baseline objective corresponds to standard quasi-maximum likelihood
training. However, likelihood-based heteroscedastic neural regression
can be sensitive to the interaction between mean-network errors and
variance-network estimates. In particular, the gradient of the residual
term with respect to the fitted mean is weighted by $1/g_{ki}(\alpha_{k})$.
Observations assigned large fitted variances therefore contribute
weakly to the mean-network gradient, while observations assigned small
fitted variances may receive high leverage. This inverse-variance
weighting is desirable when $g_{ki}(\alpha_{k})$ represents genuine
conditional noise dispersion, but it can be harmful during training
when the variance network temporarily or spuriously compensates for
mean-network error. In the present model, this issue is especially
important because the conditional variances not only quantify predictive
uncertainty but also provide the nonproportional variation that identifies
$\gamma$.

To address this problem, we adopt the $\beta$-NLL stabilization of
\citep{Seitzer2022} for the SEM framework. Let $\operatorname{sg}[\cdot]$
denote the stop-gradient operator, and let $\beta\in[0,1]$ denote
the variance-weighting exponent. For $\beta=0$, the stopped-gradient
weight equals one, so the update coincides with the original quasi-likelihood
update. Larger values of $\beta$ reduce the dependence of the training
gradients on inverse-variance weights.

We apply a single stopped-gradient weight to the entire observation-level
quasi-likelihood contribution, including the Jacobian term. Define
the observation-level variance scale
\[
s_{i}(\alpha_{1},\alpha_{2})=\left\{ g_{1i}(\alpha_{1})g_{2i}(\alpha_{2})\right\} ^{1/2}.
\]
For a mini-batch $\mathcal{B}$, define the average observation-level
variance scale
\[
\bar{s}_{\mathcal{B}}(\alpha_{1},\alpha_{2})=\frac{1}{|\mathcal{B}|}\sum_{j\in\mathcal{B}}s_{j}(\alpha_{1},\alpha_{2}).
\]
The stopped-gradient observation weight is

\[
\operatorname{sg}\left[\left\{ \frac{s_{i}(\alpha_{1},\alpha_{2})}{\bar{s}_{\mathcal{B}}(\alpha_{1},\alpha_{2})}\right\} ^{\beta}\right].
\]
The stabilized objective is
\begin{align*}
\mathcal{L}_{\mathcal{B}}^{(\beta)}(\theta;\beta,\phi)= & \frac{1}{|\mathcal{B}|}\sum_{i\in\mathcal{B}}\operatorname{sg}\left[\left\{ \frac{s_{i}(\alpha_{1},\alpha_{2})}{\bar{s}_{\mathcal{B}}(\alpha_{1},\alpha_{2})}\right\} ^{\beta}\right]q_{i}(\theta)\\
 & +\mathcal{R}(\theta;\phi).
\end{align*}
All parameters, including $\gamma$, are updated using the gradient
of $\mathcal{L}_{\mathcal{B}}^{(\beta)}(\theta;\beta,\phi)$. The
corresponding update direction is
\[
d_{\theta}^{(\beta)}=-\nabla_{\theta}\mathcal{L}_{\mathcal{B}}^{(\beta)}(\theta;\beta,\phi).
\]
This specification preserves the observation-level structure of the
quasi-likelihood contribution, because the Jacobian and residual terms
are weighted together. It may therefore be preferable when the joint
learning of interaction parameters and conditional variances is empirically
important. At the same time, $\mathcal{L}_{\mathcal{B}}^{(\beta)}(\theta;\beta,\phi)$
is no longer the original quasi-likelihood when $\beta>0$. We therefore
interpret it as a stabilized training criterion rather than as the
likelihood used for model comparison.

The fitted model is evaluated using the original unpenalized validation
negative quasi-log-likelihood (VNQLL). For a validation set $\mathcal{B}_{\mathrm{val}}$,
define
\begin{equation}
\mathcal{L}_{\mathrm{val}}^{(0)}(\theta)=\frac{1}{|\mathcal{B}_{\mathrm{val}}|}\sum_{i\in\mathcal{B}_{\mathrm{val}}}q_{i}(\theta).\label{eq:valnqll0}
\end{equation}
The $\beta$-weighted objectives are not used as validation criteria,
because their numerical values depend on the fitted variance weights
and do not have the same likelihood interpretation as $\mathcal{L}_{\mathrm{val}}^{(0)}(\theta)$.
The validation criterion is used both for early stopping and for comparing
hyperparameter configurations.

We impose the compact stability restriction
\[
|\gamma_{k}|\leq\bar{\gamma}<1,\qquad k=1,2.
\]
This restriction implies
\[
1-\gamma_{1}\gamma_{2}\geq1-\bar{\gamma}^{2}>0,
\]
and therefore prevents the simultaneous system from approaching singularity.
In computation, the restriction is imposed through the differentiable
reparameterization
\[
\gamma_{k}=\bar{\gamma}\tanh(\widetilde{\gamma}_{k}),\qquad\widetilde{\gamma}_{k}\in\mathbb{R},\qquad k=1,2.
\]

The unconstrained parameters $\widetilde{\gamma}_{1}$ and $\widetilde{\gamma}_{2}$
are optimized jointly with the neural-network parameters. This parameterization
prevents the training path from approaching explosive feedback systems
or nearly singular transformations between observed outcomes and structural
shocks.

The full training procedure is as follows. For each candidate regularization
vector $\phi$, each candidate value of $\beta$, and each training
rule in the candidate set, the model is initialized and trained using
mini-batch stochastic gradient optimization. At each prespecified
validation epoch, the current parameters are evaluated on the validation
set using $\mathcal{L}_{\mathrm{val}}^{(0)}(\theta)$. Within the
prespecified minimum and maximum training lengths, training is stopped
when the validation criterion fails to decrease by at least a prespecified
tolerance for a prespecified number of consecutive validation evaluations.
In the baseline implementation, validation is performed after every
epoch. The retained estimate is taken from the evaluated epoch with
the smallest validation loss, with ties resolved in favor of the earliest
epoch.

In the implementation, $\beta$ can be tuned manually or selected
from a grid. In this study, we set $\beta=0.5$ by default, motivated
by the simulation evidence in \citep{Seitzer2022}, and conduct an
ablation study of this choice. 

\subsection{Diagnostics for identification and training reliability}

\label{subsec:diagnostics}

Numerical convergence alone does not establish that a fitted model
contains the nonproportional heteroscedastic variation required to
identify the interaction parameters. We therefore propose a set of
diagnostics to evaluate three complementary aspects of the fitted
model on a held-out diagnostic sample that is not used for parameter
updating, hyperparameter selection, or early stopping. The diagnostics
assess whether the fitted model exhibits the observable implications
and curvature needed by the proposed identification argument. They
can reveal weak identifying variation or apparent incompatibility
with the selected conditional-moment restrictions, but they cannot
verify structural exogeneity, shock orthogonality, or intervention
invariance.

First, we assess the degree of nonproportional variation in the fitted
structural variances. Let $n_{d}$ denote the diagnostic-sample size,
and let
\begin{align*}
\widehat{e}_{1i} & =y_{1i}-\widehat{\gamma}_{1}y_{2i}-f_{1}(x_{i};\widehat{\nu}_{1}),\\
\widehat{e}_{2i} & =y_{2i}-\widehat{\gamma}_{2}y_{1i}-f_{2}(x_{i};\widehat{\nu}_{2}).
\end{align*}
Define the variance-ratio component of the fitted curvature matrix
as
\[
\widehat{H}_{\rho}=\begin{pmatrix}n_{d}^{-1}\sum_{i=1}^{n_{d}}\widehat{\rho}_{i}^{-1} & 1\\
1 & n_{d}^{-1}\sum_{i=1}^{n_{d}}\widehat{\rho}_{i}
\end{pmatrix},
\]
where 
\[
\widehat{\rho}_{i}=\frac{\widehat{g}_{1i}}{\widehat{g}_{2i}},\qquad\widehat{g}_{ki}=g_{k}(x_{i};\widehat{\alpha}_{k}),\qquad k=1,2.
\]
The plug-in analogue of the profiled curvature matrix $I_{\gamma}^{*}$
in Proposition~\ref{prop:profiled-curvature} is $\widehat{I}_{\gamma}^{*}=\widehat{d}^{-2}\widehat{H}_{\rho}$,
with 
\[
\widehat{d}=1-\widehat{\gamma}_{1}\widehat{\gamma}_{2}.
\]
To isolate the identifying variation in the variance ratio, we report
the minimum eigenvalue, condition number, and determinant of $\widehat{H}_{\rho}$.
The minimum eigenvalue summarizes the weakest variance-ratio direction,
while the condition number measures anisotropy across the two directions.
The determinant is nonnegative and equals zero if and only if the
fitted variance ratio is constant over the diagnostic sample. We also
inspect the empirical distribution of $\log\widehat{\rho}_{i}$ and
a scatter plot of $\log\widehat{g}_{1i}$ against $\log\widehat{g}_{2i}$.

Second, we assess approximate conditional diagonalization and conditional
variance calibration. At the true parameter values,
\[
\operatorname{E}\left[\frac{\varepsilon_{1i}\varepsilon_{2i}}{\sqrt{g_{10}(x_{i})g_{20}(x_{i})}}h(x_{i})\right]=0
\]
and
\[
\operatorname{E}\left[\left\{ \frac{\varepsilon_{ki}^{2}}{g_{k0}(x_{i})}-1\right\} h(x_{i})\right]=0,\qquad k=1,2,
\]
where $g_{10}$ and $g_{20}$ are the true conditional variance functions,
for suitable functions $h$. Multiplying these conditional restrictions
by suitable integrable functions of $x_{i}$ yields unconditional
moment restrictions. We evaluate a finite collection of such moments
using a feature vector specified before examining the diagnostic outcomes.
The resulting quadratic and componentwise discrepancy measures indicate
whether residual dependence or variance-calibration errors remain
along the selected covariate directions.

Third, we examine concentration in the residual-gradient weights induced
by inverse-variance weighting. For equation $k$, the residual-driven
part of the observation-weighted $\beta$-NLL gradient is proportional
to
\[
\widehat{\varpi}_{ki}^{(\beta)}=\frac{(\widehat{g}_{1i}\widehat{g}_{2i})^{\beta/2}}{\widehat{g}_{ki}},\qquad k=1,2,
\]
apart from a common normalization factor. For scale-comparable summaries,
we normalize these weights to have diagnostic-sample mean one and
report their upper quantiles, maxima, and normalized effective sample
sizes. A small effective sample size or a heavy upper tail indicates
that the residual-driven gradient is concentrated on relatively few
observations.

These quantities are descriptive model diagnostics rather than formal
specification tests. They condition on an estimated model, examine
only a finite collection of conditional-moment implications, and do
not incorporate first-stage estimation uncertainty or model-selection
uncertainty. They should therefore be interpreted jointly rather than
as independent pass-fail criteria. Full definitions and implementation
details are provided in Appendix~B.

\section{Simulation study}

\label{sec:simulation-study}

\subsection{Monte Carlo design}

\label{subsec:monte-carlo-design}

We evaluate the proposed estimator in a simulation environment designed
to reflect the main difficulties of heteroscedastic neural causal
learning: bidirectional feedback, nonlinear nuisance functions, feature-dependent
structural noise, irrelevant covariates, and non-Gaussian shocks.
Because there is no standard benchmark design for this problem, we
construct a design to make the identifying heteroscedastic signal
explicit while remaining flexible enough to test neural-network estimation.

For each Monte Carlo replication, data are generated from the simultaneous
structural system (\ref{eq:truesys1})--(\ref{eq:truesys2}). The
true bidirectional interaction parameters are fixed at $(\gamma_{1},\gamma_{2})=(0.5,0.4)$,
so that $1-\gamma_{1}\gamma_{2}=0.8$. Thus the system contains nonnegligible
feedback in both directions while remaining away from the singular
boundary. The observed outcomes are obtained by solving this two-equation
system for $(y_{1i},y_{2i})$. The covariate vector $x_{i}=(x_{i1},\ldots,x_{iJ})^{\top}$
is generated from a Gaussian copula with correlated coordinates and
bounded marginal support on $(0,1)$. We fix the feature dimension
at $J=100$. 

The structural mean functions $f_{1}$ and $f_{2}$ and the variance-index
functions $\widetilde{g}_{1}$ and $\widetilde{g}_{2}$ are nonlinear
functions of a subset of the covariates. The conditional variance
functions are specified as (\ref{eq:condval}) with $g_{\min}=0.01$.
The functional forms are taken from \citep{Bauer2019} and include
interactions among the covariates. The mean functions use coordinates
up to $x_{i7}$, and the variance-index functions use coordinates
up to $x_{i,10}$; the remaining coordinates are irrelevant. This
allows us to examine whether the estimator can recover the structural
interaction parameters when many observed features are available but
only a subset contains useful signal. The exact formulas for the covariate
generation, nonlinear benchmark functions, and mean and variance functions
are reported in Appendix~C.

We use three sample sizes: $n\in\{5000,10000,20000\}$. For each replication,
the scale of the structural mean functions is adjusted so that the
empirical signal-to-noise ratio in each structural equation is set
equal to $R\in\{4,1\}$ by construction. The case $R=4$ represents
a relatively favorable setting with a strong structural mean signal,
whereas $R=1$ represents a noisier setting in which accurate use
of heteroscedastic variation and regularization becomes more important.

The standardized shocks are generated from centered and scaled chi-squared
random variables. Hence the structural shocks are non-Gaussian, while
their conditional mean and variance follow the model specification.
This makes the diagonal Gaussian objective a quasi-likelihood rather
than a correctly specified full likelihood.

The same data-generating process is used for all estimators. The parameter
$\beta$ in the SEM-$\beta$-NLL procedures is therefore not part
of the data-generating process; it is a training-stabilization parameter
used only in estimation. This distinction is important because the
Monte Carlo design evaluates whether stabilized heteroscedastic neural
training improves recovery of the structural interaction parameters
under the same underlying simultaneous-equation system.

\subsection{Estimators and implementation}

\label{subsec:estimators-and-implementation}

We compare the proposed estimator with three classes of alternatives.
The comparison is designed to disentangle the contributions of (i)
simultaneous-equation structure with heteroscedastic identification
and (ii) neural-network function approximation.

The neural simultaneous-equation model, denoted SEM-DNN, is the model
described in Sections~\ref{sec:Method} and \ref{sec:learning-algorithm}.
The conditional mean functions $f_{1}$ and $f_{2}$, and the variance-index
functions $\widetilde{g}_{1}$ and $\widetilde{g}_{2}$, are represented
by feedforward neural networks. We fix the $\beta$-stabilization
level at $\beta=0.5$. This value is treated as a default stabilization
level rather than as a parameter of the statistical model. Sensitivity
to $\beta$ is examined in the ablation studies. The fitted model
is selected using the original unpenalized VNQLL (\ref{eq:valnqll0}),
rather than the $\beta$-weighted training objective. SEM-DNN denotes
the architecture and likelihood model. Unless otherwise stated, it
is trained with the observation-level $\beta$-NLL stabilization at
$\beta=0.5$. 

The first benchmark, denoted SEM-PAB, is a parametric heteroscedastic
quasi-likelihood estimator with the same simultaneous-equation structure
as SEM-DNN. It approximates the unknown mean and variance-index functions
using beta-function-weighted polynomial bases, following the flexible
specification used in \citep{Milunovich2018}. This benchmark keeps
the same identification strategy as SEM-DNN but replaces neural networks
with a structured parametric basis expansion.

The second benchmark, denoted SEM-Kernel, also keeps the simultaneous-equation
and heteroscedastic quasi-likelihood structure, but represents the
unknown functions using random Fourier features for a radial basis
function kernel \citep{Rahimi2007}. This estimator provides a kernel-style
alternative to neural function approximation.

The third benchmark, denoted Single-DNN, ignores simultaneity and
estimates the two equations separately. It treats the right-hand-side
outcome in each equation as an ordinary predictor and fits two regularized
neural regressions using squared-error loss. This estimator does not
model conditional heteroscedasticity and does not use the diagonalization
argument in Section~\ref{subsec:identification}. It is included
to show the effect of reduced-form association and simultaneity bias
when the bidirectional structure is ignored.

All likelihood-based simultaneous-equation estimators, including SEM-DNN,
SEM-PAB, and SEM-Kernel, are trained using the same $\beta$-NLL stabilization
with $\beta=0.5$. We evaluate performance using mean bias, root mean
squared error (RMSE), and standard deviation across 100 Monte Carlo
replications. For likelihood-based estimators, we also report the
mean of the VNQLL. Finally, to assess computational cost, we report
average runtime including hyperparameter tuning, early stopping, and
final training.

\subsection{Main Monte Carlo results}

\label{subsec:main-mc-results}

\begin{figure*}
\begin{centering}
\subfloat[$R=4,\widehat{\gamma}_{1}$]{\begin{centering}
\includegraphics[scale=0.6]{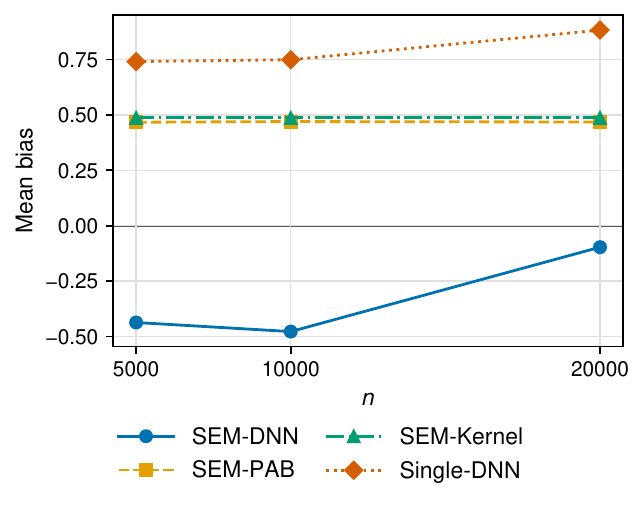}
\par\end{centering}
} \subfloat[$R=4,\widehat{\gamma}_{2}$]{\begin{centering}
\includegraphics[scale=0.6]{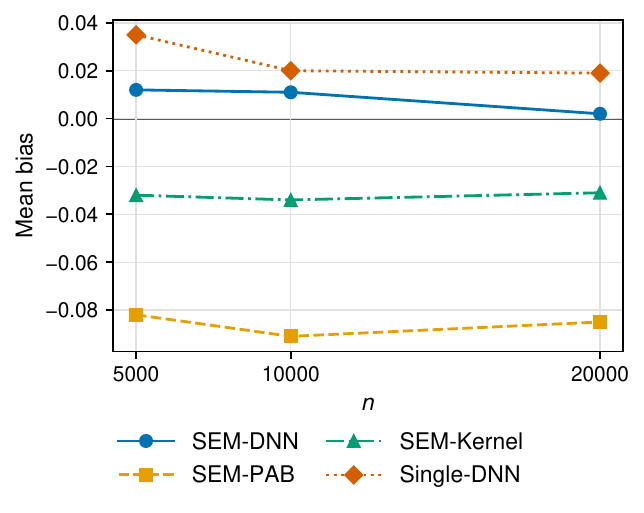}
\par\end{centering}
}\\
\subfloat[$R=1,\widehat{\gamma}_{1}$]{\begin{centering}
\includegraphics[scale=0.6]{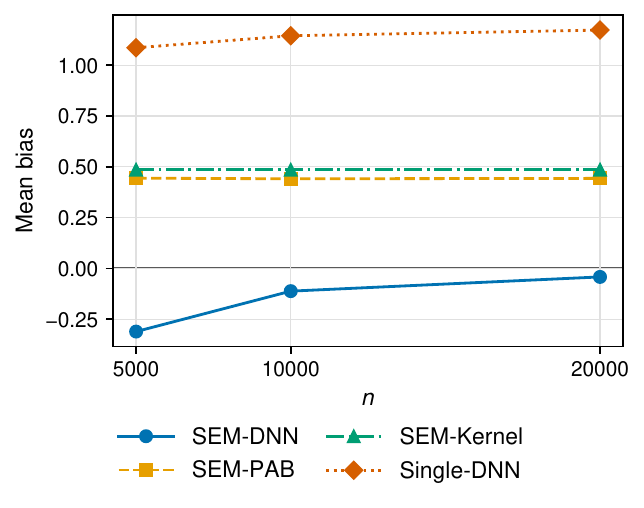}
\par\end{centering}
} \subfloat[$R=1,\widehat{\gamma}_{2}$]{\begin{centering}
\includegraphics[scale=0.6]{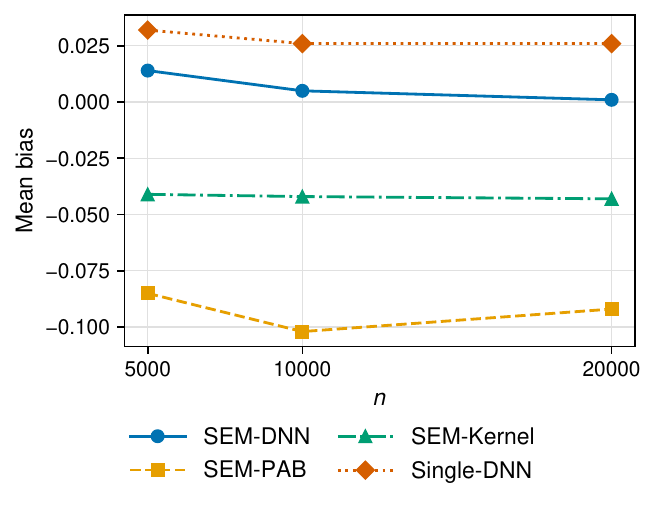}
\par\end{centering}
}
\par\end{centering}
\caption{Monte Carlo mean bias of the structural-interaction estimates across
100 replications. Panels report $\widehat{\gamma}_{1}$ and $\widehat{\gamma}_{2}$
for signal-to-noise ratios $R=4$ and $R=1$ as functions of sample
size.}
\label{fig-mean-bias}
\end{figure*}

\begin{figure*}
\begin{centering}
\subfloat[$R=4,\widehat{\gamma}_{1}$]{\begin{centering}
\includegraphics[scale=0.6]{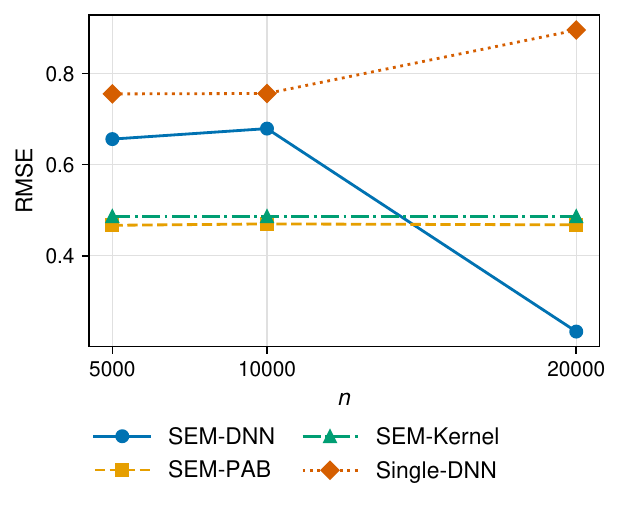}
\par\end{centering}
} \subfloat[$R=4,\widehat{\gamma}_{2}$]{\begin{centering}
\includegraphics[scale=0.6]{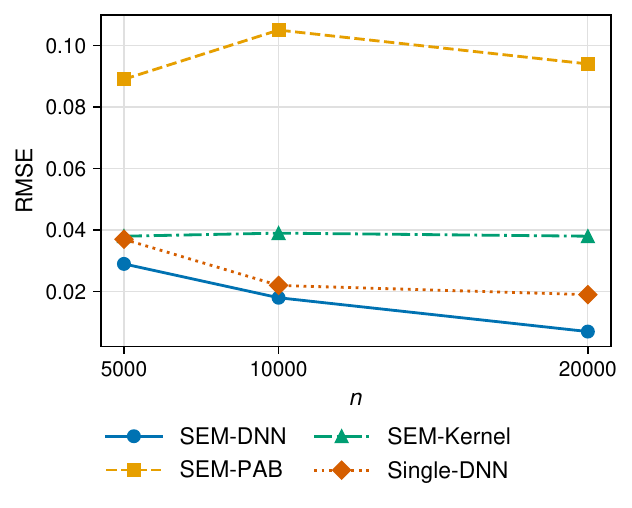}
\par\end{centering}
}\\
\subfloat[$R=1,\widehat{\gamma}_{1}$]{\begin{centering}
\includegraphics[scale=0.6]{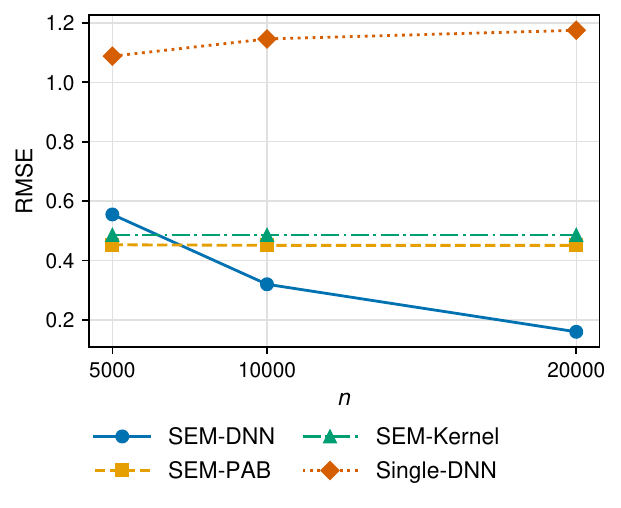}
\par\end{centering}
} \subfloat[$R=1,\widehat{\gamma}_{2}$]{\begin{centering}
\includegraphics[scale=0.6]{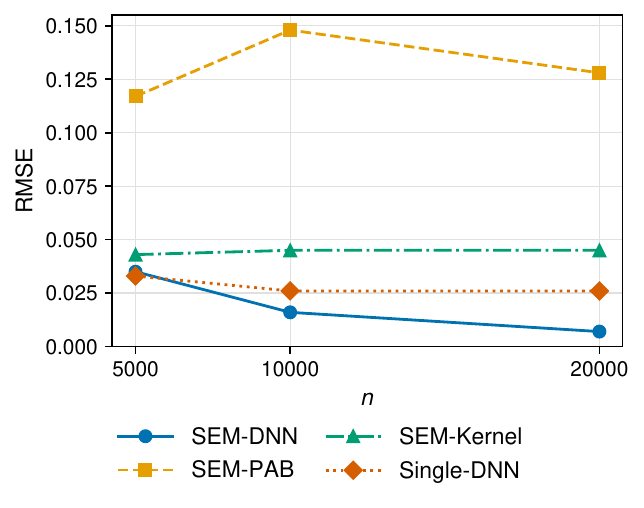}
\par\end{centering}
}
\par\end{centering}
\caption{Monte Carlo root mean squared error (RMSE) of the structural-interaction
estimates across 100 replications. Panels report $\widehat{\gamma}_{1}$
and $\widehat{\gamma}_{2}$ for signal-to-noise ratios $R=4$ and
$R=1$ as functions of sample size.}

\label{fig-rmse}
\end{figure*}

\begin{figure*}
\begin{centering}
\subfloat[$R=4,\widehat{\gamma}_{1}$]{\begin{centering}
\includegraphics[scale=0.6]{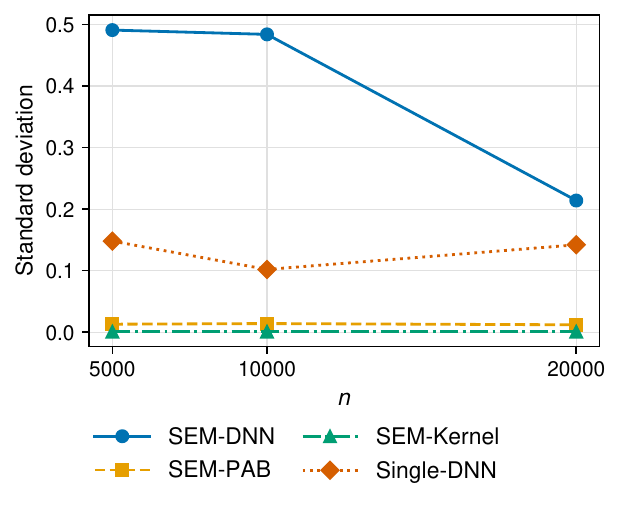}
\par\end{centering}
} \subfloat[$R=4,\widehat{\gamma}_{2}$]{\begin{centering}
\includegraphics[scale=0.6]{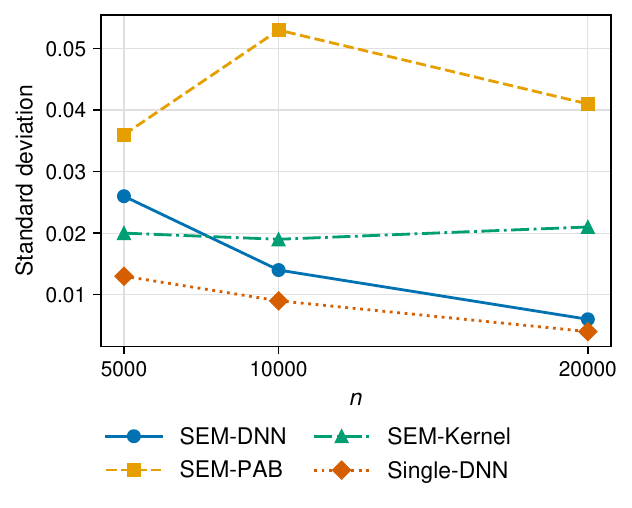}
\par\end{centering}
}\\
\subfloat[$R=1,\widehat{\gamma}_{1}$]{\begin{centering}
\includegraphics[scale=0.6]{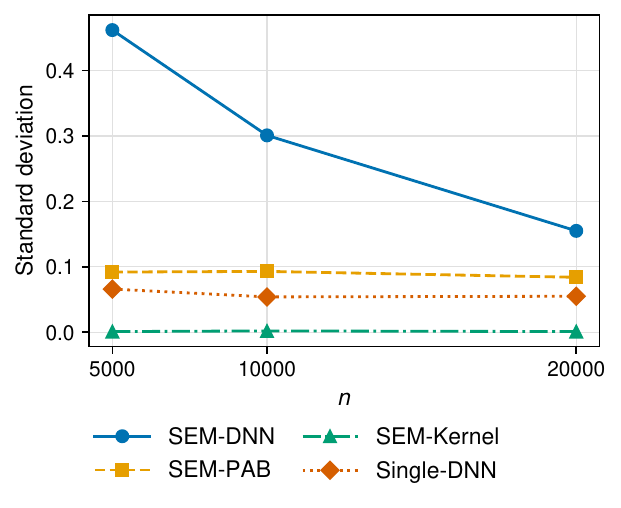}
\par\end{centering}
} \subfloat[$R=1,\widehat{\gamma}_{2}$]{\begin{centering}
\includegraphics[scale=0.6]{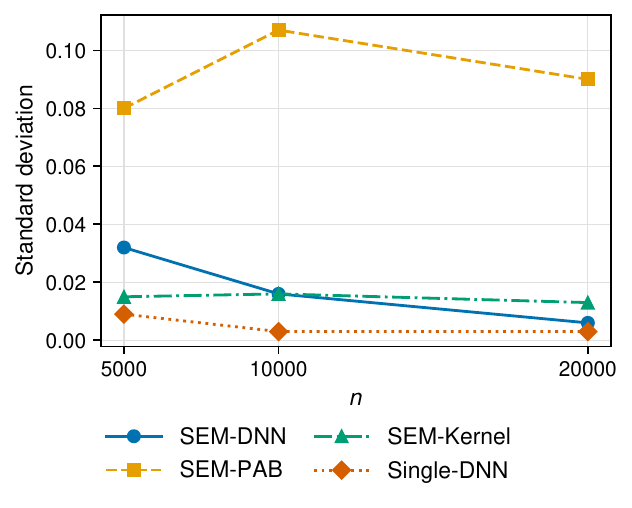}
\par\end{centering}
}
\par\end{centering}
\caption{Monte Carlo standard deviation of the structural-interaction estimates
across 100 replications. Panels report $\widehat{\gamma}_{1}$ and
$\widehat{\gamma}_{2}$ for signal-to-noise ratios $R=4$ and $R=1$
as functions of sample size.}

\label{fig-standard-deviation}
\end{figure*}

Figures~\ref{fig-mean-bias}--\ref{fig-standard-deviation} summarize
the Monte Carlo bias, RMSE, and standard deviation of $\widehat{\gamma}_{1}$
and $\widehat{\gamma}_{2}$ across estimators. SEM-DNN's accuracy
generally improves with sample size, particularly for $\gamma_{1}$,
although the pattern is not monotonic. In particular, for $R=4$,
the RMSE of $\widehat{\gamma}_{1}$ increases slightly from $n=5000$
to $n=10000$. Aggregate RMSE alone cannot distinguish the effects
of weak identification, nuisance-estimation error, and stochastic-optimization
variability; Appendix~D.10 examines these factors more directly.
In the original design, $\widehat{\gamma}_{2}$ remains nearly unbiased,
and its RMSE decreases as $n$ increases.

The competing estimators show different failure modes. SEM-PAB and
SEM-Kernel use the same heteroscedastic simultaneous-equation identification
strategy, but their restricted function approximations produce persistent
bias, most visibly for $\widehat{\gamma}_{1}$. This suggests that,
in the nonlinear design considered here, approximating the structural
mean and variance functions is not only a predictive issue but also
affects recovery of the interaction parameters. Single-DNN performs
poorly for $\widehat{\gamma}_{1}$ and its bias does not vanish with
larger samples. This is consistent with the simultaneity problem:
treating the other outcome as an ordinary regressor learns reduced-form
association rather than the structural interaction coefficient.

\begin{figure*}
\begin{centering}
\subfloat[$R=4$]{\begin{centering}
\includegraphics[scale=0.6]{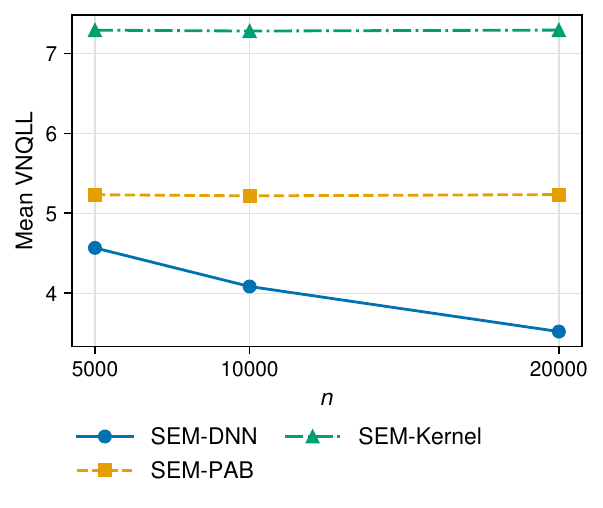}
\par\end{centering}
} \subfloat[$R=1$]{\begin{centering}
\includegraphics[scale=0.6]{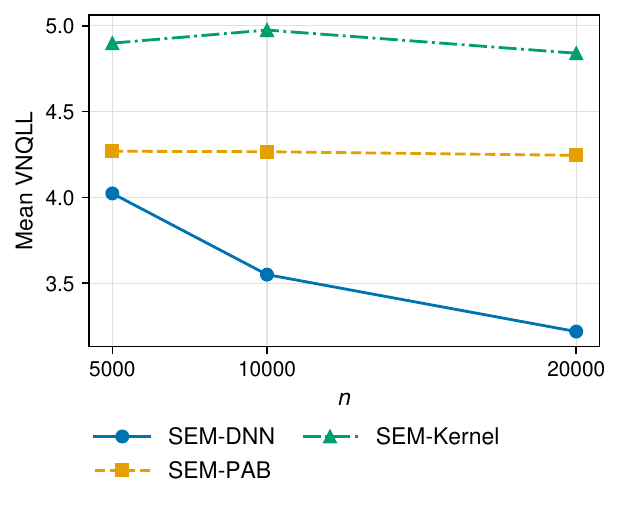}
\par\end{centering}
}
\par\end{centering}
\caption{Mean validation negative quasi-log-likelihood (VNQLL) across 100 Monte
Carlo replications for signal-to-noise ratios $R=4$ and $R=1$, plotted
as a function of sample size. Single-DNN is omitted because its criterion
is not comparable.}

\label{fig-mean-vnqll}
\end{figure*}

Figure~\ref{fig-mean-vnqll} reports the mean VNQLL for the likelihood-based
estimators. SEM-DNN attains the lowest validation criterion in both
signal-to-noise settings, and the criterion improves as $n$ increases.
In contrast, SEM-PAB is relatively flat across sample sizes, while
SEM-Kernel has a substantially worse validation fit. These results
support the interpretation that the neural specification provides
a better joint approximation of the structural means and conditional
variances, rather than merely reducing squared prediction error.

\begin{figure*}
\begin{centering}
\subfloat[$R=4$]{\begin{centering}
\includegraphics[scale=0.6]{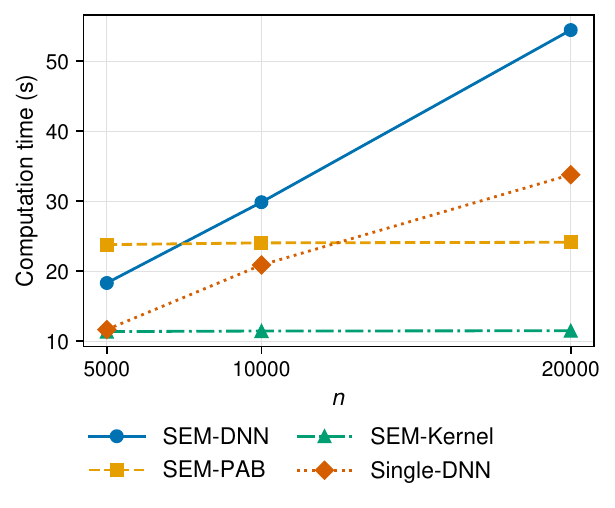}
\par\end{centering}
} \subfloat[$R=1$]{\begin{centering}
\includegraphics[scale=0.6]{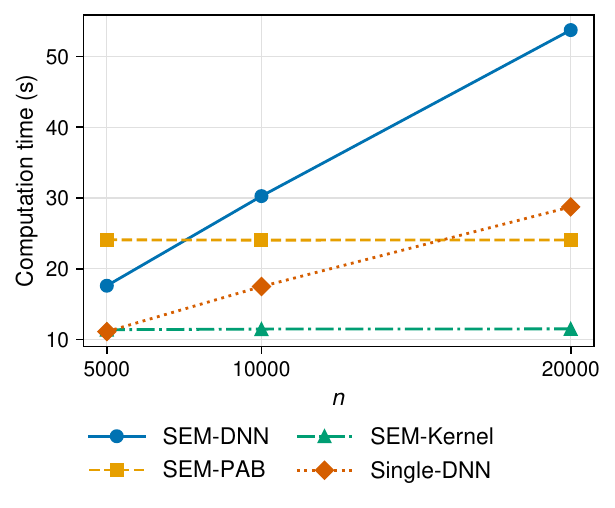}
\par\end{centering}
}
\par\end{centering}
\caption{Mean runtime in seconds across 100 Monte Carlo replications for signal-to-noise
ratios $R=4$ and $R=1$, including hyperparameter tuning, early stopping,
and final training. For Single-DNN, runtime is the sum of the runtimes
for the two separate optimizations.}

\label{fig-runtime}
\end{figure*}

Figure~\ref{fig-runtime} shows the computational cost. SEM-DNN is
more expensive than the parametric and kernel benchmarks, and its
runtime increases with $n$. The additional cost comes from tuning
and jointly training four neural networks together with the structural
interaction parameters. Thus, the main Monte Carlo evidence indicates
a clear accuracy--computation tradeoff: SEM-DNN delivers the most
reliable recovery of the bidirectional effects at the largest sample
size, but this improvement is obtained at higher computational cost.
Developing a faster algorithm for the proposed estimator is an important
direction for future work.

\subsection{Ablation studies}

\label{subsec:ablation-studies}

Sections~D.1--D.9 of the appendix report nine ablation studies.
In these ablations, the sample size is fixed at $n=10000$; all other
aspects of the design match those of the main Monte Carlo study unless
otherwise stated. Four additional studies are reported in Sections~D.10--D.13.

The first four ablations, reported in Sections~D.1--D.4, are directly
related to the identification mechanism and the training pathologies
discussed in Sections~\ref{sec:Method} and \ref{sec:learning-algorithm}.
The purpose is not to exhaustively tune the estimator, but to examine
whether the empirical behavior of SEM-DNN is consistent with the theory.
We focus on four questions: whether structural recovery deteriorates
when the two variance functions become nearly proportional, whether
regularization of the variance networks is necessary, whether regularization
of the mean networks is necessary, and how sensitive the estimator
is to the stabilization level $\beta$.

The remaining five ablations, reported in Sections~D.5--D.9, are
implementation-oriented stress tests. They examine sensitivity to
architecture size, estimation without the stability constraint on
$\gamma$, tied versus untied regularization parameters, the variance
lower bound $g_{\min}$, and the stability bound $\bar{\gamma}$.

The ablation results are broadly consistent with the proposed identification
and training arguments. When the two conditional variance functions
are made closer to proportional, the validation quasi-log-likelihood
deteriorates and recovery becomes more asymmetric across the two directed
effects. This supports the view that nonproportional heteroscedasticity
is not merely a variance-modeling feature, but a central source of
diagonalization information for identifying $\gamma_{1}$ and $\gamma_{2}$.
The regularization ablations show that the baseline results are not
driven mechanically by the variance-network penalty, while mean-network
regularization plays a stabilizing role in preserving the residual
variation used for heteroscedastic identification. The $\beta$-NLL
ablation further indicates that moderate stabilization, especially
$\beta=0.5$, improves training robustness relative to ordinary quasi-likelihood
training without changing the validation criterion.

The implementation-oriented stress tests also support the robustness
of the baseline specification. Moderate increases in network capacity
improve or preserve structural recovery, whereas the smallest architecture
is less effective. This suggests that sufficient approximation capacity
is important for learning the nonlinear mean and variance functions.
The bounded reparameterization of $\gamma$, untied regularization,
and the variance lower bound $g_{\min}$ mainly act as numerical stabilization
devices; changing these choices does not overturn the main conclusions.
Overall, the ablation studies indicate that SEM-DNN performs best
when the model retains enough flexibility to learn nonlinear heteroscedastic
structure while using moderate stabilization to control mean--variance
allocation and near-singular feedback behavior.

\subsection{Focused diagnostic and robustness experiments}

\label{subsec:focused-simulation}

Sections~D.10--D.13 of the appendix present four supplementary analyses.
Section~D.10 examines population identification, oracle estimation,
and initialization sensitivity, showing that weak identification alone
does not explain SEM-DNN's estimation variability. Section~D.11 compares
parameter-matched shallow, deep, and hybrid architectures: the baseline
achieves the best validation fit but does not consistently deliver
the most accurate structural estimates. Section~D.12 examines alternative
DGPs and strengthened kernel benchmarks, revealing coefficient-specific
differences in performance. 

Section~D.13 additionally compares SEM-DNN with Local ISA-LiNG \citep{Dai2024}
and BACKSHIFT \citep{Rothenhaeusler2015} using separate DGPs satisfying
the identifying assumptions of SEM-DNN and each competing method.
The designs feature linear feedback and identically zero structural
mean functions, whereas SEM-DNN continues to estimate two mean and
two variance networks. Local ISA-LiNG achieves substantially smaller
RMSEs in both of its designs, although SEM-DNN's RMSEs decrease markedly
under stronger heteroscedastic variation. In the BACKSHIFT experiment,
SEM-DNN achieves RMSEs of $0.045$ and $0.047$, compared with $0.037$
for both coefficients under BACKSHIFT. These results demonstrate the
accuracy advantages of specialized procedures in settings closely
aligned with their modeling assumptions, while showing that SEM-DNN
can recover structural coefficients under these overlapping restrictions.
They do not establish a comparative advantage from nonlinear nuisance-function
learning, which is unnecessary in the present DGPs. Together with
the nonlinear experiments and architectural comparisons elsewhere
in the paper, they clarify the tradeoff between the broader joint
learning formulation of SEM-DNN and the finite-sample efficiency of
specialized alternatives.

\section{Empirical illustration}

\label{sec:empirical-illustration}

\subsection{Data and empirical specification}

\label{subsec:data-and-specification}

We illustrate SEM-DNN using the ready-to-eat cereal category in the
Dominick’s scanner database (James M. Kilts Center, University of
Chicago Booth School of Business). The data combine weekly store--universal
product code (UPC) movement and price records with product attributes
and store-level demographic characteristics. Appendix~E provides
the full protocol for sample construction, variable definition, preprocessing,
and model selection.

Let $s$, $j$, and $t$ index stores, UPCs, and weeks. For each retained
store--UPC pair, we define a historical window $\mathcal{W}_{H}$,
a four-week exclusion gap, and a four-week focal window $\mathcal{W}_{0}$.
The empirical outcomes are
\[
y_{1,sj}=\log\left(\sum_{t\in\mathcal{W}_{0}}\texttt{MOVE}_{sjt}\right),
\]
and
\[
y_{2,sj}=\frac{1}{|\mathcal{W}_{0}|}\sum_{t\in\mathcal{W}_{0}}\log\left(\frac{\texttt{PRICE}_{sjt}}{\texttt{QTY}_{sjt}}\right).
\]
Here, $\texttt{MOVE}_{sjt}$ is the number of packages sold, $\texttt{PRICE}_{sjt}$
is the recorded total price, and $\texttt{QTY}_{sjt}$ is the number
of packages covered by that price, so that $\texttt{PRICE}_{sjt}/\texttt{QTY}_{sjt}$
is the unit package price. Thus, $\gamma_{1}$ measures the structural
effect of the average log package price (equivalently, the log of
the geometric mean package price) on log focal-window movement, whereas
$\gamma_{2}$ measures the reverse contemporaneous interaction.

The predetermined covariate vector $x_{sj}$ contains variables constructed
from the historical window, time-invariant product attributes, and
store characteristics. It includes historical movement, availability,
relative-price and sale-code measures, package information, store
demographics, and training-sample text-derived singular value decomposition
(SVD) features constructed using the training sample and product descriptions.
Variables constructed from focal-window outcomes are excluded. All
data-dependent quantities used for screening, imputation, scaling,
category pooling, text processing, and dimensionality reduction are
estimated using the training stores only. Learned outcome-network
embeddings are not used as diagnostic covariates.

Stores are partitioned into mutually exclusive training, validation,
and diagnostic samples. Neural-network hyperparameters, early stopping,
and seed-specific checkpoints are selected using the original unpenalized
VNQLL. Structural coefficients are summarized across numerically admissible
seeds by their componentwise medians and interquartile ranges. Model-specific
figures and diagnostics are based on a single representative run:
the seed at the lower median rank of the validation-loss ordering.
The diagnostic partition is used only after model selection and is
not used for retuning.

The baseline sample requires positive recorded movement and a valid
positive package price in each of the four focal weeks. It should
therefore be interpreted as a sample of store--UPC pairs with observed
prices and positive sales in every focal week. A causal interpretation
additionally requires
\[
\operatorname{Cov}\left(\varepsilon_{1,sj},\varepsilon_{2,sj}\mid x_{sj},S_{sj}=1\right)=0,
\]
where $S_{sj}=1$ denotes inclusion in the analysis sample. Unobserved
promotions, cost or demand shocks, inventory conditions, and endogenous
sample selection may violate this restriction. We consequently present
the application as an empirical illustration of SEM-DNN and its identification
diagnostics rather than definitive estimates of cereal demand and
pricing effects.

\subsection{Results}

\label{subsec:illustration-results}

The final sample contains 6,878 store--UPC pairs, of which 4,272,
949, and 1,657 belong to the training, validation, and diagnostic
partitions, respectively. All ten seed runs under the selected configuration
satisfy the numerical admissibility criteria. As reported in Table~\ref{tab:realdata-result},
the componentwise median estimates are $\widehat{\gamma}_{1}^{\mathrm{med}}=0.347$
and $\widehat{\gamma}_{2}^{\mathrm{med}}=-0.238$. Their interquartile
ranges across seeds, $[-0.123,0.573]$ and $[-0.560,0.586]$, are
wide and include zero, indicating substantial sensitivity to stochastic
optimization. These ranges summarize variation across admissible training
runs and should not be interpreted as confidence intervals. The representative
run---the seed at the lower median rank of the validation-loss ordering---gives
$\widehat{\gamma}_{1}=0.719$ and $\widehat{\gamma}_{2}=-0.651$,
with $1-\widehat{\gamma}_{1}\widehat{\gamma}_{2}=1.468$, safely away
from singularity.

\begin{table}
\caption{Results of the empirical application}

\label{tab:realdata-result}
\begin{centering}
\begin{tabularx}{\columnwidth}{>{\raggedright\arraybackslash}Xr}
\hline 
Quantity & \multicolumn{1}{r}{Reported value}\tabularnewline
\hline 
Median $\widehat{\gamma}_{1}$ & 0.347\tabularnewline
Interquartile range of $\widehat{\gamma}_{1}$ & {[}-0.123,0.573{]}\tabularnewline
Median $\widehat{\gamma}_{2}$ & -0.238\tabularnewline
Interquartile range of $\widehat{\gamma}_{2}$ & {[}-0.560,0.586{]}\tabularnewline
\hline 
Representative-run $\widehat{\gamma}_{1}^{(r^{\star})}$ & 0.719\tabularnewline
Representative-run $\widehat{\gamma}_{2}^{(r^{\star})}$ & -0.651\tabularnewline
$1-\widehat{\gamma}_{1}^{(r^{\star})}\widehat{\gamma}_{2}^{(r^{\star})}$ & 1.468\tabularnewline
$\mathcal{L}_{\mathrm{val}}^{(0)}(\widehat{\theta}^{(r^{\star})})$  & -2.578\tabularnewline
\hline 
$\lambda_{\min}(\widehat{H}_{\rho})$ & 0.056\tabularnewline
$\lambda_{\max}(\widehat{H}_{\rho})/\lambda_{\min}(\widehat{H}_{\rho})$ & 48.611\tabularnewline
$\det(\widehat{H}_{\rho})$ & 0.155\tabularnewline
\hline 
$J_{u}$ & 621.365\tabularnewline
$T_{u,\max}$ & 22.722\tabularnewline
$J_{z,1}$ & 59.309\tabularnewline
$J_{z,2}$ & 43.694\tabularnewline
$T_{z,1,\max}$ & 7.000\tabularnewline
$T_{z,2,\max}$ & 5.354\tabularnewline
$\widehat{\mu}_{z,1}$ & 1.428\tabularnewline
$\widehat{\mu}_{z,2}$ & 1.289\tabularnewline
\hline 
$\mathrm{IQR}_{\widetilde{\varpi}_{1}^{(\beta)}}$ & 0.529\tabularnewline
$\mathrm{IQR}_{\widetilde{\varpi}_{2}^{(\beta)}}$ & 0.373\tabularnewline
$Q_{0.95}(\widetilde{\varpi}_{1i}^{(\beta)})$ & 1.793\tabularnewline
$Q_{0.95}(\widetilde{\varpi}_{2i}^{(\beta)})$ & 1.320\tabularnewline
$\max(\widetilde{\varpi}_{1i}^{(\beta)})$ & 2.469\tabularnewline
$\max(\widetilde{\varpi}_{2i}^{(\beta)})$ & 1.921\tabularnewline
$\mathrm{ESS}_{1,d}^{(\beta)}$ & 0.861\tabularnewline
$\mathrm{ESS}_{2,d}^{(\beta)}$ & 0.933\tabularnewline
\hline 
Admissible runs & 10\tabularnewline
Failed runs & 0\tabularnewline
\hline 
\end{tabularx}\medskip{}
\par\end{centering}
\centering{}%
\noindent\begin{minipage}[t]{1\columnwidth}%
Note: Diagnostics are computed for the representative run $(r^{\star})$.%
\end{minipage}
\end{table}

\begin{figure*}
\begin{centering}
\subfloat[$\log\widehat{\rho}_{i}$]{\begin{centering}
\includegraphics[scale=0.6]{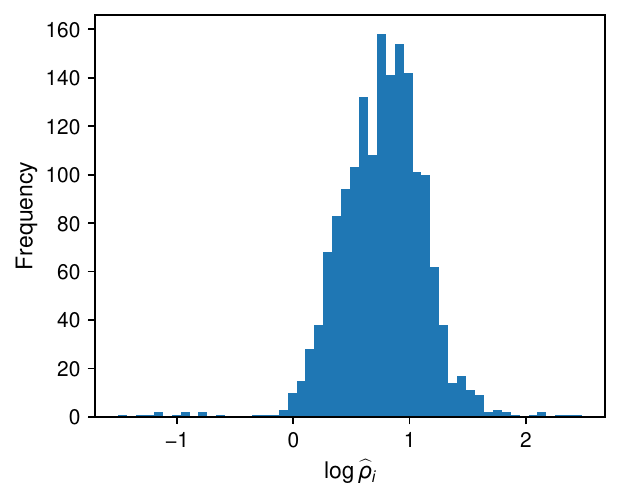}
\par\end{centering}
} \subfloat[$\log\widehat{g}_{1i},\log\widehat{g}_{2i}$]{\begin{centering}
\includegraphics[scale=0.6]{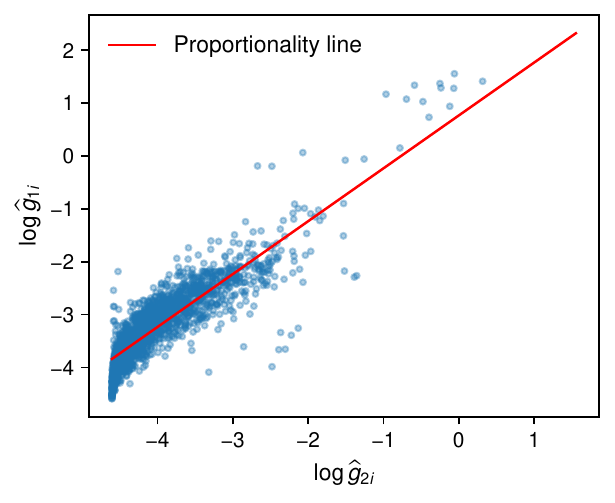}
\par\end{centering}
}
\par\end{centering}
\caption{Fitted variance-ratio diagnostics for the representative empirical
run. Panel (a) shows the distribution of $\log\widehat{\rho}_{i}=\log\widehat{g}_{1i}-\log\widehat{g}_{2i}$;
panel (b) plots $\log\widehat{g}_{1i}$ against $\log\widehat{g}_{2i}$,
with the proportionality line.}

\label{fig:dist-var-ratio}
\end{figure*}

The held-out diagnostics provide mixed evidence about the fitted structural
specification. The variance-ratio matrix has a minimum eigenvalue
of $0.056$ and a determinant of $0.155$. These positive values indicate
nonconstant fitted variance ratios and hence some heteroscedastic
variation relevant for identification. Figure~\ref{fig:dist-var-ratio}
corroborates this result: $\log\widehat{\rho}_{i}$ exhibits clear
cross-sectional dispersion, and the fitted log variances depart from
the proportionality line. However, the condition number of $48.611$
implies markedly uneven curvature across the two interaction directions.
Moreover, the residual-diagonalization discrepancies are relatively
large: $J_{u}=621.365$ and $T_{u,\max}=22.722$. They indicate remaining
conditional residual dependence along the examined covariate directions.
The variance-scaling statistics are $\widehat{\mu}_{z,1}=1.428$ and
$\widehat{\mu}_{z,2}=1.289$, showing that diagnostic-sample residual
dispersion exceeds the fitted variances on average. In contrast, the
normalized effective sample sizes of $0.861$ and $0.933$ suggest
that inverse-variance weighting is not concentrated on a small number
of observations.

The preceding coefficient summaries describe variation across ten
numerically admissible training seeds, whereas the held-out diagnostics
in Table~\ref{tab:realdata-result} refer to the prespecified representative
run. Appendix~E.11 therefore reports the fitted coefficient pair
and validation VNQLL for each seed and recomputes the held-out curvature,
residual-diagonalization, and variance-calibration diagnostics for
all ten saved fits. This analysis provides a descriptive assessment
of optimization sensitivity; the diagnostic sample is not used to
select or replace the representative run. Notably, two runs attain
nearly identical validation quasi-likelihood values despite yielding
opposite coefficient-sign patterns. Thus, favorable validation fit
and positive fitted curvature do not ensure stable structural estimates
or satisfactory held-out diagnostics.

The coefficient medians are comparatively stable across the robustness
specifications in Appendix~E. Changes to UPC support, historical
coverage, focal-price availability and aggregation, recorded-sale-code
restriction, price screening, outcome winsorization, observation weighting,
the sample split, and the focal window produce median estimates ranging
approximately from $0.294$ to $0.361$ for $\gamma_{1}$ and from
$-0.248$ to $-0.223$ for $\gamma_{2}$. Nevertheless, the wide dispersion
across seeds and the diagnostic discrepancies remain important qualifications.
The application therefore illustrates how SEM-DNN produces bidirectional
interaction estimates and evaluates the heteroscedastic variation
that identifies them, rather than yielding definitive causal elasticity
estimates for cereal quantity and price.

\section{Conclusion}

\label{sec:conclusion}

This paper develops SEM-DNN, a heteroscedastic neural simultaneous-equation
estimator for bidirectional contemporaneous structural interactions.
The method jointly estimates nonlinear structural mean functions,
feature-dependent structural variances, and two interaction coefficients
using a diagonal Gaussian quasi-likelihood with the simultaneous-system
Jacobian. Within the maintained structural model, nonproportional
variation in the conditional shock variances uniquely identifies the
stable interaction parameter through conditional covariance diagonalization.
Under neural-profile compatibility conditions, the implemented neural
criterion inherits the corresponding local population curvature despite
the nonuniqueness of network parameterizations. When the simultaneous
equations are interpreted as autonomous mechanisms invariant under
the relevant interventions, the identified coefficients are direct
causal interaction effects. 

The Monte Carlo results show that SEM-DNN's structural recovery depends
on the DGP, the interaction coefficient considered, and the quality
of joint nuisance-function estimation. Supplementary comparisons with
Local ISA-LiNG and BACKSHIFT demonstrate that specialized causal-learning
methods can achieve greater finite-sample accuracy and computational
efficiency in simplified linear settings satisfying their identifying
assumptions. In particular, Local ISA-LiNG achieves substantially
smaller RMSEs in both of its designs, while BACKSHIFT exhibits a more
modest accuracy advantage in its three-environment experiment. These
findings highlight the costs associated with flexible joint estimation
and caution against interpreting SEM-DNN as a uniformly superior estimator.

The methodological contribution of SEM-DNN lies in integrating adaptive
neural representation learning with simultaneous-equation estimation.
Unlike the fixed-feature kernel benchmark considered here, the proposed
framework learns nonlinear structural mean functions and feature-dependent
conditional variances jointly with reciprocal interaction coefficients
within a Jacobian-adjusted quasi-likelihood objective. This formulation
is supported by a conditional neural-profile curvature analysis, variance-aware
training stabilization, and held-out diagnostics for assessing identification
strength and model adequacy. It provides a complementary approach
for specified bidirectional systems with complex nuisance relationships,
where informative conditional variance variation is available but
independent non-Gaussian disturbances or designated shift-intervention
environments need not be assumed. Its statistical identification remains
grounded in the maintained conditional mean-zero, conditional shock-orthogonality,
and nonproportional variance assumptions, rather than in neural-network
flexibility itself.

The scanner application illustrates how the estimator can be combined
with held-out diagnostics for heteroscedastic identification strength,
residual diagonalization, variance calibration, and gradient leverage.
The fitted variance ratios contain relevant identifying variation,
but seedwise dispersion and conditional-moment discrepancies indicate
material empirical uncertainty. The application should therefore be
read as an illustration of assumption-dependent structural learning
rather than definitive estimates of cereal demand and pricing effects.

Developing valid inference remains an important topic for future research.
It must account for learned nuisance functions, model and checkpoint
selection, weak heteroscedastic identification, and nonconvex optimization.
More generally, causal interpretation depends on structural invariance,
conditional exogeneity, conditional shock orthogonality, and nonproportional
variance assumptions. These assumptions yield implications that can
be tested or assessed diagnostically, but they cannot be fully verified
from the observational distribution. Promising extensions include
weak-identification-robust inference, scalable optimization, multivariate
systems, and dynamic or time-varying feedback.

\bibliographystyle{elsarticle-num}
\bibliography{reference}



\section*{Supplementary material}

# Appendix to "Learning Bidirectional Causal Interactions with Heteroscedastic Neural Networks"

Masahiro Tanaka*

July 24, 2026

# A Proofs

## A.1 Proof of Proposition 1

Let

$$\Gamma_0 = \Gamma(\gamma_0), \qquad d_0 = \det \Gamma_0, \qquad D_0(x_i) = \operatorname{diag}\{g_{10}(x_i), g_{20}(x_i)\}.$$

Throughout Appendix A, we suppress the superscript $*$ on $v^*_{\gamma,k}$ to simplify notation. For a candidate value $\gamma = \gamma_0 + h$, where $h = (h_1, h_2)^\top$, the mean-profiled residual satisfies

$$r_{\gamma i} = \Gamma(\gamma) y_i - \mathrm{E}\{\Gamma(\gamma) y_i \mid x_i\} = \Gamma(\gamma)\Gamma_0^{-1}\varepsilon_i.$$

Hence its conditional covariance matrix is

$$V_\gamma(x_i) = \operatorname{Var}(r_{\gamma i} \mid x_i) = \Gamma(\gamma)\Gamma_0^{-1} D_0(x_i)\{\Gamma(\gamma)\Gamma_0^{-1}\}^\top.$$

Let

$$v_{\gamma,k}(x_i) = \operatorname{Var}(r_{\gamma,ki} \mid x_i), \qquad k = 1, 2,$$

denote the diagonal entries of $V_\gamma(x_i)$, and let

$$c_\gamma(x_i) = \operatorname{Cov}(r_{\gamma,1i}, r_{\gamma,2i} \mid x_i)$$

denote its off-diagonal entry.

By definition of the unrestricted profiled criterion,

$$Q^*(\gamma) = \log|\det \Gamma(\gamma)| - \frac{1}{2}\mathrm{E}\{\log v_{\gamma,1}(x_i) + \log v_{\gamma,2}(x_i)\} + C,$$

where $C$ is a constant independent of $\gamma$. Since

$$\det V_\gamma(x_i) = \left\{\frac{\det \Gamma(\gamma)}{\det \Gamma_0}\right\}^2 g_{10}(x_i) g_{20}(x_i),$$

we have the exact identity

$$Q^*(\gamma) - Q^*(\gamma_0) = -\frac{1}{2}\mathrm{E}\left[\log\left\{\frac{v_{\gamma,1}(x_i) v_{\gamma,2}(x_i)}{\det V_\gamma(x_i)}\right\}\right].$$

For a $2 \times 2$ covariance matrix,

$$\det V_\gamma(x_i) = v_{\gamma,1}(x_i) v_{\gamma,2}(x_i) - c_\gamma(x_i)^2.$$

Therefore

$$\frac{v_{\gamma,1}(x_i) v_{\gamma,2}(x_i)}{\det V_\gamma(x_i)} = \left\{1 - \frac{c_\gamma(x_i)^2}{v_{\gamma,1}(x_i) v_{\gamma,2}(x_i)}\right\}^{-1},$$

*Faculty of Economics, Fukuoka University, Fukuoka, Japan. Address: 8-19-1, Nanakuma, Jonan, Fukuoka, Japan 814-0180. E-mail: m.tanaka.tt@fukuoka-u.ac.jp.

and hence

$$Q^*(\gamma) - Q^*(\gamma_0) = \frac{1}{2}\,\mathrm{E}\left[\log\left\{1 - \frac{c_\gamma(x_i)^2}{v_{\gamma,1}(x_i)v_{\gamma,2}(x_i)}\right\}\right].$$

Thus the local behavior of $Q^*(\gamma)$ is determined by the conditional covariance induced by a misspecified transformation $\Gamma(\gamma)$.

We now expand this covariance. Since

$$\Gamma(\gamma)\Gamma_0^{-1} = \frac{1}{d_0}\begin{pmatrix} 1 - \gamma_1\gamma_{20} & \gamma_{10} - \gamma_1 \\ \gamma_{20} - \gamma_2 & 1 - \gamma_2\gamma_{10} \end{pmatrix},$$

with $d_0 = 1 - \gamma_{10}\gamma_{20}$, setting $\gamma = \gamma_0 + h$ gives

$$\Gamma(\gamma_0 + h)\Gamma_0^{-1} = \frac{1}{d_0}\begin{pmatrix} d_0 - h_1\gamma_{20} & -h_1 \\ -h_2 & d_0 - h_2\gamma_{10} \end{pmatrix}.$$

Therefore

$$\begin{aligned} c_{\gamma_0+h}(x_i) &= \frac{1}{d_0^2}\left[(d_0 - h_1\gamma_{20})(-h_2)g_{10}(x_i) + (-h_1)(d_0 - h_2\gamma_{10})g_{20}(x_i)\right] \\ &= -\frac{1}{d_0}\{h_2 g_{10}(x_i) + h_1 g_{20}(x_i)\} + \frac{h_1h_2}{d_0^2}\{\gamma_{20}g_{10}(x_i) + \gamma_{10}g_{20}(x_i)\}. \end{aligned}$$

It follows that

$$c_{\gamma_0+h}(x_i) = -\frac{1}{d_0}\{h_2 g_{10}(x_i) + h_1 g_{20}(x_i)\} + O(\|h\|^2\{g_{10}(x_i) + g_{20}(x_i)\}).$$

Also,

$$v_{\gamma_0+h,1}(x_i) = g_{10}(x_i) + O(\|h\|)\{g_{10}(x_i) + g_{20}(x_i)\},$$

and

$$v_{\gamma_0+h,2}(x_i) = g_{20}(x_i) + O(\|h\|)\{g_{10}(x_i) + g_{20}(x_i)\}.$$

Consequently,

$$\frac{c_{\gamma_0+h}(x_i)^2}{v_{\gamma_0+h,1}(x_i)v_{\gamma_0+h,2}(x_i)} = \frac{1}{d_0^2}\frac{\{h_2 g_{10}(x_i) + h_1 g_{20}(x_i)\}^2}{g_{10}(x_i)g_{20}(x_i)} + o(\|h\|^2)$$

in expectation. Under the bounded-ratio condition, the log-correlation term is locally dominated by

$$C\|h\|^2\{1 + \rho_0(x_i) + \rho_0(x_i)^{-1}\},$$

for all sufficiently small $\|h\|$, where the dominating random variable is integrable. Hence dominated convergence justifies interchanging the second-order expansion and expectation.

Since the preceding ratio is $O(\|h\|^2)$, the expansion

$$\log(1 - z) = -z + o(z)$$

yields

$$\begin{aligned} Q^*(\gamma_0 + h) - Q^*(\gamma_0) &= -\frac{1}{2d_0^2}\,\mathrm{E}\left[\frac{\{h_2 g_{10}(x_i) + h_1 g_{20}(x_i)\}^2}{g_{10}(x_i)g_{20}(x_i)}\right] + o(\|h\|^2) \\ &= -\frac{1}{2d_0^2}\,\mathrm{E}\left[h_2^2\frac{g_{10}(x_i)}{g_{20}(x_i)} + 2h_1h_2 + h_1^2\frac{g_{20}(x_i)}{g_{10}(x_i)}\right] + o(\|h\|^2) \\ &= -\frac{1}{2d_0^2}\,\mathrm{E}\left[h_2^2\rho_0(x_i) + 2h_1h_2 + h_1^2\rho_0(x_i)^{-1}\right] + o(\|h\|^2). \end{aligned}$$

This proves the claimed local expansion. Equivalently,

$$Q^*(\gamma_0 + h) - Q^*(\gamma_0) = -\frac{1}{2}h^\top I_\gamma^* h + o(\|h\|^2),$$

where

$$I_\gamma^* = \frac{1}{d_0^2}\begin{pmatrix} \mathrm{E}\{\rho_0(x_i)^{-1}\} & 1 \\ 1 & \mathrm{E}\{\rho_0(x_i)\} \end{pmatrix}. \tag{A.1}$$

It remains to verify positive definiteness. The diagonal elements of $I_\gamma^*$ are positive because $\rho_0(x_i) > 0$ almost surely. Its determinant is

$$\det(I_\gamma^*) = \frac{1}{d_0^4}\left[\mathrm{E}\{\rho_0(x_i)\}\,\mathrm{E}\{\rho_0(x_i)^{-1}\} - 1\right].$$

By the Cauchy–Schwarz inequality,

$$\mathrm{E}\{\rho_0(x_i)\}\,\mathrm{E}\{\rho_0(x_i)^{-1}\} \geq \{\mathrm{E}(1)\}^2 = 1.$$

Equality holds if and only if $\rho_0(x_i)^{1/2}$ is proportional to $\rho_0(x_i)^{-1/2}$ almost surely, which is equivalent to $\rho_0(x_i)$ being almost surely constant. Since Assumption 2 rules out this case,

$$\mathrm{E}\{\rho_0(x_i)\}\,\mathrm{E}\{\rho_0(x_i)^{-1}\} > 1.$$

Therefore $\det(I_\gamma^*) > 0$, and hence $I_\gamma^* \succ 0$. This completes the proof. □

## A.2 Proof of Lemma 1

Fix a candidate value $\gamma$ in the stable parameter region. After the conditional mean functions have been profiled out,

$$r_{\gamma i} = \Gamma(\gamma)y_i - \mathrm{E}\{\Gamma(\gamma)y_i \mid x_i\}.$$

Since

$$\Gamma(\gamma_0)y_i = f_0(x_i) + \varepsilon_i$$

and $\mathrm{E}(\varepsilon_i \mid x_i) = 0$, we have

$$y_i = \Gamma_0^{-1} f_0(x_i) + \Gamma_0^{-1}\varepsilon_i,$$

where

$$\Gamma_0 = \Gamma(\gamma_0), \qquad d_0 = \det\Gamma_0 = 1 - \gamma_{10}\gamma_{20}.$$

Therefore

$$\mathrm{E}\{\Gamma(\gamma)y_i \mid x_i\} = \Gamma(\gamma)\Gamma_0^{-1} f_0(x_i),$$

and hence

$$r_{\gamma i} = \Gamma(\gamma)\Gamma_0^{-1}\varepsilon_i.$$

Let

$$V_\gamma(x_i) = \mathrm{Var}(r_{\gamma i} \mid x_i),$$

and denote its diagonal entries by

$$v_{\gamma,1}(x_i) = \mathrm{Var}(r_{\gamma,1i} \mid x_i), \qquad v_{\gamma,2}(x_i) = \mathrm{Var}(r_{\gamma,2i} \mid x_i).$$

Because $|\gamma_k| \leq \bar{\gamma} < 1$ and $|\gamma_{k0}| \leq \bar{\gamma} < 1$, both $\Gamma(\gamma)$ and $\Gamma_0$ are nonsingular. Since $g_{10}(x_i) > 0$ and $g_{20}(x_i) > 0$ almost surely, $V_\gamma(x_i)$ is positive definite almost surely.

For fixed $\gamma$, the unrestricted profiled diagonal Gaussian quasi-log-likelihood sets the diagonal conditional variance functions equal to the diagonal conditional second moments $v_{\gamma,1}(x_i)$ and $v_{\gamma,2}(x_i)$. Thus, up to constants that do not depend on $\gamma$,

$$Q^*(\gamma) = \log|\det\Gamma(\gamma)| - \frac{1}{2}\mathrm{E}\{\log v_{\gamma,1}(x_i) + \log v_{\gamma,2}(x_i)\}.$$

Moreover,

$$V_\gamma(x_i) = \Gamma(\gamma)\Gamma_0^{-1}\,\mathrm{diag}\{g_{10}(x_i), g_{20}(x_i)\}\{\Gamma(\gamma)\Gamma_0^{-1}\}^\top.$$

Taking determinants gives

$$\det V_\gamma(x_i) = \left\{\frac{\det\Gamma(\gamma)}{\det\Gamma_0}\right\}^2 g_{10}(x_i)g_{20}(x_i).$$

At the true value $\gamma_0$,

$$V_{\gamma_0}(x_i) = \mathrm{diag}\{g_{10}(x_i), g_{20}(x_i)\}.$$

Therefore,

$$Q^*(\gamma) = Q^*(\gamma_0) - \frac{1}{2}\mathrm{E}\left[\log\left\{\frac{v_{\gamma,1}(x_i)v_{\gamma,2}(x_i)}{\det V_\gamma(x_i)}\right\}\right].$$

For any positive definite $2 \times 2$ covariance matrix, the product of the diagonal entries is at least as large as the determinant. In the present notation,

$$v_{\gamma,1}(x_i)v_{\gamma,2}(x_i) - \det V_\gamma(x_i) = \operatorname{Cov}(r_{\gamma,1i}, r_{\gamma,2i} \mid x_i)^2 \geq 0.$$

Hence

$$\frac{v_{\gamma,1}(x_i)v_{\gamma,2}(x_i)}{\det V_\gamma(x_i)} \geq 1$$

almost surely, and consequently $Q^*(\gamma) \leq Q^*(\gamma_0)$. Equality holds if and only if

$$\operatorname{Cov}(r_{\gamma,1i}, r_{\gamma,2i} \mid x_i) = 0$$

almost surely. It remains to characterize the values of $\gamma$ satisfying this conditional diagonalization condition.

Since

$$\Gamma(\gamma)\Gamma_0^{-1} = \frac{1}{d_0}\begin{pmatrix} 1-\gamma_1\gamma_{20} & \gamma_{10}-\gamma_1 \\ \gamma_{20}-\gamma_2 & 1-\gamma_2\gamma_{10} \end{pmatrix},$$

Assumption 2 gives

$$\operatorname{Cov}(r_{\gamma,1i}, r_{\gamma,2i} \mid x_i) = \frac{1}{d_0^2}\Big[(1-\gamma_1\gamma_{20})(\gamma_{20}-\gamma_2)g_{10}(x_i) \\ + (\gamma_{10}-\gamma_1)(1-\gamma_2\gamma_{10})g_{20}(x_i)\Big].$$

Thus the zero-covariance condition is equivalent to

$$(1-\gamma_1\gamma_{20})(\gamma_{20}-\gamma_2)g_{10}(x_i) + (\gamma_{10}-\gamma_1)(1-\gamma_2\gamma_{10})g_{20}(x_i) = 0$$

almost surely. Dividing by $g_{20}(x_i) > 0$ gives

$$(1-\gamma_1\gamma_{20})(\gamma_{20}-\gamma_2)\frac{g_{10}(x_i)}{g_{20}(x_i)} + (\gamma_{10}-\gamma_1)(1-\gamma_2\gamma_{10}) = 0$$

almost surely.

Because $g_{10}(x_i)/g_{20}(x_i)$ is not almost surely constant, both coefficients in the preceding affine function of $g_{10}(x_i)/g_{20}(x_i)$ must be zero:

$$(1-\gamma_1\gamma_{20})(\gamma_{20}-\gamma_2) = 0,$$

and

$$(\gamma_{10}-\gamma_1)(1-\gamma_2\gamma_{10}) = 0.$$

The compact stability restriction implies

$$|\gamma_1\gamma_{20}| < 1, \qquad |\gamma_2\gamma_{10}| < 1.$$

Therefore

$$1-\gamma_1\gamma_{20} \neq 0, \qquad 1-\gamma_2\gamma_{10} \neq 0.$$

It follows that $\gamma_2 = \gamma_{20}$ and $\gamma_1 = \gamma_{10}$. Hence $\gamma = \gamma_0$ is the only candidate in the stable parameter region that diagonalizes the conditional covariance matrix of the profiled residuals.

Combining this uniqueness result with the likelihood comparison above shows that $\gamma_0$ is the unique maximizer of the unrestricted profiled population criterion $Q^*(\gamma)$ in the stable parameter region. This proves the lemma. □

## A.3 Proof of Theorem 1

We prove the result in three steps. First, we show that the local neural profiled criterion inherits the second-order expansion of the unrestricted profiled criterion. Second, we use this expansion to obtain strict local identification of $\gamma_0$ on the quotient slice. Third, we relate the inherited curvature to the Schur-complement representation in local neural-network coordinates.

Let $h = \gamma - \gamma_0$. By Assumption 3 (N2), the local neural profiled criterion satisfies

$$Q_{\delta,\eta_0}(\gamma_0 + h) = Q^*(\gamma_0 + h) + o(\|h\|^2), \qquad h \to 0.$$

At $h = 0$, the neural networks at $\eta_0 \in \mathcal{E}_0$ represent the true structural mean and conditional variance functions. Hence

$$Q_{\delta,\eta_0}(\gamma_0) = Q(\gamma_0, \eta_0) = Q^*(\gamma_0).$$

Therefore

$$Q_{\delta,\eta_0}(\gamma_0 + h) - Q_{\delta,\eta_0}(\gamma_0) = Q^*(\gamma_0 + h) - Q^*(\gamma_0) + o(\|h\|^2).$$

By Proposition 1,

$$Q^*(\gamma_0 + h) - Q^*(\gamma_0) = -\frac{1}{2} h^\top I_\gamma^* h + o(\|h\|^2),$$

where

$$I_\gamma^* = \frac{1}{d_0^2} \begin{pmatrix} \mathrm{E}\{\rho_0(x_i)^{-1}\} & 1 \\ 1 & \mathrm{E}\{\rho_0(x_i)\} \end{pmatrix}, \qquad \rho_0(x_i) = \frac{g_{10}(x_i)}{g_{20}(x_i)}.$$

Combining the preceding displays gives

$$Q_{\delta,\eta_0}(\gamma_0 + h) - Q_{\delta,\eta_0}(\gamma_0) = -\frac{1}{2} h^\top I_\gamma^* h + o(\|h\|^2).$$

This proves the claimed local expansion for the neural profiled criterion.

By Assumption 2, $\rho_0(x_i)$ is not almost surely constant. Proposition 1 then implies $I_\gamma^* \succ 0$. Therefore there exist constants $c > 0$ and $r_{\mathrm{loc}} > 0$ such that, for all $0 < \|h\| < r_{\mathrm{loc}}$,

$$h^\top I_\gamma^* h \geq c\|h\|^2$$

and the remainder term is dominated by the leading quadratic term. After possibly shrinking $r_{\mathrm{loc}}$,

$$Q_{\delta,\eta_0}(\gamma_0 + h) - Q_{\delta,\eta_0}(\gamma_0) \leq -\frac{c}{4}\|h\|^2 < 0$$

for every $0 < \|h\| < r_{\mathrm{loc}}$. Hence $\gamma_0$ is a strict local maximizer of $Q_{\delta,\eta_0}(\gamma)$. This establishes local identification of $\gamma_0$ on the quotient slice.

It remains to connect this curvature to the local representation in neural-network parameters. Write the local nuisance perturbations as $\eta = \eta_0 + a$, $a \in \mathcal{T}_{\eta_0}$. By Assumption 3 (N1), $Q$ is twice continuously differentiable near $(\gamma_0, \eta_0)$ on the slice. The first-order conditions at the true representative are

$$\nabla_\gamma Q(\gamma_0, \eta_0) = 0, \qquad \nabla_\eta Q(\gamma_0, \eta_0)[a] = 0 \quad \text{for all } a \in \mathcal{T}_{\eta_0}.$$

Thus, uniformly for $(h, a)$ close to zero with $a \in \mathcal{T}_{\eta_0}$,

$$Q(\gamma_0 + h, \eta_0 + a) - Q(\gamma_0, \eta_0) = -\frac{1}{2} \begin{pmatrix} h \\ a \end{pmatrix}^\top J \begin{pmatrix} h \\ a \end{pmatrix} + o(\|h\|^2 + \|a\|^2),$$

where

$$J = -\nabla^2 Q(\gamma_0, \eta_0) = \begin{pmatrix} J_{\gamma\gamma} & J_{\gamma\eta} \\ J_{\eta\gamma} & J_{\eta\eta} \end{pmatrix}.$$

By Assumption 3 (N3),

$$\ker(J_{\eta\eta}) = \mathcal{K}_{\eta_0}, \qquad \mathcal{T}_{\eta_0} = \mathcal{K}_{\eta_0}^\perp,$$

and

$$\ker(J_{\eta\eta}) \subseteq \ker(J_{\gamma\eta}).$$

Since $J_{\eta\eta}$ is symmetric,

$$\mathrm{Range}(J_{\eta\eta}) = \ker(J_{\eta\eta})^\perp.$$

For any $z \in \ker(J_{\eta\eta})$ and any $h \in \mathbb{R}^2$, the compatibility condition gives

$$z^\top J_{\eta\gamma} h = h^\top J_{\gamma\eta} z = 0.$$

Hence

$$J_{\eta\gamma} h \in \ker(J_{\eta\eta})^\perp = \mathrm{Range}(J_{\eta\eta}).$$

Therefore the Moore–Penrose expression

$$J_{\eta\eta}^+ J_{\eta\gamma} h$$

is well defined and lies in

$$\mathrm{Range}(J_{\eta\eta}) = \mathcal{T}_{\eta_0}.$$

For fixed $h$, define

$$a_h = -J_{\eta\eta}^{+} J_{\eta\gamma} h.$$

Completing the square gives

$$\begin{pmatrix} h \\ a \end{pmatrix}^{\top} J \begin{pmatrix} h \\ a \end{pmatrix} = h^{\top} \left( J_{\gamma\gamma} - J_{\gamma\eta} J_{\eta\eta}^{+} J_{\eta\gamma} \right) h + (a - a_h)^{\top} J_{\eta\eta} (a - a_h).$$

Thus the quadratic part of the locally profiled criterion is obtained by setting $a = a_h$. Since $a_h = O(\|h\|)$, it belongs to the ball $\{a \in \mathcal{T}_{\eta_0} : \|a\| < \delta\}$ for all sufficiently small $h$. Therefore

$$Q_{\delta,\eta_0}(\gamma_0 + h) - Q_{\delta,\eta_0}(\gamma_0) = -\frac{1}{2} h^{\top} I_{\gamma} h + o(\|h\|^2),$$

where

$$I_{\gamma} = J_{\gamma\gamma} - J_{\gamma\eta} J_{\eta\eta}^{+} J_{\eta\gamma}.$$

The same profile has already been shown to satisfy

$$Q_{\delta,\eta_0}(\gamma_0 + h) - Q_{\delta,\eta_0}(\gamma_0) = -\frac{1}{2} h^{\top} I_{\gamma}^{*} h + o(\|h\|^2).$$

Since both are second-order expansions of the same local profiled criterion, their quadratic forms must coincide. Hence $I_{\gamma} = I_{\gamma}^{*}$. In particular, $I_{\gamma} \succ 0$.

The preceding argument proves strict local identification on the quotient slice. If the full nuisance parameterization has no competing local nuisance maximizer outside the quotient-slice neighborhood, then every local maximizer of the full neural profile near $(\gamma_0, \mathcal{E}_0)$ is represented by a point on this slice. The strict local maximum on the quotient slice therefore implies local identification of $\gamma_0$ in the full neural-network parameterization. This completes the proof. □

# B Details of the diagnostics

The proposed estimator uses conditional heteroscedasticity to distinguish bidirectional structural interactions from reduced-form association. A fitted model may achieve a favorable validation quasi-likelihood while providing little information about $\gamma$, particularly when the fitted conditional variances are nearly proportional, the fitted structural residuals remain conditionally dependent, or inverse-variance weighting concentrates residual-driven gradients on a small subset of observations.

Model parameters and the training epoch are selected using the original unpenalized VNQLL rather than the $\beta$-weighted training criterion. After selection, the diagnostics are evaluated once on a held-out diagnostic sample. This separation limits mechanical overfitting of the diagnostic moments and places fitted specifications on a common quasi-log-likelihood scale. The diagnostics are descriptive model checks, not selection criteria or formal specification tests.

## B.1 Local identification diagnostics

Let $n_d$ denote the diagnostic-sample size and define

$$\widehat{\rho}_i = \frac{\widehat{g}_{1i}}{\widehat{g}_{2i}}, \qquad \widehat{g}_{ki} = g_k(x_i; \widehat{\alpha}_k), \qquad k = 1, 2,$$

and

$$\widehat{d} = 1 - \widehat{\gamma}_1 \widehat{\gamma}_2.$$

Define the variance-ratio component of the fitted curvature matrix by

$$\widehat{H}_{\rho} = \begin{pmatrix} \overline{\widehat{\rho}^{-1}} & 1 \\ 1 & \overline{\widehat{\rho}} \end{pmatrix},$$

where

$$\bar{\widehat{\rho}} = \frac{1}{n_d}\sum_{i=1}^{n_d}\widehat{\rho}_i, \qquad \overline{\widehat{\rho}^{-1}} = \frac{1}{n_d}\sum_{i=1}^{n_d}\widehat{\rho}_i^{-1}.$$

The plug-in analogue of the profiled curvature matrix $I_\gamma^*$ in Proposition 1 is

$$\widehat{I}_\gamma^* = \frac{1}{\widehat{d}^2}\widehat{H}_\rho.$$

We report

$$\lambda_{\min}(\widehat{H}_\rho), \qquad \kappa(\widehat{H}_\rho) = \frac{\lambda_{\max}(\widehat{H}_\rho)}{\lambda_{\min}(\widehat{H}_\rho)},$$

and

$$\det(\widehat{H}_\rho) = \overline{\widehat{\rho}}\,\overline{\widehat{\rho}^{-1}} - 1.$$

The determinant is nonnegative by the Cauchy–Schwarz inequality and equals zero if and only if $\widehat{\rho}_i$ is constant over the diagnostic sample. Thus, $\lambda_{\min}(\widehat{H}_\rho)$ measures the weakest fitted curvature direction attributable to nonproportional variation in the fitted variances, while a large condition number indicates substantial curvature anisotropy across the two structural-coefficient directions.

The empirical distribution of $\log\widehat{\rho}_i$ is displayed together with a scatter plot of $\log\widehat{g}_{1i}$ against $\log\widehat{g}_{2i}$. Exact proportionality of the fitted conditional variances implies

$$\log\widehat{g}_{1i} = c + \log\widehat{g}_{2i}$$

for a constant $c$, and therefore corresponds to a line with slope one.

## B.2 Approximate conditional diagonalization

Define the fitted structural residuals

$$\widehat{e}_{1i} = y_{1i} - \widehat{\gamma}_1 y_{2i} - f_1(x_i;\widehat{\nu}_1),$$

and

$$\widehat{e}_{2i} = y_{2i} - \widehat{\gamma}_2 y_{1i} - f_2(x_i;\widehat{\nu}_2).$$

The variance-standardized residual cross-product is

$$u_i = \frac{\widehat{e}_{1i}\widehat{e}_{2i}}{\sqrt{\widehat{g}_{1i}\widehat{g}_{2i}}}.$$

At the correctly specified structural parameter and nuisance functions, Assumption 2 (S2) implies

$$\mathrm{E}(u_i \mid x_i) = 0.$$

This conditional restriction implies

$$\mathrm{E}\{u_i a(x_i)\} = 0$$

for every scalar-valued integrable function $a$. We examine a finite collection of these implications using a prespecified vector of transformations

$$h(x_i) = \left(h_1(x_i), \ldots, h_{p_h}(x_i)\right)^\top,$$

whose first component is an intercept. Define

$$m_i^{(u)} = u_i h(x_i).$$

For a diagnostic sample of size $n_d$, let

$$\widehat{M}_u = \frac{1}{n_d}\sum_{i=1}^{n_d} m_i^{(u)}$$

and

$$\widehat{\Omega}_u = \frac{1}{n_d}\sum_{i=1}^{n_d}\left(m_i^{(u)} - \widehat{M}_u\right)\left(m_i^{(u)} - \widehat{M}_u\right)^\top.$$

We summarize the joint magnitude of the moments by

$$J_u = n_d \widehat{M}_u^\top \widehat{\Omega}_u^+ \widehat{M}_u,$$

where $\widehat{\Omega}_u^+$ denotes the Moore–Penrose inverse. We also report the largest absolute componentwise standardized moment,

$$T_{u,\max} = \max_\ell \left| \frac{\sqrt{n_d}\widehat{M}_{u,\ell}}{\sqrt{\widehat{\Omega}_{u,\ell\ell}}} \right|.$$

The maximum is taken over components satisfying $\widehat{\Omega}_{u,\ell\ell} > 0$.

The Moore–Penrose inverse is computed from the symmetric eigendecomposition

$$\widehat{\Omega}_u = V \operatorname{diag}\left(\widehat{\lambda}_1, \ldots, \widehat{\lambda}_{p_h}\right) V^\top, \qquad \widehat{\lambda}_1 \geq \cdots \geq \widehat{\lambda}_{p_h}.$$

Numerically negative eigenvalues are replaced by zero. For the relative tolerance $\tau = 10^{-10}$, define

$$\widehat{\Omega}_u^+ = V \operatorname{diag}\left(\frac{\mathbf{1}\{\widehat{\lambda}_j > \tau\widehat{\lambda}_1\}}{\widehat{\lambda}_j}\right)_{j=1}^{p_h} V^\top.$$

The corresponding effective rank is

$$\widehat{r}_u = \sum_{j=1}^{p_h} \mathbf{1}\{\widehat{\lambda}_j > \tau\widehat{\lambda}_1\}.$$

The componentwise summaries indicate the selected covariate directions along which residual dependence remains, whereas $J_u$ provides an omnibus standardized discrepancy over the retained moment space. Smaller values indicate closer agreement with the examined implications of conditional residual diagonalization.

## B.3 Approximate variance calibration of the structural residuals

For each equation, define the variance-standardized squared residual

$$\widehat{z}_{ki} = \frac{\widehat{e}_{ki}^2}{\widehat{g}_{ki}}, \qquad k = 1, 2.$$

Correct conditional variance specification implies

$$\mathrm{E}(\widehat{z}_{ki} \mid x_i) \approx 1$$

when the fitted structural parameter and nuisance functions are close to their population counterparts. The unconditional mean of $\widehat{z}_{ki}$ assesses overall variance scaling, but it cannot detect conditional calibration errors that offset across the covariate distribution.

Define

$$a_{ki} = \widehat{z}_{ki} - 1$$

and

$$m_{ki}^{(z)} = a_{ki} h(x_i).$$

Let

$$\widehat{M}_{z,k} = \frac{1}{n_d} \sum_{i=1}^{n_d} m_{ki}^{(z)}$$

and

$$\widehat{\Omega}_{z,k} = \frac{1}{n_d} \sum_{i=1}^{n_d} \left(m_{ki}^{(z)} - \widehat{M}_{z,k}\right)\left(m_{ki}^{(z)} - \widehat{M}_{z,k}\right)^\top.$$

Using the same Moore–Penrose convention as in Section B.2, define

$$J_{z,k} = n_d \widehat{M}_{z,k}^\top \widehat{\Omega}_{z,k}^+ \widehat{M}_{z,k}$$

and

$$T_{z,k,\max} = \max_{\ell:\widehat{\Omega}_{z,k,\ell\ell}>0} \frac{\sqrt{n_d}\,|\widehat{M}_{z,k,\ell}|}{\widehat{\Omega}_{z,k,\ell\ell}^{1/2}}.$$

We also report the empirical mean

$$\widehat{\mu}_{z,k} = \frac{1}{n_d}\sum_{i=1}^{n_d} \widehat{z}_{ki}.$$

A value of $\widehat{\mu}_{z,k}$ close to one indicates approximately correct unconditional variance scaling. The quantities $J_{z,k}$ and $T_{z,k,\max}$ detect calibration discrepancies associated with the selected transformations of the covariates.

## B.4 Feature vector for conditional-moment diagnostics

For the residual-dependence and variance-calibration diagnostics, we use a common, prespecified vector of transformations of the predetermined variables. Let $x_i^{\text{diag}}$ denote the numerical diagnostic-feature vector constructed from the raw predetermined information using transformations fitted exclusively on the training sample. It contains the standardized continuous variables, binary missingness indicators, one-hot encodings of the pooled categorical variables, and the fixed text singular value decomposition (SVD) components described in Appendix E. Learned neural-network embeddings are not included, because their values depend on the fitted outcome model and would make the diagnostic basis model-specific.

Let $n_{\text{train}}$ denote the training-sample size, and define the training mean

$$\bar{x}_{\text{train}} = \frac{1}{n_{\text{train}}} \sum_{i\in\mathcal{I}_{\text{train}}} x_i^{\text{diag}}.$$

The principal-component loading vectors are obtained from the centered training observations

$$x_i^{\text{diag}} - \bar{x}_{\text{train}}, \qquad i \in \mathcal{I}_{\text{train}}.$$

Let $v_1, \dots, v_q$ denote the loading vectors associated with the $q$ largest training-sample eigenvalues, ordered from largest to smallest. The corresponding unstandardized principal-component score is

$$s_{ij} = v_j^\top \left(x_i^{\text{diag}} - \bar{x}_{\text{train}}\right), \qquad j = 1, \dots, q.$$

For each component, let

$$\bar{s}_j = \frac{1}{n_{\text{train}}} \sum_{i\in\mathcal{I}_{\text{train}}} s_{ij}$$

and

$$\widehat{\sigma}_{s,j} = \left[\frac{1}{n_{\text{train}}-1} \sum_{i\in\mathcal{I}_{\text{train}}} (s_{ij} - \bar{s}_j)^2\right]^{1/2}.$$

The standardized principal-component score is

$$\text{PC}_{ij} = \frac{s_{ij} - \bar{s}_j}{\widehat{\sigma}_{s,j}}.$$

All centering constants, scaling constants, and principal-component loading vectors are estimated using the training sample only and are then held fixed when transforming validation and diagnostic observations.

Using these scores, define

$$c_j = \frac{1}{n_{\text{train}}} \sum_{i\in\mathcal{I}_{\text{train}}} \text{PC}_{ij}^2, \qquad j = 1, \dots, q,$$

and construct the diagnostic feature vector

$$h(x_i) = \left(1, \text{PC}_{i1}, \dots, \text{PC}_{iq}, \text{PC}_{i1}^2 - c_1, \dots, \text{PC}_{iq}^2 - c_q\right)^\top.$$

The linear principal-component terms detect first-order variation in the examined conditional moments, whereas the centered squared terms detect simple nonlinear variation along the same directions. Centering by $c_j$ makes each squared term have training-sample mean zero. Because sample standard deviation normalization is used, $c_j$

need not equal one exactly in finite samples; the computed training-sample value is retained. The feature dimension is

$$p_h = 1 + 2q.$$

Our default choice is $q = 8$, with $q = 4$ and $q = 16$ used in robustness analyses. The same transformation $h(x_i)$ is used to construct the residual-dependence diagnostic $J_u$ and the variance-calibration diagnostics $J_{z,k}$. Comparisons across different values of $q$ are interpreted cautiously because both the dimension and the effective rank of the examined moment space may change.

### B.5 Residual-gradient leverage

We next inspect residual-gradient leverage induced by inverse-variance weighting. For equation $k$, define the unnormalized residual-gradient weight

$$\widehat{\varpi}_{ki}^{(\beta)} = \frac{(\widehat{g}_{1i}\widehat{g}_{2i})^{\beta/2}}{\widehat{g}_{ki}}, \qquad k = 1, 2.$$

Apart from a common mini-batch normalization factor, this is the multiplicative variance-dependent weight appearing in the residual-driven mean-network gradient and in the residual component of the structural score. It does not equal the complete per-observation gradient norm, which also depends on the residual magnitude, neural-network Jacobians, the log-variance contribution, and the structural Jacobian contribution.

Because the weight is defined only up to a common multiplicative factor, we normalize it to have diagnostic-sample mean one:

$$\widetilde{\varpi}_{ki}^{(\beta)} = \frac{\widehat{\varpi}_{ki}^{(\beta)}}{n_d^{-1} \sum_{j=1}^{n_d} \widehat{\varpi}_{kj}^{(\beta)}}.$$

For each equation, we report the interquartile range, 95th percentile, and maximum of $\widetilde{\varpi}_{ki}^{(\beta)}$. This normalization makes the reported levels comparable across values of $\beta$ and across fitted specifications.

We also compute the normalized effective sample size

$$\mathrm{ESS}_{k,d}^{(\beta)} = \frac{1}{n_d} \frac{\left(\sum_{i=1}^{n_d} \widehat{\varpi}_{ki}^{(\beta)}\right)^2}{\sum_{i=1}^{n_d} \left(\widehat{\varpi}_{ki}^{(\beta)}\right)^2}.$$

The same value is obtained using $\widetilde{\varpi}_{ki}^{(\beta)}$. The normalized effective sample size lies in $[1/n_d, 1]$. A value close to one indicates relatively uniform variance-dependent residual-gradient weights, whereas a small value indicates concentration on a limited subset of observations. Comparing $\mathrm{ESS}_{k,d}^{(\beta)}$ and the upper tail of $\widetilde{\varpi}_{ki}^{(\beta)}$ across values of $\beta$ indicates whether stabilization reduces excessive leverage associated with very small fitted conditional variances.

## C Simulation details

### C.1 Data-generating process

This section describes the data-generating process used in the Monte Carlo experiments. We vary the sample size and signal-to-noise ratio while fixing the covariate dimension at $J = 100$:

$$n \in \{5000, 10000, 20000\}, \qquad R \in \{4, 1\}.$$

For each design point $(n, R)$, we generate 100 independent Monte Carlo replications. The design therefore evaluates the estimator under a fixed, nonlinear, high-dimensional data-generating process while varying sample size and signal strength. The primary estimands are the structural interaction parameters $\gamma_1$ and $\gamma_2$.

**Structural system.** For each observation $i = 1, \dots, n$, the outcomes are generated from the simultaneous structural system

$$y_{1i} = \gamma_1 y_{2i} + f_1(x_i) + \varepsilon_{1i},$$
$$y_{2i} = \gamma_2 y_{1i} + f_2(x_i) + \varepsilon_{2i}.$$

The true bidirectional interaction parameters are fixed at $\gamma_1 = 0.5$ and $\gamma_2 = 0.4$. Therefore,

$$1 - \gamma_1\gamma_2 = 0.8,$$

so that the simultaneous system is stable and bounded away from singularity.

Equivalently, the observed outcomes are obtained by solving the simultaneous system:

$$y_{1i} = \frac{f_1(x_i) + \gamma_1 f_2(x_i) + \varepsilon_{1i} + \gamma_1\varepsilon_{2i}}{1 - \gamma_1\gamma_2},$$

$$y_{2i} = \frac{\gamma_2 f_1(x_i) + f_2(x_i) + \gamma_2\varepsilon_{1i} + \varepsilon_{2i}}{1 - \gamma_1\gamma_2}.$$

**Covariates.** The covariate vector is

$$x_i = (x_{i1}, \dots, x_{iJ})^\top.$$

It is generated from a Gaussian copula so that the coordinates are correlated but each marginal variable lies in $(0, 1)$. Let

$$\breve{x}_i = (\breve{x}_{i1}, \dots, \breve{x}_{iJ})^\top$$

denote a latent Gaussian vector with Toeplitz dependence

$$(\Sigma_x)_{jj'} = \varrho_x^{|j-j'|}.$$

We set $\varrho_x = 0.5$. The latent vector is generated recursively as $\breve{x}_{i1} \sim N(0, 1)$, and, for $j = 2, \dots, J$,

$$\breve{x}_{ij} = \varrho_x \breve{x}_{i,j-1} + \sqrt{1 - \varrho_x^2}\xi_{ij}, \qquad \xi_{ij} \sim N(0, 1).$$

The innovations $\xi_{ij}$ are independent across $i$ and $j$. The observed covariates are then defined by $x_{ij} = \Phi(\breve{x}_{ij})$, where $\Phi(\cdot)$ denotes the standard normal cumulative distribution function. No further standardization is applied; the transformed covariates are clipped to $[10^{-6}, 1 - 10^{-6}]$ to avoid boundary singularities.

**Functional forms.** The functional forms are taken from [1]. The unscaled structural mean functions are

$$\begin{aligned}\widetilde{f}_1(x) =& \cot\left(\frac{\pi}{1 + \exp(x_1^2 + 2x_2 + \sin(6x_4^3) - 3)}\right) \\ &+ \exp(3x_3 + 2x_4 - 5x_5 + \sqrt{x_6 + 0.9x_7 + 0.1})\end{aligned}$$

and

$$\widetilde{f}_2(x) = \frac{2}{x_1 + 0.008} + 3\log(x_2^7 x_3 + 0.1)x_4.$$

For each replication, the final mean functions are

$$f_k(x_i) = s_k \widetilde{f}_k(x_i), \qquad k = 1, 2.$$

The conditional variance functions are

$$g_k(x) = g_{\min} + \psi\{\widetilde{g}_k(x)\}, \qquad k = 1, 2,$$

where $g_{\min} = 0.01$ and

$$\psi(t) = \log(1 + \exp(t))$$

is the softplus link. The variance-index functions are

$$\begin{aligned}\widetilde{g}_1(x) =& 2\log(x_1x_2 + 4x_3 + |\tan(x_4)| + 0.1) \\ &+ x_3^4x_5^2x_6 - x_4x_7 + (3x_8^2 + x_9 + 2)^{0.1+4x_{10}^2}\end{aligned}$$

and

$$\begin{aligned}\widetilde{g}_2(x) =& x_1 + \tan(x_2) + x_3^3 + \log(x_4 + 0.1) + 3x_5 \\ &+ x_6 + \sqrt{x_7 + 0.1}.\end{aligned}$$

**Signal-to-noise scaling.** For each replication and each equation, the scale of the structural mean function is chosen so that the empirical signal-to-noise ratio equals $R \in \{4, 1\}$ by construction. Let

$$\widehat{\mathrm{E}}_n(a_i) = \frac{1}{n}\sum_{i=1}^{n} a_i, \qquad \widehat{\mathrm{Var}}_n(a_i) = \frac{1}{n}\sum_{i=1}^{n}\left\{a_i - \widehat{\mathrm{E}}_n(a_i)\right\}^2.$$

After generating the covariates and computing $g_k(x_i)$, the scale parameter for equation $k$ is set to

$$s_k = \left[\frac{R\widehat{\mathrm{E}}_n(g_k(x_i))}{\widehat{\mathrm{Var}}_n(\widetilde{f}_k(x_i))}\right]^{1/2}, \qquad k = 1, 2.$$

Therefore,

$$\frac{\widehat{\mathrm{Var}}_n(f_k(x_i))}{\widehat{\mathrm{E}}_n(g_k(x_i))} = \frac{\widehat{\mathrm{Var}}_n(s_k\widetilde{f}_k(x_i))}{\widehat{\mathrm{E}}_n(g_k(x_i))} = R.$$

The scaling is computed using the full simulated sample within each replication.

**Structural shocks.** Conditional on $x_i$, the structural shocks are generated as

$$\varepsilon_{ki} = \sqrt{g_k(x_i)}u_{ki}, \qquad k = 1, 2.$$

The standardized shocks $u_{ki}$ are independent across observations and equations. They are generated from centered and scaled chi-squared random variables:

$$u_{ki} = \frac{v_{ki} - 10}{\sqrt{20}}, \qquad v_{ki} \sim \chi^2(10), \qquad k = 1, 2.$$

Thus, $\mathrm{E}(u_{ki}) = 0$ and $\mathrm{Var}(u_{ki}) = 1$. It follows that, for $k = 1, 2$,

$$\mathrm{E}(\varepsilon_{ki} \mid x_i) = 0,$$

and

$$\mathrm{Var}(\varepsilon_{ki} \mid x_i) = g_k(x_i).$$

Because the shocks are independent across equations,

$$\mathrm{Cov}(\varepsilon_{1i}, \varepsilon_{2i} \mid x_i) = 0.$$

The shocks are non-Gaussian, so the diagonal Gaussian objective used by the estimators is a quasi-log-likelihood rather than a correctly specified likelihood.

**Summary of one replication.** For each design point $(n, R)$, one Monte Carlo replication is generated as follows.

1. Generate $x_i \in (0, 1)^J$, $i = 1, \ldots, n$, from the Gaussian copula with $\varrho_x = 0.5$.
2. Compute the variance functions $g_1(x_i)$ and $g_2(x_i)$.
3. Compute the unscaled mean functions $\widetilde{f}_1(x_i)$ and $\widetilde{f}_2(x_i)$.
4. Choose $s_1$ and $s_2$ so that the empirical signal-to-noise ratio in each structural equation equals $R$.
5. Set $f_k(x_i) = s_k\widetilde{f}_k(x_i)$, $k = 1, 2$.
6. Generate standardized shocks $u_{1i}$ and $u_{2i}$ from centered and scaled $\chi^2(10)$ variables.
7. Set $\varepsilon_{ki} = \sqrt{g_k(x_i)}u_{ki}$, $k = 1, 2$.
8. Generate outcomes by solving the simultaneous system with $\gamma_1 = 0.5$ and $\gamma_2 = 0.4$.

## C.2 Hyperparameter optimization and implementation

This appendix describes the implementation details used for the Monte Carlo experiments. Unless otherwise stated, all SEM-DNN specifications using $\beta$-weighted negative-log-likelihood ($\beta$-NLL) stabilization use the same data split, neural architecture, optimization budget, validation criterion, and regularization-search protocol.

**Training-validation split.** For each Monte Carlo replication, the sample is randomly split into a training set and a validation set. The training set contains 80% of the observations, and the validation set contains the remaining 20%. The same training-validation split is used across all estimators within a replication. Parameter updates are computed using the training set, while early stopping and hyperparameter selection use the validation set.

**Network architecture.** For the neural simultaneous-equation estimators, each of the four neural networks $f_1$, $f_2$, $\widetilde{g}_1$, and $\widetilde{g}_2$ has two hidden layers with 256 units per layer. Gaussian error linear unit (GELU) activation functions are used in all hidden layers. The output layer of each mean network is linear. The output layer of each variance-index network is also linear, and the final conditional variance is obtained through

$$g_k(x_i; \alpha_k) = g_{\min} + \psi\{\widetilde{g}_k(x_i; \alpha_k)\}, \qquad k = 1, 2,$$

with $g_{\min} = 0.01$.

**Original penalized training objective.** For notational brevity, write $g_{ki}(\alpha_k) = g_k(x_i; \alpha_k)$. Let $q_i(\theta)$ denote the per-observation negative diagonal Gaussian quasi-log-likelihood, ignoring constants that do not depend on the parameters:

$$q_i(\theta) = -\log|1 - \gamma_1\gamma_2| + \frac{1}{2}\sum_{k=1}^{2}\left\{\log g_{ki}(\alpha_k) + \frac{e_{ki}(\gamma, \nu_k)^2}{g_{ki}(\alpha_k)}\right\},$$

where

$$e_{1i}(\gamma, \nu_1) = y_{1i} - \gamma_1 y_{2i} - f_1(x_i; \nu_1), \qquad e_{2i}(\gamma, \nu_2) = y_{2i} - \gamma_2 y_{1i} - f_2(x_i; \nu_2).$$

For a training mini-batch $\mathcal{B} \subset \mathcal{I}_{\text{train}}$, the ordinary SEM-DNN objective is

$$\mathcal{L}^{(0)}_{\text{train}}(\theta; \phi) = \frac{1}{|\mathcal{B}|}\sum_{i\in\mathcal{B}} q_i(\theta) + \mathcal{R}(\theta; \phi),$$

where

$$\theta = (\gamma^\top, \alpha_1^\top, \alpha_2^\top, \nu_1^\top, \nu_2^\top)^\top$$

and

$$\phi = (\phi_{\nu_1}, \phi_{\nu_2}, \phi_{\alpha_1}, \phi_{\alpha_2})^\top.$$

The regularization term is

$$\mathcal{R}(\theta; \phi) = \sum_{k=1}^{2}\{\phi_{\nu_k}\varsigma(\nu_k) + \phi_{\alpha_k}\varsigma(\alpha_k)\},$$

with

$$\varsigma(\vartheta) = \|\vartheta \odot \iota_\vartheta\|_2^2.$$

Here $\iota_\vartheta$ is a mask vector selecting network weights and $\odot$ denotes elementwise multiplication. Bias parameters are not penalized. The structural coefficients $\gamma_1$ and $\gamma_2$ are also not penalized.

**$\beta$-NLL stabilization rule.** Let $\text{sg}[\cdot]$ denote the stop-gradient operator. For a training mini-batch $\mathcal{B}$, define

$$s_i(\alpha_1, \alpha_2) = \{g_{1i}(\alpha_1)g_{2i}(\alpha_2)\}^{1/2}, \qquad \bar{s}_{\mathcal{B}} = \frac{1}{|\mathcal{B}|}\sum_{j\in\mathcal{B}} s_j(\alpha_1, \alpha_2),$$

and

$$w^{(\beta)}_{i,\mathcal{B}} = \text{sg}\left[\left\{\frac{s_i(\alpha_1, \alpha_2)}{\bar{s}_{\mathcal{B}}}\right\}^\beta\right].$$

The stabilized training objective is

$$\mathcal{L}^{(\beta)}_{\text{train}}(\theta; \phi) = \frac{1}{|\mathcal{B}|}\sum_{i\in\mathcal{B}} w^{(\beta)}_{i,\mathcal{B}} q_i(\theta) + \mathcal{R}(\theta; \phi).$$

The unstabilized SEM-DNN rule corresponds to $\beta = 0$. The main stabilized specifications use $\beta = 0.5$, fixed before validation. In the $\beta$-sensitivity ablation, we repeat the same protocol for

$$\beta \in \{0, 0.25, 0.5, 0.75, 1\}.$$

**Validation objective.** All simultaneous-equation estimators are selected using the same original unpenalized VNQLL,

$$\mathcal{L}^{(0)}_{\text{val}}(\theta) = \frac{1}{|\mathcal{B}_{\text{val}}|} \sum_{i \in \mathcal{B}_{\text{val}}} q_i(\theta).$$

The $\beta$-weighted objective is used only to define training updates. It is not used for early stopping, hyperparameter selection, or model comparison.

**Stability parameterization.** The structural coefficients are constrained to satisfy

$$|\gamma_k| \le \bar{\gamma} < 1, \qquad k = 1, 2.$$

In all main simulations, we set $\bar{\gamma} = 0.99$. The constraint is imposed by optimizing unconstrained parameters $\widetilde{\gamma}_1, \widetilde{\gamma}_2 \in \mathbb{R}$ and setting

$$\gamma_k = \bar{\gamma} \tanh(\widetilde{\gamma}_k), \qquad k = 1, 2.$$

This implies

$$1 - \gamma_1\gamma_2 \ge 1 - \bar{\gamma}^2 = 1 - 0.99^2 = 0.0199.$$

**Optimizer and gradient stabilization.** All neural simultaneous-equation estimators are trained using Adam [5]. The Adam learning rate is $\alpha_{\text{Adam}} = 0.001$ with momentum parameters $\beta_{\text{Adam},1} = 0.9$, $\beta_{\text{Adam},2} = 0.999$, and numerical stability constant $\epsilon_{\text{Adam}} = 10^{-8}$. Adaptive gradient clipping is applied as a numerical stabilization device with a clipping threshold of 0.01 [2]. The clipping threshold is fixed in advance and is not selected by validation.

**Mini-batches and early stopping.** The mini-batch size is fixed at 4000. The maximum number of epochs is 2000. Early stopping is applied using the original validation criterion $\mathcal{L}^{(0)}_{\text{val}}(\theta)$. Training is terminated when $\mathcal{L}^{(0)}_{\text{val}}(\theta)$ fails to improve for 50 consecutive epochs. The reported parameter estimate is taken from the epoch with the smallest $\mathcal{L}^{(0)}_{\text{val}}(\theta)$, not from the final epoch and not from the epoch minimizing a $\beta$-weighted training objective. In the main experiments, $\beta$ is fixed at 0.5. Therefore the quasi-random search is over regularization coefficients only. In the $\beta$-sensitivity ablation, the same quasi-random regularization protocol is repeated separately for each fixed value of $\beta$.

**Regularization search space.** The regularization coefficients are constrained to be positive by optimizing their base-10 logarithms. In the main four-dimensional specification, let

$$\lambda = (\lambda_{\nu 1}, \lambda_{\nu 2}, \lambda_{\alpha 1}, \lambda_{\alpha 2})^\top, \qquad \lambda_j = \log_{10} \phi_j,$$

with each component searched over $[-8, -1]$. The main untied SEM-DNN, SEM-PAB, and SEM-Kernel specifications use this four-dimensional search. The tied-regularization, no-variance-regularization, and no-mean-regularization ablations reduce the search to two log-penalty components, each over the same interval.

Some ablation specifications reduce the number of searched penalty parameters to two. In the tied-regularization ablation, the penalties are constrained as

$$\phi_\nu = \phi_{\nu 1} = \phi_{\nu 2}, \qquad \phi_\alpha = \phi_{\alpha 1} = \phi_{\alpha 2},$$

so that the search variables are

$$\lambda_\nu = \log_{10} \phi_\nu, \qquad \lambda_\alpha = \log_{10} \phi_\alpha.$$

In the no-variance-regularization ablation, $\phi_{\alpha 1} = \phi_{\alpha 2} = 0$, and only $\lambda_{\nu 1}$ and $\lambda_{\nu 2}$ are searched. In the no-mean-regularization ablation, $\phi_{\nu 1} = \phi_{\nu 2} = 0$, and only $\lambda_{\alpha 1}$ and $\lambda_{\alpha 2}$ are searched. All searched log-penalty components in these two-dimensional specifications are again restricted to $[-8, -1]$.

**Quasi-random hyperparameter optimization and final training.** Regularization coefficients are selected by validation-based quasi-random search. Candidate configurations are generated using a scrambled Sobol sequence on the unit hypercube and then linearly transformed to the relevant log-penalty search space. The scrambling seed is set as a deterministic function of the design point, replication, and estimator, so that the hyperparameter search is reproducible while remaining independent across replications. The Sobol design is used because it gives more uniform coverage of the low-dimensional regularization space than ordinary pseudo-random search for a fixed number of candidate configurations.

The number of first-stage Sobol candidates depends on the dimension of the regularization search. Four-dimensional SEM specifications use 32 first-stage candidates, whereas two-dimensional SEM specifications use 20 first-stage candidates. Thus, the main untied SEM-DNN, SEM-PAB, and SEM-Kernel specifications use 32 first-stage candidates. The two-dimensional SEM ablations, including no-variance regularization, no-mean regularization, and tied regularization, use 20 first-stage candidates.

The staged continuation protocol is otherwise unchanged. In the first stage, all Sobol configurations are trained from initialization with a maximum budget of 30 epochs, and the 8 configurations with the most favorable VNQLL value are promoted. For each promoted configuration, the saved weights and optimizer state are restored from the epoch within that stage at which the validation criterion was minimized, not necessarily from the terminal epoch of the stage. In the second stage, the promoted configurations are continued from their restored stage-best checkpoints and trained through at most epoch 100, and the best 3 configurations are promoted using the same validation-based rule. In the third stage, these 3 configurations are continued from their restored stage-best checkpoints with a budget up to epoch 200, and the best configuration is selected. The selected configuration is then restored to its best validation checkpoint from the third stage and continued until at most epoch 2000. The reported estimate is taken from the checkpoint that attains the smallest original validation criterion $\mathcal{L}_{\text{val}}^{(0)}(\theta)$ over the entire continuation path.

For the Single-DNN benchmark, hyperparameter optimization is conducted separately for the two equations. Each equation uses a one-dimensional log-penalty search over

$$\log_{10} \phi_\nu \in [-8, -1],$$

with 20 first-stage Sobol candidates and the same staged continuation budgets, promotion counts, validation-based early stopping rule, and final restoration rule.

**Initialization of structural coefficients.** For each training run, the initial structural coefficients are sampled as $\gamma_k^{(0)} \sim U(-0.9, 0.9)$, $k = 1, 2$. They are then mapped to the unconstrained parameters by

$$\tilde{\gamma}_k^{(0)} = \operatorname{arctanh}\left(\frac{\gamma_k^{(0)}}{\bar{\gamma}}\right), \qquad k = 1, 2.$$

**Initialization of mean networks.** After drawing the initial structural coefficients $\gamma_1^{(0)}$ and $\gamma_2^{(0)}$, let $n_{\text{train}} = |\mathcal{I}_{\text{train}}|$ and define the transformed outcomes

$$z_{1i}^{(0)} = y_{1i} - \gamma_1^{(0)} y_{2i}, \qquad z_{2i}^{(0)} = y_{2i} - \gamma_2^{(0)} y_{1i}.$$

For each equation $k = 1, 2$, let

$$\bar{z}_k^{(0)} = \frac{1}{|\mathcal{I}_{\text{train}}|} \sum_{i \in \mathcal{I}_{\text{train}}} z_{ki}^{(0)}.$$

The hidden-layer parameters of the mean network are initialized using the standard random initialization. The final-layer weights are initialized at zero, and the final-layer bias is set equal to $\bar{z}_k^{(0)}$. Consequently, the initial fitted mean is the constant function $f_k^{(0)}(x) = \bar{z}_k^{(0)}$, so that the preliminary residuals are centered in the training sample before initializing the variance networks.

**Initialization of variance networks.** The variance-index networks are initialized so that the initial fitted variances match the empirical unconditional residual variances in the training sample. Let

$$r_{1i}^{(0)} = y_{1i} - \gamma_1^{(0)} y_{2i} - f_1^{(0)}(x_i),$$

and

$$r_{2i}^{(0)} = y_{2i} - \gamma_2^{(0)} y_{1i} - f_2^{(0)}(x_i)$$

denote the preliminary residuals from the initialized mean networks. For equation $k$, define

$$v_k^{(0)} = \frac{1}{n_{\text{train}}} \sum_{i \in \mathcal{I}_{\text{train}}} \left( r_{ki}^{(0)} \right)^2,$$

where $n_{\text{train}}$ is the training-sample size.

The final-layer weights of the variance-index network are initialized at zero. The final-layer bias is set so that

$$g_{\min} + \psi(b_k) = v_k^{(0)}.$$

Thus,

$$b_k = \log\{\exp(v_k^{(0)} - g_{\min}) - 1\}.$$

For numerical stability, $v_k^{(0)}$ is truncated below at $g_{\min} + \epsilon$, where $\epsilon > 0$ is a small constant. Consequently, the initial fitted variance is constant across observations and equal to the empirical unconditional residual variance, while subsequent optimization allows the variance network to learn heteroscedastic variation.

**Implementation of regularization and $\beta$-NLL ablations.** The $\beta$-sensitivity ablation repeats the main four-dimensional neural simultaneous-equation training protocol for

$$\beta \in \{0, 0.25, 0.5, 0.75, 1\}.$$

For $\beta = 0$, the stabilized objective reduces to ordinary quasi-log-likelihood training. Because all four regularization coefficients remain freely selected in this ablation, each value of $\beta$ uses 32 first-stage Sobol candidates. In every case, early stopping, hyperparameter selection, final training, and reporting use the original validation criterion $\mathcal{L}_{\text{val}}^{(0)}(\theta)$.

The no-variance-regularization ablation sets

$$\phi_{\alpha 1} = \phi_{\alpha 2} = 0$$

and selects only the two mean-network penalties $\phi_{\nu 1}$ and $\phi_{\nu 2}$ by validation. The no-mean-regularization ablation sets

$$\phi_{\nu 1} = \phi_{\nu 2} = 0$$

and selects only the two variance-network penalties $\phi_{\alpha 1}$ and $\phi_{\alpha 2}$ by validation. These two ablations therefore use the two-dimensional Sobol search protocol with 20 first-stage candidates. The structural coefficients $\gamma_1$ and $\gamma_2$ remain unpenalized, and the same validation criterion, early-stopping rule, optimizer, mini-batch size, and initialization scheme are used as in the main SEM-DNN specification.

**Random seeds and comparability across estimators.** Within each Monte Carlo replication, the same simulated dataset and the same training-validation split are used across estimators. Random seeds for network initialization and mini-batch ordering are fixed as deterministic functions of the replication index and the quasi-random search trial index. This makes comparisons across regularization specifications and estimators less sensitive to arbitrary initialization differences.

**Software and hardware.** All programs are written in Python and PyTorch. The experiments are executed under Ubuntu on a workstation with an AMD Ryzen 7 5700G CPU and an NVIDIA GeForce RTX 5060 Ti GPU.

## C.3 Alternative estimators

This section gives the full specifications of the three alternative estimators used in the Monte Carlo study: SEM-PAB, SEM-Kernel, and Single-DNN. SEM-PAB and SEM-Kernel retain the simultaneous-equation structure and heteroscedastic quasi-log-likelihood of SEM-DNN but replace the neural networks with alternative function approximators. Single-DNN instead ignores simultaneity and estimates the two equations separately. It is included to illustrate the consequences of treating a simultaneous causal system as two ordinary supervised-learning problems. SEM-PAB and SEM-Kernel also use the same SEM-$\beta$-NLL stabilization rule, so they serve as benchmarks for the same identification strategy under different nuisance-function approximations.

**SEM-PAB: simultaneous equations with polynomial approximation bases.** SEM-PAB is a parametric heteroscedastic quasi-log-likelihood estimator with the same structural system as SEM-DNN:

$$y_{1i} = \gamma_1 y_{2i} + f_1(x_i; \nu_1) + \varepsilon_{1i},$$
$$y_{2i} = \gamma_2 y_{1i} + f_2(x_i; \nu_2) + \varepsilon_{2i}.$$

It also uses the same diagonal Gaussian quasi-log-likelihood,

$$\ell_i(\theta) = \log|\det\Gamma(\gamma)| - \log(2\pi) - \frac{1}{2}\left\{\log(g_{1i}g_{2i}) + \frac{e_{1i}^2}{g_{1i}} + \frac{e_{2i}^2}{g_{2i}}\right\},$$

where

$$e_{1i} = y_{1i} - \gamma_1 y_{2i} - f_1(x_i; \nu_1), \qquad e_{2i} = y_{2i} - \gamma_2 y_{1i} - f_2(x_i; \nu_2).$$

The difference from SEM-DNN is that the mean and variance-index functions are represented using beta-function-weighted polynomial approximation bases. This specification follows the flexible approximation strategy used in [6]. In the original version, this approximation is used for the variance functions; here we use the same type of approximation for both the mean and variance-index functions.

For $r = 1, \ldots, 4$, define normalized beta weights

$$b_{kjr}(a,b) = \frac{\left(\frac{r}{5}\right)^{\psi(a)-1}\left(1-\frac{r}{5}\right)^{\psi(b)-1}}{\sum_{r'=1}^{4}\left(\frac{r'}{5}\right)^{\psi(a)-1}\left(1-\frac{r'}{5}\right)^{\psi(b)-1}},$$

where

$$\psi(t) = \log\{1+\exp(t)\}$$

is the softplus link. The softplus transformation ensures positive beta-shape parameters. The variance-index functions are specified as follows:

$$g_k(x_i; \alpha_k) = g_{\min} + \psi\left(\sum_{j=1}^{J}\widetilde{g}_{kij}\right), \qquad k = 1, 2,$$

where

$$\widetilde{g}_{kij} = \left\{\psi(\alpha_{kj0}) + \psi(\alpha_{kj1})\sum_{r=1}^{4} b_{kjr}(\alpha_{kj2}, \alpha_{kj3})x_{ij}^r\right\}^2.$$

For each $j$,

$$\alpha_{kj} = (\alpha_{kj0}, \alpha_{kj1}, \alpha_{kj2}, \alpha_{kj3})^\top,$$

and

$$\alpha_k = (\alpha_{k1}^\top, \ldots, \alpha_{kJ}^\top)^\top.$$

This functional form ensures that $\widetilde{g}_{kij} > 0$. The variance lower bound is the same as in SEM-DNN, $g_{\min} = 0.01$. The mean functions are specified analogously, but they are allowed to be negative,

$$f_k(x_i; \nu_k) = \sum_{j=1}^{J}\widetilde{f}_{kij}, \qquad k = 1, 2,$$

where

$$\widetilde{f}_{kij} = \nu_{kj0} + \nu_{kj1}\sum_{r=1}^{4} b_{kjr}(\nu_{kj2}, \nu_{kj3})x_{ij}^r.$$

For each $j$,

$$\nu_{kj} = (\nu_{kj0}, \nu_{kj1}, \nu_{kj2}, \nu_{kj3})^\top,$$

and

$$\nu_k = (\nu_{k1}^\top, \ldots, \nu_{kJ}^\top)^\top.$$

SEM-PAB is estimated by minimizing the regularized negative quasi-log-likelihood:

$$\mathcal{L}_{\text{PAB}}(\theta; \phi) = -\frac{1}{n_{\text{train}}}\sum_{i\in\mathcal{I}_{\text{train}}}\ell_i(\theta) + \sum_{k=1}^{2}\left(\phi_{\nu k}\|\nu_k\|_2^2 + \phi_{\alpha k}\|\alpha_k\|_2^2\right).$$

The structural coefficients $\gamma_1$ and $\gamma_2$ are not penalized. Optimization uses Adam with full-batch updates and $\beta$-NLL stabilization with $\beta = 0.5$; regularization tuning, validation, and early stopping follow the same validation protocol as SEM-DNN whenever applicable. The same training-validation split, VNQLL, and regularization-selection protocol are used. SEM-PAB uses the same stability parameterization and constraint on $\gamma$ as SEM-DNN.

**SEM-Kernel: simultaneous equations with random Fourier features.** SEM-Kernel retains the simultaneous-equation structure and the heteroscedastic quasi-log-likelihood objective. It represents the unknown functions using random Fourier features for a radial basis function kernel [7].

The radial basis function kernel is

$$K(x,x') = \exp\left(-\frac{1}{2\tau}\|x-x'\|_2^2\right),$$

where $\tau > 0$ is the bandwidth parameter. The bandwidth is chosen using the median heuristic [3, 4]. Specifically, for an index set $\mathcal{I}$ sampled without replacement from the full sample, with $|\mathcal{I}| = 1000$, we set

$$\tau = \text{median}\left\{\|x_i - x_{i'}\|_2^2 : i \neq i',\ i, i' \in \mathcal{I}\right\}.$$

The random Fourier feature dimension is fixed at $D = 500$. For $d = 1, \ldots, D$, draw

$$\omega_d \sim N(0, \tau^{-1} I_J), \qquad \upsilon_d \sim U(0, 2\pi).$$

The random feature map is

$$z(x_i) = \sqrt{\frac{2}{D}}\left(\cos(\omega_1^\top x_i + \upsilon_1), \ldots, \cos(\omega_D^\top x_i + \upsilon_D)\right)^\top.$$

The auxiliary random variables $\{(\omega_d, \upsilon_d)\}_{d=1}^D$ are regenerated for each Monte Carlo replication and fixed during hyperparameter tuning within that replication. The same random feature map is used for the mean and variance-index functions.

For $k = 1, 2$, the mean functions are

$$f_k(x_i; \nu_k) = \nu_k^\top z(x_i),$$

and the variance functions are

$$g_k(x_i; \alpha_k) = g_{\min} + \psi\{\widetilde{g}_k(x_i; \alpha_k)\},$$

with

$$\widetilde{g}_k(x_i; \alpha_k) = \alpha_k^\top z(x_i).$$

Here again

$$\psi(t) = \log\{1 + \exp(t)\}$$

and $g_{\min} = 0.01$.

SEM-Kernel is estimated by minimizing the regularized negative quasi-log-likelihood:

$$\mathcal{L}_{\text{Kernel}}(\theta; \phi) = -\frac{1}{n_{\text{train}}} \sum_{i \in \mathcal{I}_{\text{train}}} \ell_i(\theta) + \sum_{k=1}^{2} \left(\phi_{\nu k}\|\nu_k\|_2^2 + \phi_{\alpha k}\|\alpha_k\|_2^2\right).$$

For optimization, we use Adam with full-batch updates and $\beta$-NLL stabilization with $\beta = 0.5$. We use the same validation-based regularization-selection and early-stopping protocol as in SEM-DNN. The same stability parameterization and constraint on $\gamma$ as SEM-DNN are used.

**Single-DNN: separate neural regressions ignoring simultaneity.** Single-DNN estimates the two equations separately and treats the outcome on the right-hand side as an ordinary predictor. It does not model conditional heteroscedasticity and does not use the simultaneous-equation likelihood. This estimator is included to quantify the consequences of ignoring the bidirectional structural system.

For the first equation, Single-DNN solves

$$\min_{\gamma_1, \nu_1} \frac{1}{n_{\text{train}}} \sum_{i \in \mathcal{I}_{\text{train}}} \{y_{1i} - \gamma_1 y_{2i} - f_1(x_i; \nu_1)\}^2 + \phi_\nu \varsigma(\nu_1).$$

For the second equation, it solves

$$\min_{\gamma_2, \nu_2} \frac{1}{n_{\text{train}}} \sum_{i \in \mathcal{I}_{\text{train}}} \{y_{2i} - \gamma_2 y_{1i} - f_2(x_i; \nu_2)\}^2 + \phi_\nu \varsigma(\nu_2).$$

The two optimization problems are independent. The mean networks $f_1$ and $f_2$ use the same architecture as the SEM-DNN mean networks: two hidden layers with 256 units per layer and GELU activations. The slope coefficients $\gamma_1$ and $\gamma_2$ are unconstrained in Single-DNN.

Table D.1: Results for alternative variance structure

| Measure | $R$ | Param | $c$ | | |
|---|---|---|---|---|---|
| | | | 0 | 0.5 | 1 |
| Mean bias | 4 | $\gamma_1$ | -0.478 | -0.164 | -0.085 |
| | | $\gamma_2$ | 0.011 | 0.013 | -0.063 |
| | 1 | $\gamma_1$ | -0.112 | -0.100 | -0.052 |
| | | $\gamma_2$ | 0.005 | 0.014 | -0.015 |
| RMSE | 4 | $\gamma_1$ | 0.679 | 0.319 | 0.235 |
| | | $\gamma_2$ | 0.018 | 0.160 | 0.345 |
| | 1 | $\gamma_1$ | 0.320 | 0.293 | 0.237 |
| | | $\gamma_2$ | 0.016 | 0.157 | 0.248 |
| Standard deviation | 4 | $\gamma_1$ | 0.484 | 0.275 | 0.220 |
| | | $\gamma_2$ | 0.014 | 0.161 | 0.341 |
| | 1 | $\gamma_1$ | 0.301 | 0.277 | 0.232 |
| | | $\gamma_2$ | 0.016 | 0.157 | 0.248 |
| VNQLL | 4 | | 4.084 | 5.276 | 5.623 |
| | 1 | | 3.550 | 4.378 | 4.797 |
| Runtime (s) | 4 | | 29.856 | 33.564 | 33.668 |
| | 1 | | 30.260 | 34.998 | 34.329 |

The regularization coefficient $\phi_\nu$ is selected by quasi-random search using the same log-scale search range as in SEM-DNN:

$$\log_{10} \phi_\nu \in [-8, -1].$$

The validation criterion is mean squared error. Hyperparameter optimization is conducted separately for the two equations.

# D Ablation studies

## D.1 Alternative variance structure

We modify the data-generating process so that the two structural variance functions become increasingly correlated and, ultimately, proportional. The baseline design uses nonproportional conditional variances $g_1(x_i)$ and $g_2(x_i)$, which satisfy the identifying condition in Lemma 1. This ablation weakens that source of identifying variation while holding the remaining features of the design as close as possible to the baseline. Specifically, after computing the baseline variance functions $g_1(x_i)$ and $g_2(x_i)$, we define

$$g_1^{\text{prop}}(x_i) = g_1(x_i), \qquad g_2^{\text{prop}}(x_i) = (1-c)g_2(x_i) + cg_1(x_i), \tag{D.1}$$

where $c \in \{0, 0.5, 1\}$ controls the degree to which the second variance function is moved toward the first. When $c = 0$, the baseline nonproportional design is recovered. When $c = 1$, the two variance functions are identical, so the variance ratio is constant and the point-identification condition in Lemma 1 fails.

All other structural functional forms are held fixed; shocks are regenerated and signal-to-noise ratio normalizations are recomputed under the modified variances. While the standardized shocks $u_{ki}$ are kept unchanged, the structural shocks $\varepsilon_{ki}$ are regenerated using $g_1^{\text{prop}}(x_i)$ and $g_2^{\text{prop}}(x_i)$. The mean-function scaling constants $s_1$ and $s_2$ are then recomputed using the proportional variances so that the target empirical signal-to-noise ratio remains $R$ in each structural equation. For all ablation studies, including this one, we fix the sample size at $n = 10000$.

Table D.1 shows that making the conditional variances closer to proportional redistributes, rather than uniformly increases, estimation error across the two interaction coefficients. As $c$ increases, the root mean squared error (RMSE) of $\widehat{\gamma}_1$ decreases, whereas that of $\widehat{\gamma}_2$ rises sharply; under exact proportionality ($c = 1$), the latter reaches $0.345$ for $R = 4$ and $0.248$ for $R = 1$. Thus, when the variance ratio becomes constant and point identification fails, the fitted criterion may still estimate one direction relatively accurately, but it cannot reliably separate both structural effects, with the remaining ambiguity absorbed asymmetrically across the two coefficients.

## D.2 No variance-network regularization

We remove the $L_2$ penalty from the variance networks by setting

$$\phi_{\alpha 1} = \phi_{\alpha 2} = 0,$$

Table D.2: Results for no variance-network regularization

| Measure | $R$ | Param | Specification | |
|---|---|---|---|---|
| | | | Regularized | Not regularized |
| Mean bias | 4 | $\gamma_1$ | -0.478 | -0.252 |
| | | $\gamma_2$ | 0.011 | 0.009 |
| | 1 | $\gamma_1$ | -0.112 | -0.090 |
| | | $\gamma_2$ | 0.005 | 0.004 |
| RMSE | 4 | $\gamma_1$ | 0.679 | 0.471 |
| | | $\gamma_2$ | 0.018 | 0.026 |
| | 1 | $\gamma_1$ | 0.320 | 0.310 |
| | | $\gamma_2$ | 0.016 | 0.019 |
| Standard deviation | 4 | $\gamma_1$ | 0.484 | 0.400 |
| | | $\gamma_2$ | 0.014 | 0.025 |
| | 1 | $\gamma_1$ | 0.301 | 0.298 |
| | | $\gamma_2$ | 0.016 | 0.018 |
| VNQLL | 4 | | 4.084 | 4.017 |
| | 1 | | 3.550 | 3.516 |
| Runtime (s) | 4 | | 29.856 | 28.233 |
| | 1 | | 30.260 | 28.149 |

while keeping the mean-network penalties selected by the same validation protocol as in the baseline SEM-DNN specification. This ablation tests whether regularization of $g_1$ and $g_2$ is important for structural estimation.

The variance networks enter the quasi-log-likelihood through both $\log g_k(x_i)$ and $e_{ki}^2/g_k(x_i)$. Without regularization, they may overfit idiosyncratic residual patterns and produce unstable or poorly calibrated variance estimates. Since the fitted variance functions provide the identifying heteroscedastic signal, variance overfitting can distort the residual diagonalization mechanism and thereby affect the point estimates of $\gamma$.

Table D.2 shows that removing variance-network regularization improves the estimation of $\gamma_1$, particularly when $R = 4$, reducing its RMSE from $0.679$ to $0.471$, while producing modest increases in the RMSE and empirical standard deviation of $\gamma_2$. The unregularized specification also yields slightly lower VNQLL values and shorter runtimes. Thus, variance-network regularization is not uniformly beneficial in this design: it stabilizes estimation of $\gamma_2$, but its shrinkage can impair recovery of the more difficult structural coefficient $\gamma_1$.

## D.3 No mean-network regularization

We remove the $L_2$ penalty from the mean networks by setting

$$\phi_{\nu 1} = \phi_{\nu 2} = 0,$$

while keeping the variance-network penalties selected by the same validation protocol as in the baseline SEM-DNN specification. This ablation examines whether overly flexible mean networks can damage structural estimation.

In the proposed model, the residuals are defined by

$$e_{1i} = y_{1i} - \gamma_1 y_{2i} - f_1(x_i), \qquad e_{2i} = y_{2i} - \gamma_2 y_{1i} - f_2(x_i).$$

If the mean networks are too flexible, they may absorb variation that should remain in the structural residuals. This can reduce or distort the heteroscedastic signal used to identify $\gamma$. Thus, this ablation evaluates the role of mean-network regularization in preserving a stable mean–variance decomposition.

Table D.3 indicates that removing mean-network regularization has mixed effects. For $R = 4$, it slightly reduces the RMSE of $\widehat{\gamma}_1$ from $0.679$ to $0.642$, but increases the dispersion of both coefficient estimates, particularly that of $\widehat{\gamma}_2$. For $R = 1$, the RMSE of $\widehat{\gamma}_1$ rises from $0.320$ to $0.391$, together with larger bias and empirical standard deviation. These results suggest that mean-network regularization helps preserve a stable mean–variance decomposition, although its finite-sample benefit depends on the signal-to-noise regime.

## D.4 Different $\beta$-NLL stabilization levels

We vary the stabilization exponent $\beta$ in the SEM-$\beta$-NLL training rule. We consider

$$\beta \in \{0, 0.25, 0.5, 0.75, 1\},$$

Table D.3: Results for no mean-network regularization

| Measure | $R$ | Param | Specification | |
|---|---|---|---|---|
| | | | Regularized | Not regularized |
| Mean bias | 4 | $\gamma_1$ | -0.478 | -0.415 |
| | | $\gamma_2$ | 0.011 | 0.012 |
| | 1 | $\gamma_1$ | -0.112 | -0.171 |
| | | $\gamma_2$ | 0.005 | 0.007 |
| RMSE | 4 | $\gamma_1$ | 0.679 | 0.642 |
| | | $\gamma_2$ | 0.018 | 0.033 |
| | 1 | $\gamma_1$ | 0.320 | 0.391 |
| | | $\gamma_2$ | 0.016 | 0.020 |
| Standard deviation | 4 | $\gamma_1$ | 0.484 | 0.492 |
| | | $\gamma_2$ | 0.014 | 0.031 |
| | 1 | $\gamma_1$ | 0.301 | 0.353 |
| | | $\gamma_2$ | 0.016 | 0.019 |
| VNQLL | 4 | | 4.084 | 4.090 |
| | 1 | | 3.550 | 3.523 |
| Runtime (s) | 4 | | 29.856 | 28.332 |
| | 1 | | 30.260 | 28.592 |

where $\beta = 0$ corresponds to ordinary quasi-log-likelihood training. All other aspects of the estimator, including the neural architecture, regularization search space, validation criterion, early stopping rule, and stability parameterization of $\gamma$, are kept unchanged.

This ablation evaluates whether the empirical performance of SEM-DNN depends on the degree of inverse-variance stabilization. Small values of $\beta$ remain close to the original quasi-log-likelihood but may not sufficiently reduce extreme inverse-variance gradient weights. Large values of $\beta$ reduce the influence of small fitted variances during training, but may also move the optimization path farther from the original quasi-log-likelihood.

The main experiments use $\beta = 0.5$ as a compromise between structural accuracy and inverse-variance stability. Since all candidate values of $\beta$ are evaluated using the original unpenalized VNQLL, this comparison does not favor a particular $\beta$-weighted objective by construction.

Table D.4 shows no monotonic relationship between the stabilization exponent $\beta$ and structural accuracy. For $R = 4$, $\beta = 0.25$ yields the lowest RMSEs for both coefficients, whereas for $R = 1$, the default $\beta = 0.5$ performs best for $\widehat{\gamma}_1$ and is tied with $\beta = 0.75$ for $\widehat{\gamma}_2$. Both ordinary quasi-likelihood training ($\beta = 0$) and full stabilization ($\beta = 1$) generally produce larger RMSEs and empirical dispersion. These results support moderate stabilization, with $\beta = 0.5$ providing a favorable overall compromise across signal-to-noise regimes while also requiring less computation in this experiment.

## D.5 Architecture sizes

The baseline SEM-DNN specification uses two hidden layers with 256 units per layer for each of the four networks $f_1, f_2, \widetilde{g}_1, \widetilde{g}_2$. To examine sensitivity to network capacity, we compare the baseline architecture with smaller and larger alternatives. The comparison covers the architectures $2 \times 64$, $2 \times 256$, $2 \times 512$, and $3 \times 256$, where $L \times H$ denotes $L$ hidden layers with $H$ hidden units per layer. All other settings are fixed, including the training rule, stability parameterization, early-stopping rule, hyperparameter-search procedure, and original VNQLL.

This ablation is motivated by the possibility that overparameterized mean and variance networks may induce nonmonotone behavior in structural estimation accuracy. In particular, larger networks can improve predictive fit while destabilizing the decomposition between conditional mean and conditional variance. Such behavior may appear as increased dispersion, bias, or RMSE of $\widehat{\gamma}_1$ and $\widehat{\gamma}_2$ even when the VNQLL improves.

The purpose of this experiment is diagnostic. We do not attempt a complete characterization of double-descent phenomena. Instead, we use the architecture sweep to check whether the baseline conclusions are robust to moderate changes in network capacity.

Table D.5 shows that structural accuracy generally improves with network capacity over the architectures considered. Relative to the baseline ($2 \times 256$) specification, the larger ($2 \times 512$) and deeper ($3 \times 256$) networks substantially reduce the bias, RMSE, and empirical dispersion of $\widehat{\gamma}_1$, while maintaining small errors for $\widehat{\gamma}_2$; the $3 \times 256$ architecture performs best overall. These gains are accompanied by lower VNQLL but longer runtimes. Thus, within this moderate architecture range, additional capacity improves rather than destabilizes structural estimation, although the baseline architecture remains a less costly specification.

Table D.4: Results for different $\beta$-NLL stabilization levels

| Measure | $R$ | Param | $\beta$ | | | | |
|---|---|---|---|---|---|---|---|
| | | | 0 | 0.25 | 0.5 | 0.75 | 1 |
| Mean bias | 4 | $\gamma_1$ | -0.488 | -0.442 | -0.478 | -0.434 | -0.455 |
| | | $\gamma_2$ | 0.012 | 0.010 | 0.011 | 0.011 | 0.011 |
| | 1 | $\gamma_1$ | -0.218 | -0.183 | -0.112 | -0.113 | -0.146 |
| | | $\gamma_2$ | 0.008 | 0.008 | 0.005 | 0.004 | 0.005 |
| RMSE | 4 | $\gamma_1$ | 0.692 | 0.637 | 0.679 | 0.650 | 0.698 |
| | | $\gamma_2$ | 0.022 | 0.016 | 0.018 | 0.019 | 0.022 |
| | 1 | $\gamma_1$ | 0.419 | 0.377 | 0.320 | 0.369 | 0.424 |
| | | $\gamma_2$ | 0.020 | 0.021 | 0.016 | 0.016 | 0.018 |
| Standard deviation | 4 | $\gamma_1$ | 0.493 | 0.460 | 0.484 | 0.485 | 0.532 |
| | | $\gamma_2$ | 0.019 | 0.013 | 0.014 | 0.015 | 0.019 |
| | 1 | $\gamma_1$ | 0.359 | 0.331 | 0.301 | 0.352 | 0.400 |
| | | $\gamma_2$ | 0.018 | 0.020 | 0.016 | 0.016 | 0.018 |
| VNQLL | 4 | | 4.157 | 4.059 | 4.084 | 4.095 | 4.113 |
| | 1 | | 3.607 | 3.561 | 3.550 | 3.536 | 3.623 |
| Runtime (s) | 4 | | 35.178 | 35.655 | 29.856 | 35.733 | 35.304 |
| | 1 | | 35.847 | 35.993 | 30.260 | 35.518 | 34.681 |

Table D.5: Results for different architecture sizes

| Measure | $R$ | Param | Architecture (layers $\times$ units) | | | |
|---|---|---|---|---|---|---|
| | | | $2 \times 64$ | $2 \times 256$ | $2 \times 512$ | $3 \times 256$ |
| Mean bias | 4 | $\gamma_1$ | -0.469 | -0.478 | -0.252 | -0.205 |
| | | $\gamma_2$ | 0.013 | 0.011 | 0.005 | 0.002 |
| | 1 | $\gamma_1$ | -0.250 | -0.112 | -0.116 | -0.092 |
| | | $\gamma_2$ | 0.011 | 0.005 | 0.005 | 0.002 |
| RMSE | 4 | $\gamma_1$ | 0.690 | 0.679 | 0.455 | 0.395 |
| | | $\gamma_2$ | 0.021 | 0.018 | 0.010 | 0.013 |
| | 1 | $\gamma_1$ | 0.503 | 0.320 | 0.300 | 0.232 |
| | | $\gamma_2$ | 0.025 | 0.016 | 0.013 | 0.008 |
| Standard deviation | 4 | $\gamma_1$ | 0.509 | 0.484 | 0.380 | 0.339 |
| | | $\gamma_2$ | 0.017 | 0.014 | 0.009 | 0.013 |
| | 1 | $\gamma_1$ | 0.439 | 0.301 | 0.278 | 0.215 |
| | | $\gamma_2$ | 0.023 | 0.016 | 0.012 | 0.008 |
| VNQLL | 4 | | 4.420 | 4.084 | 3.794 | 3.793 |
| | 1 | | 3.853 | 3.550 | 3.346 | 3.301 |
| Runtime (s) | 4 | | 32.132 | 29.856 | 41.815 | 45.052 |
| | 1 | | 31.442 | 30.260 | 40.729 | 44.699 |

Table D.6: Results for no stability constraint on $\gamma$

| Measure | $R$ | Param | Specification | |
|---|---|---|---|---|
| | | | Constrained | Unconstrained |
| Mean bias | 4 | $\gamma_1$ | -0.478 | -0.465 |
| | | $\gamma_2$ | 0.011 | 0.011 |
| | 1 | $\gamma_1$ | -0.112 | -0.148 |
| | | $\gamma_2$ | 0.005 | 0.005 |
| RMSE | 4 | $\gamma_1$ | 0.679 | 0.722 |
| | | $\gamma_2$ | 0.018 | 0.018 |
| | 1 | $\gamma_1$ | 0.320 | 0.351 |
| | | $\gamma_2$ | 0.016 | 0.015 |
| Standard deviation | 4 | $\gamma_1$ | 0.484 | 0.555 |
| | | $\gamma_2$ | 0.014 | 0.014 |
| | 1 | $\gamma_1$ | 0.301 | 0.320 |
| | | $\gamma_2$ | 0.016 | 0.014 |
| VNQLL | 4 | | 4.084 | 4.070 |
| | 1 | | 3.550 | 3.561 |
| Runtime (s) | 4 | | 29.856 | 35.987 |
| | 1 | | 30.260 | 35.777 |

## D.6 No stability constraint on $\gamma$

The baseline estimator imposes

$$|\gamma_k| \leq \bar{\gamma} < 1, \qquad k = 1, 2,$$

through the reparameterization

$$\gamma_k = \bar{\gamma} \tanh(\widetilde{\gamma}_k).$$

This ensures that the simultaneous system remains away from the singular boundary

$$1 - \gamma_1\gamma_2 = 0.$$

In this ablation, we remove the tanh reparameterization and optimize $\gamma_1, \gamma_2 \in \mathbb{R}$ directly. The neural mean and variance networks, regularization selection, validation criterion, and early stopping rule are otherwise unchanged. In the unconstrained specification, optimization runs that produced numerically undefined values of $\log |1 - \gamma_1\gamma_2|$ are handled by update rejection.

This experiment evaluates whether the stability constraint is merely a theoretical convenience or also an important numerical stabilization device. Without the constraint, optimization may approach regions where $|1 - \gamma_1\gamma_2|$ is small. In such regions, the Jacobian term and residual transformation can become unstable, and the fitted model may produce implausibly large feedback coefficients.

Table D.6 shows that removing the stability constraint leaves the estimation of $\gamma_2$ essentially unchanged but modestly worsens the estimation of $\gamma_1$. The unconstrained specification increases the RMSE of $\widehat{\gamma}_1$ from $0.679$ to $0.722$ for $R = 4$ and from $0.320$ to $0.351$ for $R = 1$, mainly through greater empirical dispersion, while VNQLL remains similar. It also increases runtime by approximately $20\%$. These results indicate that the bounded reparameterization provides useful numerical stabilization at little apparent cost to fit.

## D.7 Tied regularization parameters

The baseline SEM-DNN specification uses untied regularization parameters

$$\phi = (\phi_{\nu 1}, \phi_{\nu 2}, \phi_{\alpha 1}, \phi_{\alpha 2})^\top.$$

This allows the two mean networks and the two variance networks to have different penalty strengths. Because the functions $f_1, f_2, g_1, g_2$ may have different effective complexity, the untied specification is used as the default.

In this ablation, we compare the baseline untied specification with a tied specification:

$$\phi_\nu = \phi_{\nu 1} = \phi_{\nu 2}, \qquad \phi_\alpha = \phi_{\alpha 1} = \phi_{\alpha 2}.$$

The tied specification reduces the hyperparameter dimension from four to two:

$$(\phi_{\nu 1}, \phi_{\nu 2}, \phi_{\alpha 1}, \phi_{\alpha 2}) \quad \longrightarrow \quad (\phi_\nu, \phi_\alpha).$$

Table D.7: Results for tied regularization parameters

| Measure | $R$ | Param | Specification | |
|---|---|---|---|---|
| | | | Not tied | Tied |
| Mean bias | 4 | $\gamma_1$ | -0.478 | -0.396 |
| | | $\gamma_2$ | 0.011 | 0.018 |
| | 1 | $\gamma_1$ | -0.112 | -0.146 |
| | | $\gamma_2$ | 0.005 | 0.005 |
| RMSE | 4 | $\gamma_1$ | 0.679 | 0.622 |
| | | $\gamma_2$ | 0.018 | 0.042 |
| | 1 | $\gamma_1$ | 0.320 | 0.392 |
| | | $\gamma_2$ | 0.016 | 0.018 |
| Standard deviation | 4 | $\gamma_1$ | 0.484 | 0.482 |
| | | $\gamma_2$ | 0.014 | 0.038 |
| | 1 | $\gamma_1$ | 0.301 | 0.365 |
| | | $\gamma_2$ | 0.016 | 0.017 |
| VNQLL | 4 | | 4.084 | 4.089 |
| | 1 | | 3.550 | 3.590 |
| Runtime (s) | 4 | | 29.856 | 28.891 |
| | 1 | | 30.260 | 28.868 |

The same quasi-random search range is used:

$$\log_{10} \phi_\nu, \log_{10} \phi_\alpha \in [-8, -1].$$

This ablation examines the tradeoff between flexibility and computational simplicity. Tied regularization may be more stable and cheaper to tune, but it may also under-regularize one network while over-regularizing another.

Table D.7 shows that tying the regularization parameters yields mixed effects across coefficients and signal-to-noise regimes. For $R = 4$, the tied specification reduces the RMSE of $\widehat{\gamma}_1$ from $0.679$ to $0.622$, but increases that of $\widehat{\gamma}_2$ from $0.018$ to $0.042$; for $R = 1$, it raises the RMSE of $\widehat{\gamma}_1$ from $0.320$ to $0.392$, with only a small change for $\widehat{\gamma}_2$. VNQLL is similar, and runtime is modestly lower under tying. Thus, reducing the tuning dimension offers limited computational savings but sacrifices the flexibility to regularize networks with different effective complexities.

## D.8 Different values of $g_{\min}$

The baseline variance specification is

$$g_k(x_i; \alpha_k) = g_{\min} + \psi\{\widetilde{g}_k(x_i; \alpha_k)\}, \qquad k = 1, 2,$$

with $g_{\min} = 0.01$. The lower bound prevents the fitted variance from becoming too close to zero. Since the quasi-log-likelihood contains both $\log g_k(x_i; \alpha_k)$ and $e_{ki}^2 / g_k(x_i; \alpha_k)$, very small fitted variances can dominate the objective and destabilize training.

In this ablation, we compare the baseline value with the alternative values

$$g_{\min} \in \{0.001, 0.01, 0.05\}.$$

All other components of the estimator are kept unchanged. The alternative values of $g_{\min}$ are used only in estimation; the data-generating process continues to use $g_{\min} = 0.01$. Smaller values of $g_{\min}$ allow more flexible variance fitting but may increase the risk of variance collapse. Larger values provide stronger numerical stabilization but may restrict the heteroscedastic signal used for identification and induce model misspecification.

Table D.8 shows that the larger variance floor, $g_{\min} = 0.05$, improves structural estimation in both signal-to-noise regimes, reducing the bias, RMSE, and empirical dispersion of $\widehat{\gamma}_1$ while also modestly improving $\widehat{\gamma}_2$. By contrast, lowering the floor to $g_{\min} = 0.001$ provides no systematic gain and substantially worsens the RMSE of $\widehat{\gamma}_1$ when $R = 1$. Although $g_{\min} = 0.05$ requires more computation than the baseline, it also yields the lowest VNQLL. These results suggest that stronger protection against near-zero fitted variances can improve finite-sample stability without materially suppressing the heteroscedastic identifying signal.

## D.9 Different values of $\bar{\gamma}$

The baseline stability bound is $\bar{\gamma} = 0.99$, so that

$$|\gamma_k| \leq 0.99, \qquad k = 1, 2.$$

Table D.8: Results for different values of $g_{\min}$

| Measure | $R$ | Param | $g_{\min}$ | | |
|---|---|---|---|---|---|
| | | | 0.001 | 0.01 | 0.05 |
| Mean bias | 4 | $\gamma_1$ | -0.449 | -0.478 | -0.308 |
| | | $\gamma_2$ | 0.012 | 0.011 | 0.008 |
| | 1 | $\gamma_1$ | -0.172 | -0.112 | -0.091 |
| | | $\gamma_2$ | 0.006 | 0.005 | 0.003 |
| RMSE | 4 | $\gamma_1$ | 0.663 | 0.679 | 0.520 |
| | | $\gamma_2$ | 0.018 | 0.018 | 0.016 |
| | 1 | $\gamma_1$ | 0.411 | 0.320 | 0.248 |
| | | $\gamma_2$ | 0.019 | 0.016 | 0.011 |
| Standard deviation | 4 | $\gamma_1$ | 0.491 | 0.484 | 0.421 |
| | | $\gamma_2$ | 0.014 | 0.014 | 0.014 |
| | 1 | $\gamma_1$ | 0.375 | 0.301 | 0.232 |
| | | $\gamma_2$ | 0.018 | 0.016 | 0.010 |
| VNQLL | 4 | | 4.111 | 4.084 | 3.931 |
| | 1 | | 3.576 | 3.550 | 3.510 |
| Runtime (s) | 4 | | 36.057 | 29.856 | 36.854 |
| | 1 | | 36.050 | 30.260 | 36.454 |

This bound is loose relative to the true values $\gamma_1 = 0.5$ and $\gamma_2 = 0.4$ but prevents the estimator from approaching the singular region of the simultaneous system.

In this ablation, we examine sensitivity to the choice of $\bar{\gamma}$. We compare

$$\bar{\gamma} \in \{0.9, 0.99, 0.999\}.$$

The structural coefficients are parameterized as

$$\gamma_k = \bar{\gamma} \tanh(\widetilde{\gamma}_k), \qquad k = 1, 2,$$

for each value of $\bar{\gamma}$. A smaller $\bar{\gamma}$ imposes stronger stability and may improve numerical behavior, but it can also introduce boundary effects if the true interaction parameters are close to the imposed limit. A larger $\bar{\gamma}$ is less restrictive but permits training paths closer to instability.

Table D.9 indicates that the results are not highly sensitive to the stability bound over the range considered. For $R = 4$, both $\bar{\gamma} = 0.9$ and $\bar{\gamma} = 0.999$ reduce the RMSE of $\widehat{\gamma}_1$ relative to the baseline $\bar{\gamma} = 0.99$, while differences for $\widehat{\gamma}_2$ remain small. For $R = 1$, the tighter bound $\bar{\gamma} = 0.9$ performs slightly better, whereas $\bar{\gamma} = 0.999$ increases the RMSE and dispersion of $\widehat{\gamma}_1$. Thus, allowing training paths closer to the singular region provides no systematic advantage, and the baseline bound remains a reasonable compromise between flexibility and stability.

# E Empirical illustration details

## E.1 Overview and analysis timeline

This appendix documents the construction of the Dominick's ready-to-eat cereal sample used in Section 5 of the main paper. It describes the available data files, the selection of the analysis period, the linkage of weekly scanner records to product and store information, the construction of the outcomes and predetermined covariates, training-sample preprocessing, sample splitting, model selection, held-out diagnostics, and robustness specifications.

Let $s$, $j$, and $t$ index stores, universal product codes (UPCs), and Dominick's weeks, respectively. The weekly cereal file contains one scanner record for each observed store–UPC–week combination. The observational unit used in the simultaneous-equation model is a store–UPC pair constructed by summarizing a prespecified sequence of weekly records.

For a focal-window starting week $t_0$, define the historical window, exclusion gap, and focal outcome window as

$$\mathcal{W}_H = \{t_0 - 26, \dots, t_0 - 5\}, \qquad \mathcal{W}_G = \{t_0 - 4, \dots, t_0 - 1\}, \qquad \mathcal{W}_0 = \{t_0, t_0 + 1, t_0 + 2, t_0 + 3\},$$

respectively. Thus, $\mathcal{W}_H$ contains 22 weeks, $\mathcal{W}_G$ contains the four weeks immediately preceding the focal window, and $\mathcal{W}_0$ contains the four weeks from which the empirical outcomes are constructed.

Table D.9: Results for different values of $\bar{\gamma}$

| Measure | $R$ | Param | $\bar{\gamma}$ | | |
|---|---|---|---|---|---|
| | | | 0.9 | 0.99 | 0.999 |
| Mean bias | 4 | $\gamma_1$ | -0.379 | -0.478 | -0.403 |
| | | $\gamma_2$ | 0.008 | 0.011 | 0.009 |
| | 1 | $\gamma_1$ | -0.087 | -0.112 | -0.134 |
| | | $\gamma_2$ | 0.002 | 0.005 | 0.005 |
| RMSE | 4 | $\gamma_1$ | 0.629 | 0.679 | 0.619 |
| | | $\gamma_2$ | 0.016 | 0.018 | 0.019 |
| | 1 | $\gamma_1$ | 0.311 | 0.320 | 0.354 |
| | | $\gamma_2$ | 0.014 | 0.016 | 0.017 |
| Standard deviation | 4 | $\gamma_1$ | 0.505 | 0.484 | 0.472 |
| | | $\gamma_2$ | 0.014 | 0.014 | 0.017 |
| | 1 | $\gamma_1$ | 0.300 | 0.301 | 0.330 |
| | | $\gamma_2$ | 0.013 | 0.016 | 0.016 |
| VNQLL | 4 | | 4.048 | 4.084 | 4.048 |
| | 1 | | 3.556 | 3.550 | 3.537 |
| Runtime (s) | 4 | | 36.389 | 29.856 | 36.061 |
| | 1 | | 36.185 | 30.260 | 36.183 |

The historical window is used to construct predetermined scanner-history variables and training-derived preprocessing quantities. The exclusion gap is not used to construct either model inputs or outcomes. The focal window is used only to construct the two outcomes and the associated sample-availability indicators. No weekly price, movement, sale indicator, profit measure, or other scanner variable observed during $\mathcal{W}_G$ or $\mathcal{W}_0$ is included in the model input.

The weekly cereal file reports store, UPC, week, movement, deal quantity, posted deal price, a sale indicator, profit, and a record-validity indicator. It does not provide separate feature-advertising, display, inventory, stock-out, or assortment measures. Accordingly, the baseline historical scanner covariates are restricted to functions of lagged movement, usable lagged prices, the lagged sale indicator, and record availability. Product covariates are constructed from the UPC file, which supplies the product description, package-size field, case quantity, commodity code, and item identifier. Store characteristics are drawn from the demographic file when a valid store linkage is available.

The starting week $t_0$ is selected using the prespecified data-availability protocol described in Section E.3. The protocol uses only week coverage, store and UPC linkage, historical support, and focal-window record-availability conditions defined before model estimation. It does not use the realized values of the constructed outcomes, estimated structural coefficients, validation losses, residual diagnostics, variance-ratio diagnostics, or measures of optimization stability. Once selected, $t_0$ is held fixed for the baseline analysis and for all robustness analyses other than explicitly defined alternative-window specifications.

The data-processing pipeline proceeds in seven stages. First, we verify the uniqueness of the store–UPC–week keys and the validity of the scanner identifiers. Second, we link the weekly scanner records to the UPC file and, when available, to the store demographic file. Third, we assign stores to mutually exclusive training, validation, and diagnostic partitions using the deterministic store-blocked rule in Section E.9. Fourth, we apply the record-validity, product-linkage, package-size parsing, and price-availability rules. Fifth, we identify eligible store–UPC pairs using the historical-support and focal-window availability requirements. Sixth, we construct the outcomes and predetermined covariates. Finally, we fit all data-dependent preprocessing transformations exclusively on the training partition and apply them unchanged to the validation and diagnostic partitions.

Movement-record availability and price availability are treated separately. A weekly scanner row with zero movement may not contain a positive posted price and is not interpreted as providing an observed transaction price. The rules used to construct the focal-period price outcome and to retain or exclude such observations are stated explicitly in Sections E.5 and E.6. In particular, the absence of a usable weekly price is not automatically interpreted as a missing scanner row or as evidence of product unavailability.

Let $z_{sj}$ denote the resulting raw vector of predetermined covariates for store–UPC pair $(s, j)$. The neural-network input is $x_{sj} = \widehat{T}_{\text{pre}}(z_{sj})$, where $\widehat{T}_{\text{pre}}$ is estimated using the training partition only. The same realized input vector $x_{sj}$ is supplied to the two structural-mean networks and the two conditional-variance networks. The four networks may learn different internal representations, but no network-specific, outcome-dependent, validation-sample, or diagnostic-sample preprocessing is performed.

The validation partition is used only for hyperparameter selection, early stopping, and selection of the saved training epoch. The diagnostic partition is accessed only after the sample-construction rules, preprocessing operator, model specification, hyperparameter configuration, and epoch-selection procedure have been fixed. It is then used to compute the descriptive variance-ratio, residual-diagonalization, variance-calibration, and residual-gradient-leverage diagnostics defined in Appendix B.

## E.2 Raw data, identifiers, and weekly scanner records

The analysis uses the ready-to-eat cereal movement file from the Dominick's scanner database. Let $s$, $j$, and $t$ index stores, UPCs, and Dominick's weeks, respectively. The supplied cereal file is already organized at the store–UPC–week level. Each row is therefore treated as one weekly scanner record, and no transaction-level or within-week aggregation is performed.

The weekly cereal file contains the following variables: `STORE`, `UPC`, `WEEK`, `MOVE`, `QTY`, `PRICE`, `SALE`, `PROFIT`, and `OK`. The file additionally contains hexadecimal reproductions of selected numeric fields. These hexadecimal variables are retained for data-verification purposes but are not supplied to the empirical model.

Before constructing the analysis sample, we verify that the store, UPC, and week identifiers are nonmissing and that the key $(s, j, t)$ is unique. In the supplied cereal file, no duplicate store–UPC–week keys are present. Consequently, the original weekly rows are retained without summing movement, averaging prices, or otherwise combining observations.

For a weekly record $(s, j, t)$, define $M_{sjt} = \texttt{MOVE}_{sjt}$ as the reported unit movement. Movement is required to be finite and nonnegative. A value $M_{sjt} = 0$ is treated as an observed zero-movement record rather than as a missing scanner record. The variables `PRICE` and `QTY` describe the posted price offer. Define the implied price per package by

$$p_{sjt} = \frac{\texttt{PRICE}_{sjt}}{\texttt{QTY}_{sjt}}$$

when $\texttt{PRICE}_{sjt} > 0$ and $\texttt{QTY}_{sjt} > 0$. Thus, for example, an offer recorded as a total price of five dollars for two units corresponds to a package price of \$2.50. No division by package size in ounces is made in defining $p_{sjt}$. Price per ounce, when used as a data-quality measure or robustness variable, is constructed separately after linkage to the UPC file.

Define the basic row indicator

$$A^{\text{base}}_{sjt} = \mathbf{1}\left\{ \begin{array}{l} \texttt{STORE}_{sjt}, \texttt{UPC}_{sjt}, \texttt{WEEK}_{sjt} \text{ are nonmissing,} \\ M_{sjt} \text{ is finite and } M_{sjt} \geq 0 \end{array} \right\}.$$

Define $A^{\text{ok}}_{sjt} = \mathbf{1}\{\texttt{OK}_{sjt} = 1\}$, and set $A^{\text{row}}_{sjt} = A^{\text{base}}_{sjt} A^{\text{ok}}_{sjt}$. Thus, $A^{\text{row}}_{sjt} = 1$ indicates that the weekly scanner observation satisfies both the basic identifier-and-movement requirements and the database-supplied validity rule. Records with $\texttt{OK} = 0$ are not treated as valid baseline observations. Their frequency is nevertheless reported, and sensitivity analyses may examine the consequences of alternative validity rules where appropriate.

Usable-price validity is defined in Section E.4, after UPC linkage, package-size parsing, and the training-sample price screen have been introduced. A weekly row may provide valid movement information even when it does not provide a usable package price. In the supplied cereal file, no zero-movement row has a positive usable package price. Such a row therefore contributes to movement coverage and movement-based covariates, but not to price coverage or price-based variables. We do not assign an artificial zero price, carry a future price backward, or interpret the absence of a positive price as evidence that the product was unavailable.

The variable `SALE` is retained as the database-supplied weekly sale indicator. It is used only for historical covariates measured before the exclusion gap. The cereal movement file does not contain separate feature-advertising, display, inventory, stockout, or assortment indicators, and no such variables are constructed for the baseline specification. The codebook documents the values `B`, `C`, and `S` and cautions that blank values do not reliably indicate the absence of promotion. The supplied cereal file additionally contains the codes `G` and `L`. We therefore treat `SALE` as a nominal source-data code, retain an explicit blank/unrecorded category, and impose no common promotional interpretation across nonblank codes.

The variable `PROFIT` is not used as an outcome or contemporaneous covariate in the baseline simultaneous-equation model. Because its exact economic interpretation and scaling differ from the directly observed movement and posted-price fields, it is retained only for data inspection or explicitly designated robustness analyses. It is never used to construct a predetermined covariate from the exclusion-gap or focal-window weeks.

Each UPC is linked to the cereal UPC file using the reported UPC identifier. Let

$$A^{\text{upc}}_{j} = \mathbf{1}\left\{\text{UPC } j \text{ has a unique admissible match in the UPC file}\right\}.$$

The UPC file supplies the commodity code, product description, package-size field, case quantity, and item identifier. It does not supply separate standardized brand, manufacturer, flavor, or cereal-format variables. Any variables extracted from the product description are therefore identified explicitly as text-derived features rather than as native product-file fields.

The package-size field is stored as a character string and is not uniformly numeric. A deterministic parser is applied to convert interpretable ounce-denominated entries to a numeric package size $a_j$. The parser is fixed before model estimation, and the procedure records both the parsed value and a parsing-status category. Define

$$A_j^{\text{size}} = \mathbf{1}\left\{a_j \text{ is positive and unambiguously interpretable}\right\}.$$

The baseline sample uses UPCs satisfying $A_j^{\text{upc}} = 1$ and $A_j^{\text{size}} = 1$.

UPCs with no product-file match, nonunique matches, blank package-size fields, nonpositive sizes, or package-size strings that cannot be interpreted under the prespecified parser are excluded from specifications requiring package size. A robustness specification may retain otherwise matched UPCs with unavailable package size by using a missingness indicator and omitting size-normalized variables.

After the preceding checks, the weekly analysis file contains at most one row for each $(s, j, t)$, together with $M_{sjt}$, $p_{sjt}$, $A_{sjt}^{\text{row}}$, $A_{sjt}^{\text{ok}}$, $\mathsf{SALE}_{sjt}$, and the linked product characteristics. The final movement- and price-validity indicators are defined in Section E.4.

## E.3 Selection of the focal and historical windows

The focal window is selected using a deterministic data-availability rule applied before outcome construction or model estimation. For a candidate focal starting week $t$, define the historical window as $\mathcal{W}_H(t) = \{t-26, \ldots, t-5\}$, the exclusion gap as $\mathcal{W}_G(t) = \{t-4, \ldots, t-1\}$, and the focal window as $\mathcal{W}_0(t) = \{t, t+1, t+2, t+3\}$. Each candidate therefore requires 30 consecutive Dominick's week identifiers.

Let $\mathcal{W}_{\text{obs}}$ denote the set of week identifiers appearing in the cereal movement file. The candidate set is

$$\mathcal{W}_{\text{cand}} = \{t : \{t-26, \ldots, t+3\} \subseteq \mathcal{W}_{\text{obs}}\}.$$

This restriction is necessary because the cereal file does not contain every integer week over its full nominal range. Candidate windows are examined in chronological order.

Window selection uses only identifier availability, static-file linkage, weekly-row availability, and prespecified historical-support conditions. It does not use realized focal-period movement, focal-period price, the focal-period sale indicator, estimated model parameters, validation loss, residual diagnostics, fitted conditional variances, or optimization performance.

For each $t \in \mathcal{W}_{\text{cand}}$, define the preliminary historical record count for store–UPC pair $(s, j)$ by

$$N_{sj}^H(t) = \sum_{\tau \in \mathcal{W}_H(t)} \mathbf{1}\left\{(s, j, \tau) \text{ appears in the cereal file}\right\},$$

and define the preliminary focal record count by

$$N_{sj}^0(t) = \sum_{\tau \in \mathcal{W}_0(t)} \mathbf{1}\left\{(s, j, \tau) \text{ appears in the cereal file}\right\}.$$

A store–UPC pair is preliminarily eligible for candidate $t$ when all of the following conditions hold:

1. UPC $j$ has a unique match in the cereal UPC file;
2. The store identifier $s$ satisfies the store-linkage rule used in the baseline specification;
3. $N_{sj}^H(t)$ is at least $N_{\min}^H$; and
4. $N_{sj}^0(t) = 4$.

The preliminary rule is based on the existence of weekly rows, not on positive movement or positive price. In particular, a focal-week row with zero movement remains a present row for purposes of $N_{sj}^0(t)$. Record validity, usable-price availability, package-size parsing, and the final outcome-specific eligibility rules are imposed only after the focal window has been selected. The baseline preliminary window rule retains only stores having a unique match in the demographic file.

For the baseline analysis, set $N^H_{\min} = 22$, so that a preliminarily eligible pair has a weekly row in every historical week and every focal week. Let

$$N_{\text{pair}}(t) = \sum_{s,j} \mathbf{1}\left\{N^H_{sj}(t) = 22, N^0_{sj}(t) = 4, j \text{ is linked}, s \text{ satisfies the store-linkage rule}\right\}.$$

A candidate starting week is admissible when $N_{\text{pair}}(t) \geq 10000$. The threshold is a minimum database-support requirement and is fixed before model estimation. It is not chosen by comparing empirical outcomes or fitted models across candidate windows.

The baseline focal starting week is the earliest admissible candidate:

$$t_0 = \min\{t \in \mathcal{W}_{\text{cand}} : N_{\text{pair}}(t) \geq 10000\}.$$

The earliest-admissible rule supplies a deterministic tie-breaking criterion and prevents window selection from being influenced by favorable empirical results.

In the supplied cereal data, the observed week identifiers range from 1 through 399, with 367 distinct weeks. The first candidate having all 30 required week identifiers is $t = 27$. Before applying the later record-validity, package-size, and outcome-specific rules, this candidate contains 82 stores, 249 UPCs, and 17,317 store–UPC pairs with rows in all 22 historical weeks and all four focal weeks. Among demographic-file observations with a nonmissing store identifier, the store identifier is unique. All 82 stores in the preliminary selected-window sample have a unique match. Accordingly, $t_0 = 27$ under the baseline preliminary-support rule.

After selecting $t_0$, the analysis periods are fixed as $\mathcal{W}_H = \mathcal{W}_H(t_0)$, $\mathcal{W}_G = \mathcal{W}_G(t_0)$, and $\mathcal{W}_0 = \mathcal{W}_0(t_0)$. The same periods are used in the baseline analysis and in all robustness specifications except explicitly designated alternative-window analyses.

## E.4 Weekly record validity and price screening

This section defines separate indicators for scanner-row validity, movement availability, price availability, product linkage, package-size availability, and price plausibility. These distinctions are maintained because a weekly scanner row may contain valid movement information even when it does not contain a usable positive price.

Define the basic price-validity indicator by

$$A^{p,0}_{sjt} = A^{\text{row}}_{sjt} A^{\text{upc}}_j A^{\text{size}}_j \mathbf{1}\left\{\begin{array}{l} \texttt{PRICE}_{sjt} \text{ is finite and positive,} \\ \texttt{QTY}_{sjt} \text{ is finite and positive,} \\ p_{sjt} = \texttt{PRICE}_{sjt}/\texttt{QTY}_{sjt} \text{ is finite and positive} \end{array}\right\}.$$

Thus, $A^{p,0}_{sjt} = 1$ only when the row supplies valid movement information, the UPC and package size are admissible, and a finite positive package price can be constructed.

For store $s$, UPC $j$, and week $t$, let $M_{sjt}$ denote reported movement and let $p_{sjt}$ denote the package price constructed from `PRICE` and `QTY` as described in Section E.2. Define the basic row-validity indicator by $A^{\text{row}}_{sjt} = 1$ when the store, UPC, and week identifiers are nonmissing, `OK` equals one, and $M_{sjt}$ is finite and nonnegative. Otherwise, set $A^{\text{row}}_{sjt} = 0$. A row satisfying $A^{\text{row}}_{sjt} = 1$ supplies valid weekly movement information regardless of whether movement is positive. In particular, a row with $M_{sjt} = 0$ is retained as an observed zero-movement week. The absence of a positive price on such a row does not invalidate the movement observation. Define the package-price availability indicator by $A^p_{sjt} = 1$ when $A^{\text{row}}_{sjt} = 1$, `PRICE` is finite and positive, `QTY` is finite and positive, and

$$p_{sjt} = \frac{\texttt{PRICE}_{sjt}}{\texttt{QTY}_{sjt}}$$

is finite and positive. Otherwise, set $A^p_{sjt} = 0$.

In the supplied cereal file, positive package prices are observed only on positive-movement rows. Accordingly, a zero-movement row may satisfy $A^{\text{row}}_{sjt} = 1$ while satisfying $A^{p,0}_{sjt} = 0$. No zero price is interpreted as an economic package price, and no price is imputed from future weeks or focal-window observations.

$$\mathcal{P}_{\text{train}} = \left\{p^{\text{oz}}_{sjt} : s \in \mathcal{S}_{\text{train}},\ t \in \mathcal{W}_H,\ A^{p,0}_{sjt} = 1\right\}.$$

Let $A^{\text{upc}}_j = 1$ indicate that UPC $j$ has a unique match in the cereal UPC file. All UPCs appearing in the supplied cereal movement file have unique matches in the supplied UPC file. This linkage condition is nevertheless retained in the replication procedure so that linkage failures are detected if the source files are updated or reconstructed. Let $a_j$ denote package size in ounces obtained from the deterministic parsing procedure described in Section E.2.

Define $A_j^{\text{size}} = 1$ when the package-size string can be unambiguously converted to a finite positive number of ounces. Strings denoting counts, assortments, or otherwise uninterpretable units do not satisfy this condition. When $A_{sjt}^{p,0} = 1$, define

$$p_{sjt}^{\text{oz}} = \frac{p_{sjt}}{a_j}.$$

Price per ounce is used only for detecting implausible price records. The empirical price outcome is constructed from package prices, not from price per ounce.

The price-screening bounds are estimated after stores have been assigned to the training, validation, and diagnostic partitions. Only historical observations from training stores are used. Let

$$\mathcal{P}_{\text{train}} = \left\{p_{sjt}^{\text{oz}} : s \in \mathcal{S}_{\text{train}},\ t \in \mathcal{W}_H,\ A_{sjt}^{p} = 1,\ A_j^{\text{size}} = 1\right\}.$$

Because the UPC file contains only one commodity code for the cereal category, commodity-specific screening bounds cannot be estimated. Instead, the baseline analysis uses a single category-wide lower bound $p_L$ and upper bound $p_U$, defined as the empirical 0.5th and 99.5th percentiles of $\mathcal{P}_{\text{train}}$. Define the price-plausibility indicator by $A_{sjt}^{\text{plaus}} = 1$ when $A_{sjt}^{p} = 1$, $A_j^{\text{size}} = 1$, and $p_L \leq p_{sjt}^{\text{oz}} \leq p_U$. Otherwise, set $A_{sjt}^{\text{plaus}} = 0$. The estimated bounds $p_L$ and $p_U$ are held fixed when applied to the training, validation, and diagnostic partitions. Observations outside the bounds are treated as having an unusable price; their movement values are not automatically discarded. Thus, a price-screening failure affects variables or outcomes requiring a usable price but does not erase otherwise valid movement information. This separation produces two weekly analysis indicators.

For the baseline specification, define the price-validity indicator by

$$V_{sjt}^{P} = A_{sjt}^{p,0}\mathbf{1}\left\{p_L \leq p_{sjt}^{\text{oz}} \leq p_U\right\}.$$

For the robustness specification without price-per-ounce screening, set

$$V_{sjt}^{P} = A_{sjt}^{p,0}.$$

The movement-validity indicator is

$$V_{sjt}^{M} = A_{sjt}^{\text{row}} A_j^{\text{upc}}.$$

The indicator $V_{sjt}^{M}$ is used for historical movement coverage, historical movement covariates, focal movement coverage, and construction of the quantity outcome. The indicator $V_{sjt}^{P}$ is used for historical price coverage, historical price covariates, and construction of any focal price outcome. A weekly row need not satisfy $V_{sjt}^{P} = 1$ to contribute valid movement information.

The price-screening procedure is intended to identify likely coding, unit, or package-size errors. Retained prices are not truncated to the screening bounds. The baseline analysis performs no outcome winsorization through this procedure. Any separate winsorization of constructed outcomes is reported as a robustness specification.

Price plausibility screening is an outcome-availability rule. Although its thresholds are estimated without using validation or diagnostic outcomes, applying those thresholds to focal prices makes final sample membership depend on the realized focal price.

As a robustness check, the analysis is repeated using alternative category-wide screening quantiles, including the 0.1st and 99.9th percentiles and the 1st and 99th percentiles. A second robustness specification omits price-per-ounce screening and instead applies only the basic positivity and finiteness requirements. These alternatives assess whether the empirical conclusions depend materially on the treatment of extreme price records.

## E.5 Construction of the analysis sample

For focal-period movement coverage, define

$$C_{sj}^{0,M} = \prod_{t \in \mathcal{W}_0} V_{sjt}^{M}.$$

Let

$$\mathcal{W}_{sj}^{P} = \left\{t \in \mathcal{W}_0 : V_{sjt}^{P} = 1\right\},$$

and then $N_{sj}^{0,P} = \left|\mathcal{W}_{sj}^{P}\right|$. The baseline common-period sample requires a movement-valid record, $C_{sj}^{0,M} = 1$, and a usable package price in each of the four focal weeks, $N_{sj}^{0,P} = 4$. Conditional on inclusion, both outcomes are therefore constructed from the same prespecified four weeks; the identity and number of weeks entering either outcome do not vary across retained observations. In the supplied data, however, a positive usable price is recorded

only on positive-movement rows. Consequently, the focal-price requirement selects store–UPC pairs with positive recorded movement in every focal week. The baseline estimand is therefore conditional on continuously positive sales and usable price recording during the focal window. This restriction avoids observation-specific temporal aggregation but does not eliminate selection on contemporaneous movement.

For every retained store–UPC pair, define

$$Q_{sj} = \sum_{t \in \mathcal{W}_0} M_{sjt}, \qquad y_{1,sj} = \log Q_{sj},$$

and

$$y_{2,sj} = \frac{1}{4} \sum_{t \in \mathcal{W}_0} \log p_{sjt}.$$

Because the supplied cereal file contains no positive usable package price on a zero-movement row, the condition $N_{sj}^{0,P} = 4$ implies positive recorded movement in every focal week and hence $Q_{sj} > 0$. The baseline empirical population therefore consists of store–UPC pairs with continuously observed prices and positive sales. It does not represent the extensive margin between zero and positive movement.

For each eligible store–UPC pair $(s, j)$, let $\mathcal{H}_{sj}^{M}$ denote the set of historical weeks satisfying $V_{sjt}^{M} = 1$, and let $\mathcal{H}_{sj}^{P}$ denote the set satisfying $V_{sjt}^{P} = 1$. Their cardinalities are denoted by $N_{sj}^{H,M}$ and $N_{sj}^{H,P}$. The historical movement-availability rate is $A_{sj}^{H,M} = N_{sj}^{H,M}/22$, and the historical price-availability rate is $A_{sj}^{H,P} = N_{sj}^{H,P}/22$. Both rates are included because missing positive prices are substantially more common than missing movement records. Let $a_H^M$ and $a_H^P$ denote the historical thresholds for movement and price, respectively. The baseline values are $a_H^M = 0.5$ and $a_H^P = 0.25$. For each value of $K$, define

$$\mathcal{J}_K = \left\{ j : \sum_{s \in \mathcal{S}_{\text{train}}} \mathbf{1}\left\{ A_{sj}^{H,M} \geq a_H^M \right\} \geq K \right\}.$$

Combining the historical-support, UPC-support, and focal-period requirements, define

$$E_{sj} = \mathbf{1}\left\{ A_{sj}^{H,M} \geq a_H^M \right\} \mathbf{1}\left\{ A_{sj}^{H,P} \geq a_H^P \right\} \mathbf{1}\left\{ j \in \mathcal{J}_3 \right\} \mathbf{1}\left\{ C_{sj}^{0,M} = 1 \right\} \mathbf{1}\left\{ N_{sj}^{0,P} = 4 \right\}.$$

The baseline analysis sample is

$$\mathcal{I} = \{(s, j) : E_{sj} = 1\}.$$

For sample-flow reporting, pairs with complete focal movement coverage but failing the focal-price condition are divided into those with $N_{sj}^{0,P} = 0$ and those with $1 \leq N_{sj}^{0,P} \leq 3$.

The final sample is partitioned according to the store assignments defined in Section E.9. Specifically,

$$\mathcal{I}_{\text{train}} = \{(s, j) \in \mathcal{I} : s \in \mathcal{S}_{\text{train}}\}, \qquad \mathcal{I}_{\text{val}} = \{(s, j) \in \mathcal{I} : s \in \mathcal{S}_{\text{val}}\}, \qquad \mathcal{I}_{\text{diag}} = \{(s, j) \in \mathcal{I} : s \in \mathcal{S}_{\text{diag}}\}.$$

No store appears in more than one partition. Because focal-price availability depends on positive focal movement, the baseline common sample excludes store–UPC pairs with $Q_{sj} = 0$. This is a data-availability restriction for the price outcome, not a claim that zero movement is an invalid scanner observation. The replication output therefore reports the number of pairs with complete focal movement coverage that are excluded solely because no focal price is observed.

A robustness analysis relaxes the historical thresholds to $a_H^M = 0.25$ and $a_H^P = 0.10$. A second robustness analysis strengthens them to $a_H^M = 0.75$ and $a_H^P = 0.50$. The same threshold values are used in all three data partitions.

## E.6 Outcome construction

For each eligible store–UPC pair $(s, j)$ in the analysis sample $\mathcal{I}$, the two empirical outcomes are constructed from the four-week focal window:

$$\mathcal{W}_0 = \{t_0, t_0 + 1, t_0 + 2, t_0 + 3\}.$$

Let $M_{sjt}$ denote weekly movement, $p_{sjt}$ denote the package price, and $V_{sjt}^{M}$ and $V_{sjt}^{P}$ denote the movement-validity and price-validity indicators defined in Section E.4.

**Focal movement outcome.** Total focal-period movement is defined as

$$Q_{sj} = \sum_{t \in \mathcal{W}_0} M_{sjt}.$$

The baseline sample requires a movement-valid row in every focal week. Each observation therefore represents the same four-week exposure period. A movement-valid week with zero movement contributes zero to $Q_{sj}$. The baseline sample also requires a usable package price in every focal week, which in the supplied data implies positive movement in every focal week. Consequently, $Q_{sj} > 0$ for every baseline observation. The first structural outcome is therefore

$$y_{1,sj} = \log Q_{sj}.$$

This transformation reduces right skewness of focal movement while avoiding the arbitrary unit offset in $\log(1 + Q_{sj})$. Store–UPC pairs with zero total focal movement are not treated as invalid scanner observations, but they cannot enter the baseline common sample because no focal package price is observed for such pairs.

**Focal price outcome.** By the baseline eligibility rule, $N_{sj}^{0,P} = 4$. The focal price outcome is defined as the arithmetic mean of the available weekly log package prices:

$$y_{2,sj} = \frac{1}{N_{sj}^{0,P}} \sum_{t \in \mathcal{W}_{sj}^P} \log p_{sjt}.$$

Equivalently, $y_{2,sj}$ is the logarithm of the geometric mean package price over the focal weeks in which a valid price is observed. The weekly prices are not weighted by contemporaneous movement. In particular, the construction does not use $M_{sjt}$ or $Q_{sj}$ as price weights. This avoids mechanically introducing the quantity outcome into the price outcome through focal-period expenditure weights. The price outcome is based on package price rather than price per ounce. The UPC identifies a fixed packaged product, and package size is included separately as a predetermined product characteristic. Price per ounce is used only for the price-plausibility screen described in Section E.4.

**Interpretation.** The coefficient $\gamma_1$ measures the contemporaneous structural response of log four-week UPC movement to the average log package price within the same focal window, conditional on the predetermined covariates. The coefficient $\gamma_2$ measures the contemporaneous structural response of average log package price to log four-week UPC movement. Because both outcomes are logarithmic, the coefficients have elasticity-like units.

These coefficients should not, however, be interpreted as conventional demand and supply elasticities. The observational unit is a store–UPC pair, the price outcome is averaged over price-observed focal weeks, and identification relies on conditional heteroscedasticity and conditional shock orthogonality rather than on a standard market-clearing demand-and-supply system. The interpretation is also conditional on the baseline eligibility rule: the analysis concerns store–UPC pairs with positive movement and a usable package price in every focal week and therefore does not directly characterize the extensive-margin transition between zero and positive movement.

**Relaxed focal-price-availability specifications.** Under the robustness specifications that relax the baseline requirement, $N_{sj}^{0,P}$ may range from one to four. To account for this variation, the model input includes predetermined historical price-availability measures, but it does not include $N_{sj}^{0,P}$, because that count is determined during the focal outcome window. The baseline outcome gives equal weight to each price-observed focal week. It does not impute prices for zero-movement weeks, assign zero prices to those weeks, carry historical prices forward, or use prices observed after the focal window.

A robustness analysis relaxes the baseline requirement to observations satisfying $N_{sj}^{0,P} \geq 2$ or $N_{sj}^{0,P} \geq 1$. These specifications assess whether the results depend on constructing the price outcome from only one observed focal price. A further robustness specification uses the logarithm of the arithmetic mean package price,

$$y_{2,sj}^A = \log \left( \frac{1}{N_{sj}^{0,P}} \sum_{t \in \mathcal{W}_{sj}^P} p_{sjt} \right).$$

The baseline mean-of-logs construction and this log-of-mean construction coincide when the weekly package price is constant and differ when prices vary across focal weeks.

**Outlier treatment and reporting.** No additional outcome winsorization is applied in the baseline analysis after the weekly price-validity and plausibility rules in Section E.4 have been imposed. Retained values of $Q_{sj}$ and $p_{sjt}$ are not truncated. As a robustness analysis, outcome winsorization thresholds are estimated separately for $y_{1,sj}$ and $y_{2,sj}$ using the training partition only. The resulting thresholds are then held fixed when applied to the validation and diagnostic partitions.

## E.7 Predetermined covariates

All model inputs are constructed from information observed before the exclusion gap or from time-invariant store and product characteristics. Historical scanner variables are computed only from the 22-week historical window $\mathcal{W}_H$. No movement, price, sale code, profit measure, or other scanner variable observed during the exclusion gap $\mathcal{W}_G$ or focal window $\mathcal{W}_0$ is included in the covariate vector. The raw predetermined covariates are divided into three blocks: store–UPC historical variables, store-level variables, and product-level variables. All data-dependent transformations are subsequently estimated using the training partition only, as described in Section E.8.

**Store–UPC historical variables.** Historical mean weekly movement is calculated over movement-valid weeks as

$$\overline{M}_{sj}^{H} = \frac{1}{N_{sj}^{H,M}} \sum_{t \in \mathcal{H}_{sj}^{M}} M_{sjt}.$$

Historical movement dispersion is measured by the sample standard deviation of $M_{sjt}$ over $\mathcal{H}_{sj}^{M}$. When only one movement-valid historical week is available, this standard deviation is coded as missing and handled by the training-only preprocessing procedure.

To capture persistence and intermittency in movement, the model also includes:

1. the fraction of movement-valid historical weeks with positive movement;
2. the fraction with zero movement;
3. movement in the most recent movement-valid historical week;
4. mean movement during the first half of the historical window;
5. mean movement during the second half of the historical window; and
6. the difference between the second-half and first-half means.

The half-window summaries are included only when each half contains at least the prespecified minimum number of movement-valid weeks. Otherwise, the corresponding value is coded as missing.

The variable SALE is not treated as a binary indicator. For each sale code appearing in the training-store historical data, a separate historical frequency is constructed as the number of movement-valid historical weeks carrying that code divided by $N_{sj}^{H,M}$. Rare codes are pooled according to the training-only categorical pooling rule in Section E.8. A separate frequency records weeks for which SALE is blank or unavailable. No interpretation is imposed that all nonblank sale codes represent an identical form of promotion.

Historical price summaries are computed only over $\mathcal{H}_{sj}^{P}$. They include the mean and standard deviation of log package price, the most recent available historical log package price, and the difference between the second-half and first-half mean log prices when both quantities can be constructed.

For each store-week $(s, t)$, define the reference price as the median log price per ounce among cereal UPCs having a price-valid observation in that store and week. The relative log price for UPC $j$ is the difference between its log price per ounce and this store-week median. The historical mean and standard deviation of relative log price are then computed over the weeks in $\mathcal{H}_{sj}^{P}$ for which the store-week median is available. This relative-price variable measures the UPC's historical price position within the cereal category at the same store and week. It is not constructed for weeks without a usable UPC price, and missing relative prices are not replaced with zero.

**Store-level variables.** Historical cereal-category movement for store $s$ is defined as the average weekly category total over the historical window. Specifically, weekly category movement is first summed over UPCs within each store and week, and the resulting totals are then averaged over the 22 historical weeks. Weeks with no movement-valid cereal records are retained as missing rather than assigned zero.

The model includes the logarithm of one plus historical average weekly category movement. It also includes the fraction of historical weeks for which the store has at least one movement-valid cereal record and the mean number of movement-valid cereal UPCs observed per historical week.

The following time-invariant store variables are taken directly from the demographic file when a unique linkage by the store identifier is available:

1. pricing zone;
2. store cluster;
3. urban indicator;
4. average weekly store volume;
5. latitude and longitude; and
6. a prespecified set of neighborhood demographic characteristics.

The baseline neighborhood demographic set consists of variables representing age composition, ethnic composition, education, vehicle availability, log median income, income dispersion, household size, trading-area size per capita, housing value, unemployment, poverty, and racial composition. In terms of the supplied demographic-file names, the candidate baseline set is

$$\{\texttt{age9}, \texttt{age60}, \texttt{ethnic}, \texttt{educ}, \texttt{nocar}, \texttt{income}, \texttt{incsigma}, \texttt{gini},$$
$$\texttt{hsizeavg}, \texttt{density}, \texttt{hvalmean}, \texttt{unemp}, \texttt{poverty}, \texttt{nwhite}\}.$$

The substantive definitions of these variables follow the Dominick's demographic-file codebook. In particular, the source variable `income` is already defined as the logarithm of median income. It is therefore treated as an already-log-transformed continuous variable and is not subjected to an additional logarithmic transformation. The source variable `density` is defined as trading-area square miles per capita, rather than population per unit of area. We retain the source variable name `density` in the replication files, but interpret it according to this codebook definition. It should therefore not be interpreted as a conventional population-density measure.

The baseline does not automatically include every column in the demographic file. In particular, large groups of variables describing detailed shopping segments, local elasticities, competitive-store identities, or indexed market characteristics are omitted unless their economic interpretation, timing, and units are documented in advance. This restriction reduces the risk that opaque or mechanically related variables enter the model solely because they are present in the file.

Store identifiers, store names, city names, ZIP codes, and competitor identifiers are not treated as continuous covariates. Pricing zone and store cluster are processed as categorical variables. Geographic coordinates are processed as continuous variables.

For a store lacking a unique demographic-file match, the baseline sample excludes that store when store demographics are required. A robustness specification retains such stores by setting the demographic variables to missing and including a store-demographic-linkage indicator. The linkage rule is fixed before model estimation.

**Product-level variables.** The UPC file supplies the UPC identifier, product description, package-size field, case quantity, commodity code, and item identifier. The baseline product-level covariates are:

1. log parsed package size in ounces;
2. log case quantity when positive and interpretable;
3. commodity code as a categorical variable;
4. product-description text features; and
5. indicators describing package-size parsing and missing product attributes.

The item identifier `NITEM` is used only as a linkage or data-checking field and is not included as a continuous numerical covariate. Treating an arbitrary identifier as numerical would impose an unsupported ordering.

The UPC file does not contain separate standardized brand, manufacturer, flavor, or product-format fields. These variables are therefore not included as native categorical covariates in the baseline specification. Information contained in the product description is represented through the training-sample-only text procedure described in Section E.8.

A robustness specification may construct brand or manufacturer groups from the product descriptions, but only if a deterministic parsing dictionary is documented and fixed without reference to outcomes, validation loss, or diagnostic results. Any such text-derived group is labeled as constructed rather than as a source-data field.

Because the cereal UPC file contains a single commodity code for the analyzed category, the commodity code has no within-sample variation and is expected to be removed by the training-sample variance screen. It is retained in the raw covariate definition only so that this fact is documented automatically by the replication output.

**Excluded variables.** The weekly cereal file does not contain distinct feature-advertising, display, inventory, stock-out, or assortment variables. No proxies for these variables are constructed in the baseline specification. The weekly `PROFIT` field is excluded from the baseline covariate vector. Its interpretation may involve contemporaneous retail margins or other quantities that are not clearly predetermined relative to the focal outcomes. It may be examined only in a separately justified robustness analysis using historical observations.

Variables calculated from focal-window movement, focal-window price, focal-window price availability, or focal-window sale codes are excluded. In particular, $N_{sj}^{0,P}$, the number of focal weeks contributing to the price outcome, is not included because it is determined during the focal outcome window. Training, validation, or diagnostic model outputs are never used as empirical covariates.

**Raw covariate vector.** Let $z_{sj}^{H}$ collect the store–UPC historical variables, let $z_{s}^{S}$ collect the store-level historical and time-invariant variables, and let $z_{j}^{J}$ collect the product-level variables. The raw predetermined vector is $z_{sj} = (z_{sj}^{H}, z_{s}^{S}, z_{j}^{J})$.

The neural-network input is $x_{sj} = \widehat{T}_{\text{pre}}(z_{sj})$, where $\widehat{T}_{\text{pre}}$ is estimated exclusively from the training partition. The same fitted preprocessing operator is applied without modification to the validation and diagnostic partitions.

The replication output lists every raw covariate, its source file, its construction window, its transformation, its missingness rate in each partition, and whether it survives the training-sample retention rules. It also reports the numbers of historical observations contributing to each store–UPC summary and the frequencies of the observed `SALE` codes.

## E.8 Preprocessing and neural-network inputs

Let $z_{sj}$ denote the raw vector of predetermined covariates constructed in Section E.7 for store–UPC pair $(s, j)$. The neural-network input is $x_{sj} = \widehat{T}_{\text{pre}}(z_{sj})$, where $\widehat{T}_{\text{pre}}$ is fitted exclusively using observations in the training partition.

Once fitted, the same preprocessing operator is applied without modification to the validation and diagnostic partitions. Validation and diagnostic observations do not affect variable retention, imputation values, transformation parameters, category-pooling rules, text vocabularies, inverse-document-frequency weights, singular-vector loadings, or scaling constants.

The preprocessing operator contains separate transformations for continuous variables, categorical variables, and product-description text. Learned neural-network embeddings are model parameters rather than components of the fitted data-preprocessing operator.

**Prespecification of variable types.** Before preprocessing, every raw covariate is assigned to one of the following classes:

1. nonnegative level variables;
2. proportions and rates;
3. unrestricted continuous variables;
4. geographic coordinates;
5. categorical variables; or
6. product-description text.

This classification is determined from the variable definition and is fixed before model estimation. A covariate is not assigned a transformation by examining its association with either focal outcome. Nonnegative level variables include historical movement summaries, movement dispersion measures, store traffic measures, package size, case quantity, store volume, housing value, trading-area square miles per capita, and household-size measures. Proportions and rates include historical movement availability, historical price availability, positive-movement frequency, zero-movement frequency, sale-code frequencies, poverty rates, unemployment rates, demographic shares, and similar variables whose natural support is bounded. Unrestricted continuous variables include historical log-price summaries, relative log-price summaries, temporal changes, income-dispersion measures, and other variables that can take values on both sides of zero. The demographic variable `income` is also assigned to this class because the source file already records the logarithm of median income. Consequently, `income` is not classified as a nonnegative level variable and is not log-transformed a second time. The variable `density`, whose source definition is trading-area square miles per capita, is classified as a nonnegative level variable.

**Continuous-variable retention and missingness.** Let $z_{sj,r}$ denote raw continuous covariate $r$. A variable is removed when more than 50 percent of its training-partition values are missing. The missingness rate is computed over store–UPC observations in the training sample. For every retained continuous variable, define the missingness indicator

$$d_{sj,r}^{\text{miss}} = \mathbf{1}\left\{z_{sj,r} \text{ is missing}\right\}.$$

This indicator is included for every retained continuous variable, regardless of whether the training partition happens to contain a missing value. This convention ensures that missing values appearing only in a held-out partition remain distinguishable from observed values equal to the imputation constant. Let $\widehat{m}_r$ be the median of the observed training-partition values of $z_{sj,r}$. Missing values in all three partitions are replaced by $\widehat{m}_r$. A variable with no observed training value is removed rather than imputed from a held-out partition. The imputation constant is calculated before transformation. Missingness indicators are not centered, winsorized, or log-transformed.

**Continuous-variable transformations.** For a nonnegative level variable, the transformed value is

$$u_{sj,r} = \log\left(1 + \max\{z_{sj,r}^{\text{imp}}, 0\}\right),$$

where $z_{sj,r}^{\text{imp}}$ denotes the median-imputed value. Under this rule, the source variable `density` is transformed as a nonnegative level variable. No reciprocal transformation is applied to convert it into a population-per-area measure; its substantive interpretation remains trading-area square miles per capita. The use of the maximum with zero is a numerical safeguard; a negative value in a variable defined to be nonnegative is recorded as a data-quality failure before transformation. Variables already expressed as logarithms, including the demographic variable `income`, historical log package prices, and relative log prices, are not transformed again. They proceed directly to the training-sample screening, centering, and scaling steps applicable to unrestricted continuous variables. Proportions and rates are first checked against their theoretically admissible range. Values lying outside that range because of numerical error are clipped to the boundary. No empirical-quantile winsorization is applied to variables naturally bounded between zero and one. For unrestricted continuous variables, training-sample lower and upper screening thresholds are defined by the empirical 0.5th and 99.5th percentiles after imputation. Values below or above these thresholds are clipped to the corresponding threshold. The thresholds are then applied unchanged to the validation and diagnostic partitions. Latitude and longitude are not winsorized by empirical quantiles. They are checked against admissible geographic ranges and retained on their original scales before standardization. Variables such as pricing zone, store cluster, UPC, commodity code, and item identifiers are not processed as continuous quantities merely because their source-file representation is numeric.

**Standardization and removal of constant variables.** For every transformed continuous variable, the training-partition mean $\widehat{\mu}_r$ and sample standard deviation $\widehat{\sigma}_r$ are calculated after imputation and transformation. A transformed variable is removed when $\widehat{\sigma}_r < 10^{-8}$. This rule removes variables with no meaningful variation in the training partition, including the cereal commodity code if it is mistakenly supplied as a continuous variable. Each retained continuous value is standardized as

$$x_{sj,r}^{\text{cont}} = \frac{u_{sj,r} - \widehat{\mu}_r}{\widehat{\sigma}_r}.$$

The training means and standard deviations are used unchanged for the validation and diagnostic partitions. Held-out values are not recentered or rescaled within their own partitions. Binary missingness indicators are retained in their original zero-one form. Other binary variables may either remain unstandardized or be processed as categorical variables according to their substantive definition; the selected convention is fixed in the replication code.

**Categorical variables.** For each categorical variable, levels are identified from the training partition. Leading and trailing whitespace is removed, text case is standardized, and missing values are assigned an explicit `missing` level. A training level represented by fewer than 20 store–UPC observations is pooled into an `other` category. A level observed in the validation or diagnostic partition but absent from the fitted training map is also assigned to `other`. The frequency threshold is based on training observations, except for product-level categorical variables whose values are constant within UPCs. For such variables, the pooling frequency is calculated over distinct training-supported UPCs so that a product carried by many stores does not receive disproportionate influence. The level map and integer encoding are fixed after preprocessing. In the neural networks, each retained categorical variable is represented by a learned embedding. Embedding dimensions are set as

$$d_{\text{emb}} = \min\left\{16, \left\lceil \sqrt{\text{number of training levels}} \right\rceil\right\}.$$

They are not selected using the diagnostic partition. Store identifiers and UPC identifiers are excluded from the baseline neural input. Their inclusion through learned embeddings is considered only in explicitly designated robustness specifications. Arbitrary identifiers such as `NITEM` are not included as ordered numerical variables.

**Product-description text.** Product descriptions are processed at the UPC level. Each distinct UPC contributes at most one document to the fitted text transformation, regardless of the number of stores in which the UPC appears. The text transformation is fitted using the descriptions of UPCs represented in the training sample after application of the training-derived UPC-support rule. Validation and diagnostic store–UPC observations do not contribute additional documents to the fitted vocabulary. Descriptions are normalized using Unicode NFKC normalization, converted to lowercase, and stripped of punctuation. Numerical expressions corresponding to package size, weight, volume, multipack count, or case count are removed using a prespecified regular expression. The removed expressions include common variants of ounces, pounds, grams, counts, and multipack notation. Repeated whitespace is then collapsed. The unmodified source description and cleaned description are retained in the replication output for auditing. A description that is missing or empty after cleaning is assigned a text-missing indicator and represented by a zero text vector.

Term-frequency–inverse-document-frequency features are constructed from word unigrams and adjacent word bigrams. A term is eligible when it occurs in at least five distinct training UPC descriptions and in no more than 95 percent of those descriptions. When more than 2,000 eligible terms remain, the retained terms are selected in decreasing order of document frequency, with ties resolved lexicographically. The vocabulary is therefore deterministic conditional on the training partition.

Inverse-document-frequency weights are calculated using distinct training UPC descriptions as documents. No focal outcomes, validation losses, or diagnostic statistics enter the vocabulary or weight calculations. Let $D_{\text{tfidf}}$ denote the resulting training UPC-by-term matrix. A truncated singular-value decomposition is fitted to $D_{\text{tfidf}}$. The target text dimension is 32, subject to the feasible-rank restriction

$$d_{\text{text}} = \min\left\{32, n_{\text{UPC,train}} - 1, n_{\text{term}} - 1\right\},$$

where $n_{\text{UPC,train}}$ is the number of distinct training UPC documents and $n_{\text{term}}$ is the fitted vocabulary size. If either quantity is too small to support a positive-dimensional decomposition, text components are omitted and only the text-missing indicator is retained.

The fitted vocabulary, inverse-document-frequency weights, and right singular vectors are held fixed when product descriptions in all partitions are transformed. Because the baseline UPC-support set is derived from training stores, most held-out observations are expected to involve UPCs represented in the fitted text transformation. Any previously unseen or empty description is nevertheless processed using the fixed vocabulary without refitting. The resulting singular-value component scores are treated as continuous inputs. Their training-partition means and standard deviations are used for standardization. Text scores with negligible training variation are removed.

**Construction of the neural-network input.** Let $x_{sj}^{\text{cont}}$ collect the transformed continuous variables and their missingness indicators, let $c_{sj}$ collect the pooled categorical indices, and let $x_j^{\text{text}}$ collect the standardized text components and text-missing indicator. The input to each neural network consists of the concatenation of $x_{sj}^{\text{cont}}$, the learned embeddings associated with $c_{sj}$, and $x_j^{\text{text}}$. The same processed input vector is supplied to the two structural-mean networks and the two conditional-variance networks. The networks have separate parameters and may learn different internal representations, but preprocessing, imputation, text transformation, and categorical pooling are common across the four networks. No feature selection is performed separately for the two equations or separately for the mean and variance networks. No transformation is selected according to its predictive relationship with $y_{1,sj}$ or $y_{2,sj}$.

**Fixed numeric representation for diagnostics.** The conditional-moment diagnostics in Appendix B require a fixed numeric feature representation that does not depend on learned neural-network parameters. Let $x_{sj}^{\text{diag}}$ denote this representation. It contains:

1. standardized continuous covariates;
2. continuous-variable missingness indicators;
3. one-hot encodings generated from the training-defined pooled categorical maps;
4. standardized text singular-value components; and
5. the text-missing indicator.

Learned categorical embeddings are excluded because their values depend on the fitted neural model. Store and UPC identifiers are also excluded. To prevent a large categorical variable from dominating the diagnostic principal components solely through its number of levels, each one-hot block is scaled by the square root of the number of retained nonreference levels in that block. This block scaling is fixed from the training representation. Columns of $x_{sj}^{\mathrm{diag}}$ having zero or negligible training variance are removed. The principal-component loadings used in Appendix B.4 are fitted using the resulting training-partition matrix and are then applied unchanged to the diagnostic partition.

**Reproducibility output.** The replication files report:

1. the assigned type and transformation for every raw covariate;
2. training missingness rates and variable-retention decisions;
3. imputation medians, screening thresholds, means, and standard deviations;
4. categorical levels, pooled levels, and unseen-level counts by partition;
5. the distinct UPC descriptions used to fit the text transformation;
6. the cleaned descriptions and text-missing indicators;
7. the fitted vocabulary and inverse-document-frequency weights;
8. the feasible and retained text dimensions;
9. singular values and the fraction of TF–IDF matrix variation represented by the retained components;
10. the dimensions of the neural and diagnostic input vectors; and
11. the columns removed because of missingness or negligible training variation.

All fitted preprocessing objects are saved and reused without refitting for validation, diagnostic, and robustness calculations that employ the same sample split.

## E.9 Sample splitting, model selection, and seedwise reporting

Because observations sharing a store are exposed to common local demand, pricing policies, product availability, and store characteristics, the baseline split is performed at the store level. Every store–UPC observation inherits the partition assignment of its store. No store appears in more than one partition. The empirical objective gives equal weight to retained store–UPC observations. Consequently, stores carrying more retained UPCs and UPCs appearing in more retained stores receive greater aggregate weight. The baseline analysis does not reweight observations to give equal total weight to each store or UPC. Alternative weighting schemes are considered separately in Section E.10.

**Deterministic store-blocked partition.** Let $\mathcal{S}_0$ denote the set of stores appearing in the preliminarily eligible sample for the selected focal window. This set is determined using only the window, identifier-linkage, and preliminary row-availability rules in Section E.3. It is defined before applying training-derived price screens, UPC-support rules, or preprocessing transformations.

For each store $s \in \mathcal{S}_0$, construct a canonical store string by converting the store identifier to its base-10 representation without leading or trailing whitespace. Apply the xxHash64 algorithm to the UTF-8 encoding of the concatenation of a published integer seed, a fixed separator, and the canonical store string. Interpret the resulting hash as an unsigned 64-bit integer.

Stores are sorted in increasing order of their hash values. Hash ties, if any, are resolved by increasing numeric store identifier. Let $n_S = |\mathcal{S}_0|$, $n_{S,\mathrm{train}} = \lfloor 0.60 n_S \rfloor$, and $n_{S,\mathrm{val}} = \lfloor 0.20 n_S \rfloor$. The first $n_{S,\mathrm{train}}$ stores in the ranked list are assigned to the training partition, the next $n_{S,\mathrm{val}}$ stores to the validation partition, and the remaining stores to the diagnostic partition. This ranked-hash procedure produces a deterministic allocation of approximately 60%, 20%, and 20% while avoiding potentially small partitions that can arise from applying fixed hash intervals to a modest number of stores. The hash algorithm, seed, separator, encoding, canonicalization rule, and tie-breaking rule are reported in the replication materials.

The partition is computed once. It is not modified in response to final sample sizes, focal outcomes, estimated coefficients, validation losses, convergence behavior, or diagnostic results. After stores have been assigned, all training-derived sample-construction and preprocessing objects are fitted using training stores only. These objects include:

1. price-per-ounce screening bounds;
2. the UPC-support set;
3. continuous-variable imputation values;
4. clipping thresholds and standardization constants;
5. categorical pooling maps;
6. the product-description vocabulary and inverse-document-frequency weights;
7. text singular-vector loadings; and
8. diagnostic-feature transformations.

The fitted objects are then applied unchanged to observations from validation and diagnostic stores. Eligibility conditions that depend on these objects are evaluated separately in each partition.

**Hyperparameter configurations and random seeds.** Let $\mathcal{H}$ denote the prespecified finite set of hyperparameter configurations and let $\mathcal{S}_{\text{seed}}$ denote the prespecified set of training seeds. A hyperparameter configuration specifies all quantities varied in model selection, including, as applicable, the network architecture, regularization parameters, stabilization parameter, optimizer settings, and training rule. Each seed determines the neural-network initialization, training-data shuffling, and mini-batch ordering. The same ordered set $\mathcal{S}_{\text{seed}}$ is used for every $h \in \mathcal{H}$.

The hyperparameter configurations, seed values, maximum number of epochs, validation-evaluation schedule, early-stopping tolerance and patience, and minimum training length are fixed before estimation and reported in the replication materials. The data partition and all training-derived preprocessing objects are held fixed across $h \in \mathcal{H}$ and $r \in \mathcal{S}_{\text{seed}}$. Thus, comparisons across configurations and seeds reflect only the prespecified changes in the training specification and stochastic optimization path.

**Hyperparameter selection across seeds.** For configuration $h \in \mathcal{H}$, seed $r \in \mathcal{S}_{\text{seed}}$, and validation-evaluation epoch $e$, let $\mathcal{L}^{(0)}_{\text{val}}(h, r, e)$ denote the original unpenalized VNQLL. The superscript $(0)$ distinguishes this criterion from the stabilized or penalized objective used to calculate training updates. Let $\mathcal{E}(h, r)$ denote the set of epochs at which the validation criterion is evaluated before the run terminates. For each run that produces at least one finite validation evaluation and a restorable checkpoint, define

$$\widehat{e}(h, r) = \underset{e \in \mathcal{E}(h,r)}{\arg\min} \mathcal{L}^{(0)}_{\text{val}}(h, r, e),$$

with ties resolved in favor of the earliest epoch, and define the corresponding selected validation loss by

$$\ell_{\text{val}}(h, r) = \mathcal{L}^{(0)}_{\text{val}} \left\{ h, r, \widehat{e}(h, r) \right\}.$$

Early stopping monitors $\mathcal{L}^{(0)}_{\text{val}}(h, r, e)$. Within the prespecified minimum and maximum training lengths, a run stops when the validation loss has failed to decrease by at least the prespecified tolerance for the prespecified number of consecutive validation evaluations. The checkpoint at $\widehat{e}(h, r)$ is retained. The stabilized or penalized training objective affects parameter updates only; early stopping, checkpoint selection, hyperparameter comparison, and stagewise promotion are all based on $\mathcal{L}^{(0)}_{\text{val}}$.

A seed run is classified as successful for hyperparameter comparison when it completes the relevant training stage and produces a finite selected validation loss and a restorable stage-best checkpoint. Let

$$\mathcal{S}_{\text{succ}}(h) = \left\{ r \in \mathcal{S}_{\text{seed}} : \text{the run for configuration } h \text{ and seed } r \text{ is successful} \right\},$$

and let

$$N_{\text{succ}}(h) = \left| \mathcal{S}_{\text{succ}}(h) \right|.$$

A hyperparameter configuration is eligible for comparison only when $N_{\text{succ}}(h) \geq N^{\min}_{\text{succ}}$. In the baseline analysis, $|\mathcal{S}_{\text{seed}}| = 10$ and $N^{\min}_{\text{succ}} = 8$.

For an eligible configuration, failed seed runs are excluded from the validation-loss aggregates rather than being assigned infinite loss. The mean, median, and sample standard deviation of the selected validation losses are therefore calculated over the successful seeds:

$$\bar{\ell}_{\text{val}}(h) = \frac{1}{N_{\text{succ}}(h)} \sum_{r \in \mathcal{S}_{\text{succ}}(h)} \ell_{\text{val}}(h, r),$$

$$\widetilde{\ell}_{\text{val}}(h) = \text{median}_{r \in \mathcal{S}_{\text{succ}}(h)} \, \ell_{\text{val}}(h, r),$$

and

$$s_{\text{val}}(h) = \left[ \frac{1}{N_{\text{succ}}(h) - 1} \sum_{r \in \mathcal{S}_{\text{succ}}(h)} \left\{ \ell_{\text{val}}(h, r) - \overline{\ell}_{\text{val}}(h) \right\}^2 \right]^{1/2},$$

when $N_{\text{succ}}(h) > 1$.

Eligible configurations are ordered first by $\overline{\ell}_{\text{val}}(h)$, then by $\widetilde{\ell}_{\text{val}}(h)$, and then by $s_{\text{val}}(h)$, all in increasing order. Any remaining tie is resolved by the fixed published ordering of the hyperparameter configurations. The top-ranked configuration under this ordering is selected or promoted to the next stage, as applicable.

This rule permits a limited number of seed-specific numerical failures without allowing a configuration supported by too few successful runs to remain eligible. The numbers of successful and failed seeds, the failure reasons, and the seed-specific selected validation losses are retained in the replication output. No quantity calculated from the diagnostic partition is used in hyperparameter selection or stagewise promotion.

**Numerically admissible runs under the selected configuration.** For every seed $r \in \mathcal{S}_{\text{succ}}$, retain the checkpoint from epoch $\widehat{e}(\widehat{h}, r)$. Let $\widehat{\theta}^{(r)}$ denote the saved parameters and let $\widehat{\gamma}^{(r)} = (\widehat{\gamma}_1^{(r)}, \widehat{\gamma}_2^{(r)})^\top$. A selected seed run is numerically admissible when all of the following conditions hold:

1. all saved parameters are finite;
2. the saved checkpoint can be reloaded and evaluated reproducibly;
3. the training and validation losses at the selected checkpoint are finite;
4. all fitted means and conditional variances evaluated on the training and validation partitions are finite;
5. every fitted conditional variance is at least the imposed lower bound $g_{\min}$, up to the stated floating-point tolerance;
6. $|1 - \widehat{\gamma}_1^{(r)} \widehat{\gamma}_2^{(r)}| \geq c_{\text{num}}$; and
7. training reaches at least $e_{\min}$ epochs.

The constants $c_{\text{num}} > 0$, $e_{\min}$, and all numerical tolerances are fixed before estimation. We fix $c_{\text{num}} = 0.01$ and $e_{\min} = 30$. A run is not excluded because its structural coefficients are large, its estimated coefficients differ from those of other seeds, its fitted variance ratio displays weak dispersion, or its held-out residual diagnostics are unfavorable. Such quantities are empirical results rather than numerical-admissibility criteria. Let $\mathcal{S}_{\text{adm}}$ denote the set of admissible seeds. If fewer than the prespecified minimum number of admissible seeds remain, the specification is reported as numerically unsuccessful. It is not replaced by a new seed set chosen after viewing the results.

**Seedwise structural summaries.** For $k \in \{1, 2\}$, the primary estimate reported across admissible seeds is the componentwise median

$$\widehat{\gamma}_k^{\text{med}} = \text{median}_{r \in \mathcal{S}_{\text{adm}}} \, \widehat{\gamma}_k^{(r)}.$$

The tables also report the componentwise 25th and 75th percentiles and the number of admissible seeds. These summaries describe optimization variability under the fixed data split, hyperparameter configuration, and training protocol. They are not confidence intervals. In particular, they do not account for sampling variation, uncertainty induced by sample construction and model selection, or possible violations of the maintained conditional-mean and conditional-covariance restrictions. Seed-specific validation losses and structural estimates are retained together in the replication output so that the relationship between optimization variation and validation fit is visible.

**Representative fitted model.** The componentwise median pair $(\widehat{\gamma}_1^{\text{med}}, \widehat{\gamma}_2^{\text{med}})$ need not correspond to a single fitted neural model. Because the figures, fitted functions, residual calculations, and diagnostics in Appendix B require a coherent set of structural coefficients and neural-network parameters, we select one representative seed run. The representative seed is chosen based on validation loss, not by the estimated structural coefficients or diagnostic results. We order the admissible seeds by their selected validation losses, from smallest to largest, and choose the seed at the median position. When the number of admissible seeds is even, we choose the lower of the two middle seeds; ties are resolved by the smaller published seed value. This rule selects a model with typical

Table E.1: Sample-flow counts for the empirical application

| Description | Training | Validation | Diagnostic | Total |
|---|---|---|---|---|
| Preliminary store–UPC pairs in the selected window | 10411 | 3291 | 3615 | 17317 |
| Pairs retained after product and store linkage | 10268 | 3245 | 3564 | 17077 |
| Pairs retained after historical movement and price coverage requirements | 6850 | 1799 | 2393 | 11042 |
| Pairs retained after the training-derived UPC-support rule | 6849 | 1798 | 2393 | 11040 |
| Pairs retained after complete focal movement coverage | 5229 | 1143 | 1955 | 8327 |
| Pairs retained after the focal-price availability requirement | 4272 | 949 | 1657 | 6878 |
| Final analysis sample | 4272 | 949 | 1657 | 6878 |

validation performance without using any diagnostic-partition quantity. All model-specific figures and held-out diagnostics are then computed from the fitted model associated with $r^\star$.

Selecting the median-validation seed avoids choosing the most favorable validation realization while ensuring that the representative model has typical validation performance under the selected configuration. The rule does not depend on the values of $\widehat{\gamma}_1^{(r)}$, $\widehat{\gamma}_2^{(r)}$, the fitted variance-ratio diagnostics, or any diagnostic-partition quantity.

All model-specific figures and held-out diagnostics are computed using the single fitted model associated with $r^\star$. Tables continue to report the seedwise medians and interquartile ranges of the structural coefficients rather than replacing them with the representative seed's estimates.

**Use of the diagnostic partition.** The diagnostic partition is accessed only after the following objects have been fixed:

1. the focal and historical windows;
2. the store partition;
3. the sample-construction rules;
4. the training-derived preprocessing operator;
5. the neural architecture and hyperparameter grid;
6. the selected hyperparameter configuration;
7. the selected checkpoint for each seed;
8. the numerical-admissibility rule; and
9. the representative seed.

The diagnostic partition is not used for parameter updating, early stopping, hyperparameter selection, checkpoint selection, exclusion of finite but unfavorable runs, or representative-seed selection.

The representative model is then evaluated once on the diagnostic partition to calculate the variance-ratio, residual-diagonalization, variance-calibration, and residual-gradient-leverage diagnostics in Appendix B. These diagnostics are reported as descriptive model checks and do not trigger further retuning or reselection in the baseline analysis.

**Sample-flow and selection reporting.** The first reduction in Table E.1 reflects the baseline product-admissibility rules, including package-size parsing, rather than unmatched store or UPC identifiers alone. The subsequent reductions reflect the historical-coverage requirements, the training-derived UPC-support rule, complete focal-movement coverage, and the focal-price availability requirement, respectively.

## E.10 Robustness checks

The robustness analyses modify one component of the baseline empirical specification at a time. Unless stated otherwise, the focal and historical windows, store partition, weekly-record definitions, neural architecture, stabilization parameter, hyperparameter grid, early-stopping rule, numerical-admissibility rule, and representative-seed procedure remain fixed at their baseline values.

Table E.2: Robustness check: Alternative diagnostic-basis dimensions

| Quantity | Reported value | | |
|---|---|---|---|
| | $q = 4$ | $q = 8$ (Baseline) | $q = 16$ |
| $J_u$ | 570.161 | 621.365 | 707.715 |
| $T_{u,\max}$ | 22.722 | 22.722 | 22.722 |
| $J_{z,1}$ | 55.438 | 59.309 | 83.457 |
| $J_{z,2}$ | 38.782 | 43.694 | 71.852 |
| $T_{z,1,\max}$ | 7.000 | 7.000 | 7.000 |
| $T_{z,2,\max}$ | 5.354 | 5.354 | 5.354 |
| $\widehat{\mu}_{z,1}$ | 1.428 | 1.428 | 1.428 |
| $\widehat{\mu}_{z,2}$ | 1.289 | 1.289 | 1.289 |

Note: Diagnostics are computed for the representative run $(r^{\star})$.

Table E.3: Final number of observations: Alternative UPC-support thresholds

| | $K = 3$ (Baseline) | $K = 5$ | $K = 10$ |
|---|---|---|---|
| Stores | 56 | 56 | 56 |
| UPCs | 164 | 164 | 162 |
| Observations | 6878 | 6878 | 6873 |

For each robustness specification, all data-dependent quantities are re-estimated using only its training partition. These quantities include the price-screening bounds, UPC-support set, imputation and scaling constants, categorical-pooling maps, text transformation, selected hyperparameters, and diagnostic principal-component loadings. The resulting objects are then applied unchanged to the corresponding validation and diagnostic observations. The diagnostic partition is not used to select a preferred robustness specification; the results are interpreted as sensitivity analyses rather than as additional model-selection exercises.

**Alternative diagnostic-basis dimensions.** The baseline conditional-moment diagnostics use $q = 8$ training principal components and their centered squares. The same fitted representative model is evaluated using $q \in \{4, 16\}$. Only the diagnostic basis changes; the sample, outcome model, selected hyperparameters, fitted neural parameters, and representative seed remain fixed. Because the dimension and effective rank of the diagnostic moment vector change with $q$, the resulting values of $J_u$, $J_{z,1}$, and $J_{z,2}$ are interpreted jointly, with particular emphasis on the componentwise maxima, rather than as directly comparable test statistics.

Table E.2 reports the diagnostic summaries for $q \in \{4, 8, 16\}$. The substantive diagnostic conclusions are insensitive to the basis dimension. Although the omnibus quantities $J_u$, $J_{z,1}$, and $J_{z,2}$ increase as additional principal-component directions are included, their levels are not directly comparable across values of $q$. More importantly, the maximum standardized discrepancies are identical across all three bases: $T_{u,\max} = 22.722$, $T_{z,1,\max} = 7.000$, and $T_{z,2,\max} = 5.354$. The unconditional variance-calibration means are also unchanged. Thus, the principal residual-dependence and variance-calibration discrepancies are already present in the smaller basis and are not artifacts of choosing $q = 8$.

**Alternative UPC-support thresholds.** The baseline UPC-support set retains UPCs with adequate historical movement coverage in at least three training stores. To assess sensitivity to products with limited cross-store support, the analysis is repeated with $K \in \{5, 10\}$. The UPC set is reconstructed using training stores only and is then applied unchanged to the validation and diagnostic partitions. All preprocessing transformations and model parameters are re-estimated on the resulting sample. Increasing $K$ strengthens cross-store product support and can reduce the number of UPCs and store–UPC observations. Differences from the baseline may therefore reflect both stronger support and a change in sample composition.

Tables E.3 and E.4 show that increasing the support threshold from $K = 3$ to $K = 5$ leaves the sample and all reported results unchanged. Raising the threshold to $K = 10$ removes only two UPCs and five store–UPC observations. The median estimates change modestly, from $(0.347, -0.238)$ to $(0.294, -0.227)$, while their seedwise interquartile ranges remain similar. The representative-run estimates and residual-gradient-leverage measures are also stable. The fitted curvature becomes somewhat weaker and more anisotropic at $K = 10$, but the broader residual-dependence and variance-calibration patterns remain. Thus, the empirical conclusions do not appear to be driven by UPCs with limited cross-store support.

Table E.4: Robustness check: Alternative UPC-support thresholds

| Quantity | Reported value | | |
|---|---|---|---|
| | $K = 3$ (Baseline) | $K = 5$ | $K = 10$ |
| Median $\widehat{\gamma}_1$ | 0.347 | 0.347 | 0.294 |
| Interquartile range of $\widehat{\gamma}_1$ | [-0.123, 0.573] | [-0.123, 0.573] | [-0.086, 0.583] |
| Median $\widehat{\gamma}_2$ | -0.238 | -0.238 | -0.227 |
| Interquartile range of $\widehat{\gamma}_2$ | [-0.560, 0.586] | [-0.560, 0.586] | [-0.545, 0.583] |
| Representative-run $\widehat{\gamma}_1^{(r^\star)}$ | 0.719 | 0.719 | 0.719 |
| Representative-run $\widehat{\gamma}_2^{(r^\star)}$ | -0.651 | -0.651 | -0.656 |
| $1 - \widehat{\gamma}_1^{(r^\star)}\widehat{\gamma}_2^{(r^\star)}$ | 1.468 | 1.468 | 1.472 |
| $\mathcal{L}_{\text{val}}^{(0)}(\widehat{\theta}^{(r^\star)})$ | -2.578 | -2.578 | -2.482 |
| $\lambda_{\min}(\widehat{H}_\rho)$ | 0.056 | 0.056 | 0.044 |
| $\lambda_{\max}(\widehat{H}_\rho)/\lambda_{\min}(\widehat{H}_\rho)$ | 48.611 | 48.611 | 69.942 |
| $\det(\widehat{H}_\rho)$ | 0.155 | 0.155 | 0.135 |
| $J_u$ | 621.365 | 621.365 | 668.101 |
| $T_{u,\max}$ | 22.722 | 22.722 | 24.299 |
| $J_{z,1}$ | 59.309 | 59.309 | 48.373 |
| $J_{z,2}$ | 43.694 | 43.694 | 55.844 |
| $T_{z,1,\max}$ | 7.000 | 7.000 | 5.974 |
| $T_{z,2,\max}$ | 5.354 | 5.354 | 6.844 |
| $\widehat{\mu}_{z,1}$ | 1.428 | 1.428 | 1.326 |
| $\widehat{\mu}_{z,2}$ | 1.289 | 1.289 | 1.388 |
| $\text{IQR}_{\widetilde{\varpi}_1^{(\beta)}}$ | 0.529 | 0.529 | 0.583 |
| $\text{IQR}_{\widetilde{\varpi}_2^{(\beta)}}$ | 0.373 | 0.373 | 0.302 |
| $Q_{0.95}(\widetilde{\varpi}_{1i}^{(\beta)})$ | 1.793 | 1.793 | 1.829 |
| $Q_{0.95}(\widetilde{\varpi}_{2i}^{(\beta)})$ | 1.320 | 1.320 | 1.275 |
| $\max(\widetilde{\varpi}_{1i}^{(\beta)})$ | 2.469 | 2.469 | 2.451 |
| $\max(\widetilde{\varpi}_{2i}^{(\beta)})$ | 1.921 | 1.921 | 1.887 |
| $\text{ESS}_{1,d}^{(\beta)}$ | 0.861 | 0.861 | 0.848 |
| $\text{ESS}_{2,d}^{(\beta)}$ | 0.933 | 0.933 | 0.945 |
| Admissible runs | 10 | 10 | 10 |
| Failed runs | 0 | 0 | 0 |

Note: Diagnostics are computed for the representative run $(r^\star)$.

Table E.5: Final number of observations: Alternative historical-coverage requirements

| | Relaxed | Baseline | Strict |
|---|---|---|---|
| Stores | 56 | 56 | 56 |
| UPCs | 166 | 164 | 163 |
| Observations | 7016 | 6878 | 6848 |

**Alternative historical-coverage requirements.** The baseline requires $A_{sj}^{H,M} \geq 0.50$ and $A_{sj}^{H,P} \geq 0.25$. Two alternative specifications are considered. The relaxed-coverage specification uses $A_{sj}^{H,M} \geq 0.25$ and $A_{sj}^{H,P} \geq 0.10$. The strict-coverage specification uses $A_{sj}^{H,M} \geq 0.75$ and $A_{sj}^{H,P} \geq 0.50$. The UPC-support set is reconstructed under each pair of thresholds. Historical summaries are always calculated from the available movement-valid or price-valid weeks and are accompanied by their coverage measures and missingness indicators. This analysis assesses whether the estimates depend on retaining store–UPC pairs with sparse historical information.

Tables E.5 and E.6 show that changing the historical-coverage requirements alters the sample only modestly and leaves the structural estimates largely unchanged. The median estimates and their seedwise interquartile ranges are similar under the relaxed, baseline, and strict specifications, as are the representative-run estimates. The identification diagnostics are more sensitive: both alternatives produce weaker and more anisotropic fitted curvature than the baseline, particularly under relaxed coverage. Residual-dependence discrepancies also increase under the relaxed specification, whereas variance calibration and residual-gradient leverage do not deteriorate systematically. Thus, the coefficient estimates are robust to the historical-coverage thresholds, although the baseline requirements provide the strongest fitted heteroscedastic curvature.

**Alternative focal-price availability requirements.** The baseline common sample requires complete focal movement coverage and a valid positive package price in every focal week. Thus, the baseline requirement is $N_{sj}^{0,P} = 4$. To assess sensitivity to this complete-price-coverage requirement, we consider two relaxed samples: $N_{sj}^{0,P} \geq 2$ and $N_{sj}^{0,P} \geq 1$. These specifications retain store–UPC pairs for which a usable package price is observed in only a subset of the four focal weeks. Relaxing the requirement can enlarge the analysis sample but makes the construction of the price outcome depend more strongly on which focal weeks contain positive movement and usable scanner-price information. Results are therefore reported together with the corresponding changes in the numbers of observations, stores, and UPCs and in the distributions of the outcomes. The baseline store assignments are retained. Training-derived price-screening, UPC-support, and preprocessing objects are re-estimated after applying each focal-price availability requirement.

For either relaxed focal-price availability requirement, let $\mathcal{W}_{sj}^{P} \subseteq \mathcal{W}_0$ denote the set of focal weeks having a valid package price, so that $N_{sj}^{0,P} = |\mathcal{W}_{sj}^{P}|$. The baseline mean-of-logs construction is extended to the relaxed samples as

$$y_{2,sj} = \frac{1}{N_{sj}^{0,P}} \sum_{t \in \mathcal{W}_{sj}^{P}} \log p_{sjt}.$$

When $N_{sj}^{0,P} = 4$, this definition coincides with the baseline price outcome. When $N_{sj}^{0,P} < 4$, the outcome averages only over focal weeks with a valid package price. These relaxed specifications are interpreted cautiously because both $N_{sj}^{0,P}$ and the identity of the weeks in $\mathcal{W}_{sj}^{P}$ may depend on focal-period movement realizations. The focal price count is not included among the predetermined covariates because it is realized during the focal outcome window.

Tables E.7 and E.8 show that relaxing the focal-price requirement expands the sample only modestly: the number of observations increases from 6878 to 7269 when $N_{sj}^{0,P} \geq 2$ and to 7301 when $N_{sj}^{0,P} \geq 1$, with no additional stores and only four additional UPCs. The median structural estimates and their seedwise interquartile ranges remain nearly unchanged. The diagnostic results are mixed. The relaxed specifications have higher VNQLL and progressively poorer variance calibration; under $N_{sj}^{0,P} \geq 1$, residual-gradient leverage also becomes more concentrated. Although this least restrictive specification displays stronger fitted curvature and somewhat smaller residual-dependence discrepancies, its price outcome depends more heavily on the focal weeks in which a usable price is observed. The baseline requirement is therefore retained because it provides a common four-week price construction with only a modest loss of sample size.

**Alternative focal-price aggregation.** As a robustness specification, the price outcome is constructed as the logarithm of the arithmetic mean of the observed focal-week package prices:

Table E.6: Robustness check: Alternative historical-coverage requirements

| Quantity | Reported value | | |
|---|---|---|---|
| | Relaxed | Baseline | Strict |
| Median $\widehat{\gamma}_1$ | 0.346 | 0.347 | 0.356 |
| Interquartile range of $\widehat{\gamma}_1$ | [-0.123, 0.586] | [-0.123, 0.573] | [-0.095, 0.575] |
| Median $\widehat{\gamma}_2$ | -0.231 | -0.238 | -0.238 |
| Interquartile range of $\widehat{\gamma}_2$ | [-0.558, 0.577] | [-0.560, 0.586] | [-0.554, 0.599] |
| Representative-run $\widehat{\gamma}_1^{(r^\star)}$ | 0.711 | 0.719 | 0.714 |
| Representative-run $\widehat{\gamma}_2^{(r^\star)}$ | -0.663 | -0.651 | -0.663 |
| $1-\widehat{\gamma}_1^{(r^\star)}\widehat{\gamma}_2^{(r^\star)}$ | 1.472 | 1.468 | 1.473 |
| $\mathcal{L}_{\text{val}}^{(0)}(\widehat{\theta}^{(r^\star)})$ | -2.530 | -2.578 | -2.524 |
| $\lambda_{\min}(\widehat{H}_\rho)$ | 0.024 | 0.056 | 0.031 |
| $\lambda_{\max}(\widehat{H}_\rho)/\lambda_{\min}(\widehat{H}_\rho)$ | 116.515 | 48.611 | 83.801 |
| $\det(\widehat{H}_\rho)$ | 0.066 | 0.155 | 0.079 |
| $J_u$ | 718.341 | 621.365 | 630.324 |
| $T_{u,\max}$ | 24.233 | 22.722 | 22.878 |
| $J_{z,1}$ | 36.734 | 59.309 | 54.992 |
| $J_{z,2}$ | 36.591 | 43.694 | 39.658 |
| $T_{z,1,\max}$ | 4.098 | 7.000 | 6.689 |
| $T_{z,2,\max}$ | 3.630 | 5.354 | 4.957 |
| $\widehat{\mu}_{z,1}$ | 1.153 | 1.428 | 1.403 |
| $\widehat{\mu}_{z,2}$ | 1.091 | 1.289 | 1.170 |
| $\text{IQR}_{\widetilde{\varpi}_1^{(\beta)}}$ | 0.407 | 0.529 | 0.442 |
| $\text{IQR}_{\widetilde{\varpi}_2^{(\beta)}}$ | 0.417 | 0.373 | 0.442 |
| $Q_{0.95}(\widetilde{\varpi}_{1i}^{(\beta)})$ | 1.709 | 1.793 | 1.718 |
| $Q_{0.95}(\widetilde{\varpi}_{2i}^{(\beta)})$ | 1.371 | 1.320 | 1.359 |
| $\max(\widetilde{\varpi}_{1i}^{(\beta)})$ | 2.472 | 2.469 | 2.295 |
| $\max(\widetilde{\varpi}_{2i}^{(\beta)})$ | 1.621 | 1.921 | 1.780 |
| $\text{ESS}_{1,d}^{(\beta)}$ | 0.889 | 0.861 | 0.882 |
| $\text{ESS}_{2,d}^{(\beta)}$ | 0.931 | 0.933 | 0.926 |
| Admissible runs | 10 | 10 | 10 |
| Failed runs | 0 | 0 | 0 |

Note: Diagnostics are computed for the representative run $(r^\star)$.

Table E.7: Final number of observations: Alternative focal-price availability requirements

| | $N_{sj}^{0,P}=4$ (Baseline) | $N_{sj}^{0,P}\geq 2$ | $N_{sj}^{0,P}\geq 1$ |
|---|---|---|---|
| Stores | 56 | 56 | 56 |
| UPCs | 164 | 168 | 168 |
| Observations | 6878 | 7269 | 7301 |

Table E.8: Robustness check: Alternative focal-price availability requirements

| Quantity | Reported value | | |
|---|---|---|---|
| | $N_{sj}^{0,P} = 4$ (Baseline) | $N_{sj}^{0,P} \geq 2$ | $N_{sj}^{0,P} \geq 1$ |
| Median $\widehat{\gamma}_1$ | 0.347 | 0.348 | 0.344 |
| Interquartile range of $\widehat{\gamma}_1$ | [-0.123, 0.573] | [-0.113, 0.579] | [-0.103, 0.578] |
| Median $\widehat{\gamma}_2$ | -0.238 | -0.237 | -0.225 |
| Interquartile range of $\widehat{\gamma}_2$ | [-0.560, 0.586] | [-0.560, 0.591] | [-0.547, 0.584] |
| Representative-run $\widehat{\gamma}_1^{(r^\star)}$ | 0.719 | 0.720 | 0.736 |
| Representative-run $\widehat{\gamma}_2^{(r^\star)}$ | -0.651 | -0.651 | -0.627 |
| $1 - \widehat{\gamma}_1^{(r^\star)}\widehat{\gamma}_2^{(r^\star)}$ | 1.468 | 1.469 | 1.462 |
| $\mathcal{L}_{\text{val}}^{(0)}(\widehat{\theta}^{(r^\star)})$ | -2.578 | -2.467 | -2.436 |
| $\lambda_{\min}(\widehat{H}_\rho)$ | 0.056 | 0.044 | 0.284 |
| $\lambda_{\max}(\widehat{H}_\rho)/\lambda_{\min}(\widehat{H}_\rho)$ | 48.611 | 64.679 | 13.809 |
| $\det(\widehat{H}_\rho)$ | 0.155 | 0.122 | 1.112 |
| $J_u$ | 621.365 | 533.336 | 530.522 |
| $T_{u,\max}$ | 22.722 | 19.707 | 20.803 |
| $J_{z,1}$ | 59.309 | 77.919 | 109.422 |
| $J_{z,2}$ | 43.694 | 60.824 | 88.734 |
| $T_{z,1,\max}$ | 7.000 | 7.547 | 9.206 |
| $T_{z,2,\max}$ | 5.354 | 5.670 | 7.722 |
| $\widehat{\mu}_{z,1}$ | 1.428 | 1.574 | 1.789 |
| $\widehat{\mu}_{z,2}$ | 1.289 | 1.379 | 1.572 |
| $\text{IQR}_{\widetilde{\varpi}_1^{(\beta)}}$ | 0.529 | 0.563 | 0.669 |
| $\text{IQR}_{\widetilde{\varpi}_2^{(\beta)}}$ | 0.373 | 0.355 | 0.277 |
| $Q_{0.95}(\widetilde{\varpi}_{1i}^{(\beta)})$ | 1.793 | 1.878 | 1.881 |
| $Q_{0.95}(\widetilde{\varpi}_{2i}^{(\beta)})$ | 1.320 | 1.321 | 1.334 |
| $\max(\widetilde{\varpi}_{1i}^{(\beta)})$ | 2.469 | 2.279 | 4.687 |
| $\max(\widetilde{\varpi}_{2i}^{(\beta)})$ | 1.921 | 2.080 | 3.544 |
| $\text{ESS}_{1,d}^{(\beta)}$ | 0.861 | 0.845 | 0.795 |
| $\text{ESS}_{2,d}^{(\beta)}$ | 0.933 | 0.932 | 0.921 |
| Admissible runs | 10 | 10 | 10 |
| Failed runs | 0 | 0 | 0 |

Note: Diagnostics are computed for the representative run $(r^\star)$.

Table E.9: Robustness check: Alternative focal-price aggregation

| Quantity | Reported value | |
|---|---|---|
| | Baseline | Alternative |
| Median $\widehat{\gamma}_1$ | 0.347 | 0.347 |
| Interquartile range of $\widehat{\gamma}_1$ | [-0.123, 0.573] | [-0.086, 0.574] |
| Median $\widehat{\gamma}_2$ | -0.238 | -0.233 |
| Interquartile range of $\widehat{\gamma}_2$ | [-0.560, 0.586] | [-0.552, 0.598] |
| Representative-run $\widehat{\gamma}_1^{(r^\star)}$ | 0.719 | 0.719 |
| Representative-run $\widehat{\gamma}_2^{(r^\star)}$ | -0.651 | -0.653 |
| $1-\widehat{\gamma}_1^{(r^\star)}\widehat{\gamma}_2^{(r^\star)}$ | 1.468 | 1.469 |
| $\mathcal{L}^{(0)}_{\text{val}}(\widehat{\theta}^{(r^\star)})$ | -2.578 | -2.576 |
| $\lambda_{\min}(\widehat{H}_\rho)$ | 0.056 | 0.052 |
| $\lambda_{\max}(\widehat{H}_\rho)/\lambda_{\min}(\widehat{H}_\rho)$ | 48.611 | 52.634 |
| $\det(\widehat{H}_\rho)$ | 0.155 | 0.143 |
| $J_u$ | 621.365 | 621.890 |
| $T_{u,\max}$ | 22.722 | 22.601 |
| $J_{z,1}$ | 59.309 | 59.404 |
| $J_{z,2}$ | 43.694 | 42.441 |
| $T_{z,1,\max}$ | 7.000 | 7.006 |
| $T_{z,2,\max}$ | 5.354 | 5.126 |
| $\widehat{\mu}_{z,1}$ | 1.428 | 1.430 |
| $\widehat{\mu}_{z,2}$ | 1.289 | 1.277 |
| $\text{IQR}_{\widetilde{\varpi}_1^{(\beta)}}$ | 0.529 | 0.516 |
| $\text{IQR}_{\widetilde{\varpi}_2^{(\beta)}}$ | 0.373 | 0.377 |
| $Q_{0.95}(\widetilde{\varpi}_{1i}^{(\beta)})$ | 1.793 | 1.791 |
| $Q_{0.95}(\widetilde{\varpi}_{2i}^{(\beta)})$ | 1.320 | 1.327 |
| $\max(\widetilde{\varpi}_{1i}^{(\beta)})$ | 2.469 | 2.298 |
| $\max(\widetilde{\varpi}_{2i}^{(\beta)})$ | 1.921 | 1.922 |
| $\text{ESS}_{1,d}^{(\beta)}$ | 0.861 | 0.863 |
| $\text{ESS}_{2,d}^{(\beta)}$ | 0.933 | 0.932 |
| Admissible runs | 10 | 10 |
| Failed runs | 0 | 0 |

Note: Diagnostics are computed for the representative run $(r^\star)$.

$$y_{2,sj} = \log\left(\frac{1}{N_{sj}^{0,P}} \sum_{t\in\mathcal{W}_{sj}^P} p_{sjt}\right).$$

The mean-of-logs specification is retained because it is the logarithm of the geometric mean of the weekly package prices. It coincides with the logarithm of the arithmetic mean only when the weekly prices are constant; otherwise, the two differ by Jensen's inequality.

Table E.9 shows that replacing the mean of weekly log prices with the log of their arithmetic mean has negligible effects. The median estimates change from $(0.347, -0.238)$ to $(0.347, -0.233)$, with similar seedwise interquartile ranges and virtually identical representative-run estimates and VNQLLs. The fitted-curvature, residual-moment, variance-calibration, and residual-gradient-leverage diagnostics are also nearly unchanged. Thus, the empirical conclusions are insensitive to the focal-price aggregation rule; the baseline mean-log specification is retained for its direct interpretation as the log of the geometric-mean package price.

**Recorded-sale-code restriction.** The supplied file contains the nonblank `SALE` codes `B`, `C`, `S`, `G`, and `L`, whose meanings are not uniformly documented. The baseline specification uses historical frequencies of these codes as predetermined covariates but imposes no focal-period restriction. As a limited sensitivity analysis, we define a no-recorded-sale-code sample that retains only store–UPC pairs for which `SALE` is blank in all four focal weeks. This restriction excludes observations with a recorded `BonusBuy`, `Coupon`, simple price reduction, or other nonblank

Table E.10: Final number of observations: Recorded-sale-code restriction

| | Baseline | No-recorded-sale-code sample |
|---|---|---|
| Stores | 56 | 56 |
| UPCs | 164 | 147 |
| Observations | 6878 | 6122 |

code during the focal window, but it does not establish the absence of unrecorded promotions, displays, inventory constraints, wholesale-cost changes, or other contemporaneous retail decisions.

Tables E.10 and E.11 show that the restriction removes 756 observations and 17 UPCs but no stores. The median structural estimates, their seedwise interquartile ranges, and the representative-run estimates remain similar to the baseline, indicating that the estimates are not materially driven by focal observations carrying a nonblank recorded sale code. The restricted sample has a higher VNQLL and weaker, more anisotropic fitted curvature. Residual-dependence discrepancies decline but remain substantial, while the variance-calibration diagnostics are mixed: the unconditional calibration means move closer to one and the maximum standardized discrepancies fall, but the omnibus statistics increase. Residual-gradient weights become somewhat less concentrated. Because a blank SALE field does not establish the absence of promotion, the baseline specification is retained as the broader and more readily interpretable sample.

**Alternative price-screening rules.** The baseline weekly price screen uses the 0.5th and 99.5th percentiles of historical price per ounce among training-store observations. Two alternative category-wide screens are considered: the 0.1st and 99.9th percentiles (Relaxed) and the 1st and 99th percentiles (Strict). A further specification omits empirical-quantile screening and retains every weekly price satisfying the basic positivity, finiteness, product-linkage, and package-size requirements (None). For each specification, the screening bounds are re-estimated from historical observations in the training partition and applied unchanged to all partitions. The sample, outcomes, and preprocessing transformations are then reconstructed. This exercise assesses whether the results are sensitive to the treatment of extreme weekly prices and the resulting focal outcomes.

Tables E.12 and E.13 show that the structural estimates are insensitive to the weekly price-screening rule. Relative to the baseline sample of 6878 observations and 164 UPCs, eliminating empirical-quantile screening adds only 140 observations and three UPCs, whereas the strict screen removes 30 observations and one UPC. Across all four specifications, the median estimates, their seedwise interquartile ranges, and the representative-run estimates remain similar. The no-screen specification also produces fitted-curvature diagnostics close to the baseline. By contrast, the relaxed and strict quantile screens yield weaker and more anisotropic curvature, while none of the alternatives improves the validation criterion. The residual-dependence and variance-calibration diagnostics vary across screening rules without a uniform ordering: the relaxed and strict screens improve several variance-calibration measures but weaken curvature, whereas omitting screening modestly worsens variance calibration. Residual-gradient leverage remains broadly stable. Thus, the substantive coefficient estimates are not driven by the treatment of extreme weekly prices, and the baseline screen is retained because it removes implausible price observations while preserving comparatively favorable validation and identification diagnostics.

**Outcome winsorization.** The baseline outcomes are not winsorized after weekly price screening. To assess sensitivity to extreme focal-period outcomes, both outcomes are winsorized using training-derived cutoffs (Winsorized-both). For $k \in \{1, 2\}$, let $\widehat{q}_{k,0.005}$ and $\widehat{q}_{k,0.995}$ denote the empirical 0.5th and 99.5th percentiles of $y_{k,sj}$ in the training partition. Define

$$y^W_{k,sj} = \min \{\max (y_{k,sj}, \widehat{q}_{k,0.005}), \widehat{q}_{k,0.995}\}.$$

The training-derived cutoffs are applied unchanged to the validation and diagnostic partitions. Sample eligibility, predetermined covariates, weekly price-screening bounds, and store assignments remain unchanged. An additional specification winsorizes only the movement outcome (Winsorized-move), and another winsorizes only the price outcome (Winsorized-price). These comparisons help identify which outcome drives any sensitivity.

Table E.14 shows that winsorizing the focal-period outcomes at training-derived 0.5th and 99.5th percentiles has negligible effects on the results. Regardless of whether both outcomes, only the movement outcome, or only the price outcome are winsorized, the median estimates, their seedwise interquartile ranges, and the representative-run estimates remain virtually unchanged. VNQLLs are also identical up to rounding. Winsorization produces slightly weaker and more anisotropic fitted curvature, while the residual-dependence and variance-calibration diagnostics exhibit only modest, non-systematic changes. The outcome-specific specifications likewise provide no evidence that extreme observations in either outcome drive the baseline findings. Although winsorization modestly reduces some maximum residual-gradient-leverage measures, the corresponding interquartile ranges and effective sample

Table E.11: Robustness check: Recorded-sale-code restriction

| Quantity | Reported value | |
|---|---|---|
| | Baseline | No-recorded-sale-code sample |
| Median $\widehat{\gamma}_1$ | 0.347 | 0.361 |
| Interquartile range of $\widehat{\gamma}_1$ | [-0.123, 0.573] | [-0.083, 0.567] |
| Median $\widehat{\gamma}_2$ | -0.238 | -0.245 |
| Interquartile range of $\widehat{\gamma}_2$ | [-0.560, 0.586] | [-0.589, 0.609] |
| Representative-run $\widehat{\gamma}_1^{(r^\star)}$ | 0.719 | 0.703 |
| Representative-run $\widehat{\gamma}_2^{(r^\star)}$ | -0.651 | -0.669 |
| $1 - \widehat{\gamma}_1^{(r^\star)}\widehat{\gamma}_2^{(r^\star)}$ | 1.468 | 1.470 |
| $\mathcal{L}_{\mathrm{val}}^{(0)}(\widehat{\theta}^{(r^\star)})$ | -2.578 | -2.490 |
| $\lambda_{\min}(\widehat{H}_\rho)$ | 0.056 | 0.032 |
| $\lambda_{\max}(\widehat{H}_\rho)/\lambda_{\min}(\widehat{H}_\rho)$ | 48.611 | 81.877 |
| $\det(\widehat{H}_\rho)$ | 0.155 | 0.086 |
| $J_u$ | 621.365 | 566.054 |
| $T_{u,\max}$ | 22.722 | 20.994 |
| $J_{z,1}$ | 59.309 | 72.876 |
| $J_{z,2}$ | 43.694 | 59.199 |
| $T_{z,1,\max}$ | 7.000 | 3.044 |
| $T_{z,2,\max}$ | 5.354 | 2.970 |
| $\widehat{\mu}_{z,1}$ | 1.428 | 1.000 |
| $\widehat{\mu}_{z,2}$ | 1.289 | 0.975 |
| $\mathrm{IQR}_{\widetilde{\varpi}_1^{(\beta)}}$ | 0.529 | 0.416 |
| $\mathrm{IQR}_{\widetilde{\varpi}_2^{(\beta)}}$ | 0.373 | 0.303 |
| $Q_{0.95}(\widetilde{\varpi}_{1i}^{(\beta)})$ | 1.793 | 1.670 |
| $Q_{0.95}(\widetilde{\varpi}_{2i}^{(\beta)})$ | 1.320 | 1.347 |
| $\max(\widetilde{\varpi}_{1i}^{(\beta)})$ | 2.469 | 2.352 |
| $\max(\widetilde{\varpi}_{2i}^{(\beta)})$ | 1.921 | 1.554 |
| $\mathrm{ESS}_{1,d}^{(\beta)}$ | 0.861 | 0.899 |
| $\mathrm{ESS}_{2,d}^{(\beta)}$ | 0.933 | 0.957 |
| Admissible runs | 10 | 10 |
| Failed runs | 0 | 0 |

Note: Diagnostics are computed for the representative run $(r^\star)$.

Table E.12: Final number of observations: Alternative price-screening rules

| | None | Relaxed | Baseline | Strict |
|---|---|---|---|---|
| Stores | 56 | 56 | 56 | 56 |
| UPCs | 167 | 166 | 164 | 163 |
| Observations | 7018 | 7016 | 6878 | 6848 |

Table E.13: Robustness check: Alternative price-screening rules

| Quantity | Reported value | | | |
|---|---|---|---|---|
| | None | Relaxed | Baseline | Strict |
| Median $\widehat{\gamma}_1$ | 0.348 | 0.346 | 0.347 | 0.356 |
| Interquartile range of $\widehat{\gamma}_1$ | [-0.084, 0.574] | [-0.123, 0.586] | [-0.123, 0.573] | [-0.095, 0.575] |
| Median $\widehat{\gamma}_2$ | -0.223 | -0.231 | -0.238 | -0.238 |
| Interquartile range of $\widehat{\gamma}_2$ | [-0.566, 0.598] | [-0.558, 0.577] | [-0.560, 0.586] | [-0.554, 0.599] |
| Representative-run $\widehat{\gamma}_1^{(r^\star)}$ | 0.719 | 0.711 | 0.719 | 0.714 |
| Representative-run $\widehat{\gamma}_2^{(r^\star)}$ | -0.649 | -0.663 | -0.651 | -0.663 |
| $1-\widehat{\gamma}_1^{(r^\star)}\widehat{\gamma}_2^{(r^\star)}$ | 1.466 | 1.472 | 1.468 | 1.473 |
| $\mathcal{L}_{\text{val}}^{(0)}(\widehat{\theta}^{(r^\star)})$ | -2.535 | -2.530 | -2.578 | -2.524 |
| $\lambda_{\min}(\widehat{H}_\rho)$ | 0.057 | 0.024 | 0.056 | 0.031 |
| $\lambda_{\max}(\widehat{H}_\rho)/\lambda_{\min}(\widehat{H}_\rho)$ | 48.742 | 116.515 | 48.611 | 83.801 |
| $\det(\widehat{H}_\rho)$ | 0.158 | 0.066 | 0.155 | 0.079 |
| $J_u$ | 620.959 | 718.341 | 621.365 | 630.324 |
| $T_{u,\max}$ | 22.802 | 24.233 | 22.722 | 22.878 |
| $J_{z,1}$ | 69.206 | 36.734 | 59.309 | 54.992 |
| $J_{z,2}$ | 50.626 | 36.591 | 43.694 | 39.658 |
| $T_{z,1,\max}$ | 7.524 | 4.098 | 7.000 | 6.689 |
| $T_{z,2,\max}$ | 6.349 | 3.630 | 5.354 | 4.957 |
| $\widehat{\mu}_{z,1}$ | 1.459 | 1.153 | 1.428 | 1.403 |
| $\widehat{\mu}_{z,2}$ | 1.373 | 1.091 | 1.289 | 1.170 |
| $\text{IQR}_{\widetilde{\varpi}_1^{(\beta)}}$ | 0.516 | 0.407 | 0.529 | 0.442 |
| $\text{IQR}_{\widetilde{\varpi}_2^{(\beta)}}$ | 0.356 | 0.417 | 0.373 | 0.442 |
| $Q_{0.95}(\widetilde{\varpi}_{1i}^{(\beta)})$ | 1.724 | 1.709 | 1.793 | 1.718 |
| $Q_{0.95}(\widetilde{\varpi}_{2i}^{(\beta)})$ | 1.312 | 1.371 | 1.320 | 1.359 |
| $\max(\widetilde{\varpi}_{1i}^{(\beta)})$ | 2.191 | 2.472 | 2.469 | 2.295 |
| $\max(\widetilde{\varpi}_{2i}^{(\beta)})$ | 1.683 | 1.621 | 1.921 | 1.780 |
| $\text{ESS}_{1,d}^{(\beta)}$ | 0.869 | 0.889 | 0.861 | 0.882 |
| $\text{ESS}_{2,d}^{(\beta)}$ | 0.935 | 0.931 | 0.933 | 0.926 |
| Admissible runs | 10 | 10 | 10 | 10 |
| Failed runs | 0 | 0 | 0 | 0 |

Note: Diagnostics are computed for the representative run $(r^\star)$.

Table E.14: Robustness check: Outcome winsorization

| Quantity | Reported value | | | |
|---|---|---|---|---|
| | Baseline | Winsorized-both | Winsorized-move | Winsorized-price |
| Median $\widehat{\gamma}_1$ | 0.347 | 0.347 | 0.346 | 0.345 |
| Interquartile range of $\widehat{\gamma}_1$ | [-0.123, 0.573] | [-0.120, 0.571] | [-0.122, 0.574] | [-0.123, 0.573] |
| Median $\widehat{\gamma}_2$ | -0.238 | -0.234 | -0.238 | -0.238 |
| Interquartile range of $\widehat{\gamma}_2$ | [-0.560, 0.586] | [-0.572, 0.585] | [-0.572, 0.586] | [-0.559, 0.585] |
| Representative-run $\widehat{\gamma}_1^{(r^\star)}$ | 0.719 | 0.719 | 0.719 | 0.719 |
| Representative-run $\widehat{\gamma}_2^{(r^\star)}$ | -0.651 | -0.652 | -0.653 | -0.649 |
| $1-\widehat{\gamma}_1^{(r^\star)}\widehat{\gamma}_2^{(r^\star)}$ | 1.468 | 1.469 | 1.470 | 1.467 |
| $\mathcal{L}_{\text{val}}^{(0)}(\widehat{\theta}^{(r^\star)})$ | -2.578 | -2.578 | -2.579 | -2.579 |
| $\lambda_{\min}(\widehat{H}_\rho)$ | 0.056 | 0.053 | 0.050 | 0.054 |
| $\lambda_{\max}(\widehat{H}_\rho)/\lambda_{\min}(\widehat{H}_\rho)$ | 48.611 | 50.914 | 55.037 | 50.702 |
| $\det(\widehat{H}_\rho)$ | 0.155 | 0.141 | 0.136 | 0.147 |
| $J_u$ | 621.365 | 630.336 | 637.943 | 627.399 |
| $T_{u,\max}$ | 22.722 | 23.023 | 23.091 | 22.951 |
| $J_{z,1}$ | 59.309 | 64.774 | 65.660 | 59.506 |
| $J_{z,2}$ | 43.694 | 40.814 | 47.359 | 43.115 |
| $T_{z,1,\max}$ | 7.000 | 7.394 | 7.397 | 7.014 |
| $T_{z,2,\max}$ | 5.354 | 4.868 | 5.702 | 5.061 |
| $\widehat{\mu}_{z,1}$ | 1.428 | 1.458 | 1.457 | 1.428 |
| $\widehat{\mu}_{z,2}$ | 1.289 | 1.256 | 1.310 | 1.267 |
| $\text{IQR}_{\widetilde{\varpi}_1^{(\beta)}}$ | 0.529 | 0.509 | 0.523 | 0.521 |
| $\text{IQR}_{\widetilde{\varpi}_2^{(\beta)}}$ | 0.373 | 0.391 | 0.354 | 0.381 |
| $Q_{0.95}(\widetilde{\varpi}_{1i}^{(\beta)})$ | 1.793 | 1.763 | 1.781 | 1.794 |
| $Q_{0.95}(\widetilde{\varpi}_{2i}^{(\beta)})$ | 1.320 | 1.324 | 1.306 | 1.323 |
| $\max(\widetilde{\varpi}_{1i}^{(\beta)})$ | 2.469 | 2.228 | 2.252 | 2.339 |
| $\max(\widetilde{\varpi}_{2i}^{(\beta)})$ | 1.921 | 1.878 | 1.802 | 1.976 |
| $\text{ESS}_{1,d}^{(\beta)}$ | 0.861 | 0.869 | 0.864 | 0.863 |
| $\text{ESS}_{2,d}^{(\beta)}$ | 0.933 | 0.932 | 0.938 | 0.932 |
| Admissible runs | 10 | 10 | 10 | 10 |
| Failed runs | 0 | 0 | 0 | 0 |

Note: Diagnostics are computed for the representative run $(r^\star)$.

sizes remain similar. The baseline unwinsorized specification is therefore retained because the preceding weekly price screen already removes implausible prices and because additional outcome truncation does not materially affect the empirical conclusions.

**Alternative observation weighting.** The baseline empirical criterion gives equal weight to each retained store–UPC observation. Stores carrying many UPCs and UPCs sold in many stores consequently receive greater aggregate weight. Two alternative weighting schemes are considered. Under equal-store weighting, observation $(s, j)$ receives a raw weight inversely proportional to the number of retained UPCs in store $s$:

$$w_{sj}^S = \frac{1}{n_s},$$

where $n_s$ is the number of retained observations belonging to store $s$ in the relevant partition. Under equal-UPC weighting, observation $(s, j)$ receives

$$w_{sj}^J = \frac{1}{n_j},$$

where $n_j$ is the number of retained stores carrying UPC $j$ in the relevant partition. Within each partition, raw weights are normalized to have mean one. The same weighting convention is used in the training objective and validation criterion for a given specification. Diagnostic quantities retain their original equal-observation definitions, so that this comparison changes the estimation weights rather than the diagnostic definitions. These analyses assess whether the results are dominated by large stores or widely distributed products.

Table E.15: Robustness check: Alternative observation weighting

| Quantity | Reported value | | |
|---|---|---|---|
| | Baseline | Equal-store weighting | Equal-UPC weighting |
| Median $\widehat{\gamma}_1$ | 0.347 | 0.345 | 0.354 |
| Interquartile range of $\widehat{\gamma}_1$ | [-0.123, 0.573] | [-0.092, 0.568] | [-0.104, 0.559] |
| Median $\widehat{\gamma}_2$ | -0.238 | -0.240 | -0.240 |
| Interquartile range of $\widehat{\gamma}_2$ | [-0.560, 0.586] | [-0.556, 0.598] | [-0.582, 0.594] |
| Representative-run $\widehat{\gamma}_1^{(r^\star)}$ | 0.719 | 0.718 | 0.720 |
| Representative-run $\widehat{\gamma}_2^{(r^\star)}$ | -0.651 | -0.656 | -0.663 |
| $1-\widehat{\gamma}_1^{(r^\star)}\widehat{\gamma}_2^{(r^\star)}$ | 1.468 | 1.471 | 1.477 |
| $\mathcal{L}_{\text{val}}^{(0)}(\widehat{\theta}^{(r^\star)})$ | -2.578 | -2.550 | -2.422 |
| $\lambda_{\min}(\widehat{H}_\rho)$ | 0.056 | 0.051 | 0.036 |
| $\lambda_{\max}(\widehat{H}_\rho)/\lambda_{\min}(\widehat{H}_\rho)$ | 48.611 | 54.189 | 76.071 |
| $\det(\widehat{H}_\rho)$ | 0.155 | 0.140 | 0.101 |
| $J_u$ | 621.365 | 612.760 | 608.445 |
| $T_{u,\max}$ | 22.722 | 22.444 | 22.806 |
| $J_{z,1}$ | 59.309 | 59.353 | 49.442 |
| $J_{z,2}$ | 43.694 | 43.144 | 38.930 |
| $T_{z,1,\max}$ | 7.000 | 7.042 | 5.729 |
| $T_{z,2,\max}$ | 5.354 | 5.436 | 4.690 |
| $\widehat{\mu}_{z,1}$ | 1.428 | 1.427 | 1.318 |
| $\widehat{\mu}_{z,2}$ | 1.289 | 1.303 | 1.242 |
| $\text{IQR}_{\widetilde{\varpi}_1^{(\beta)}}$ | 0.529 | 0.526 | 0.512 |
| $\text{IQR}_{\widetilde{\varpi}_2^{(\beta)}}$ | 0.373 | 0.386 | 0.376 |
| $Q_{0.95}(\widetilde{\varpi}_{1i}^{(\beta)})$ | 1.793 | 1.785 | 1.848 |
| $Q_{0.95}(\widetilde{\varpi}_{2i}^{(\beta)})$ | 1.320 | 1.325 | 1.332 |
| $\max(\widetilde{\varpi}_{1i}^{(\beta)})$ | 2.469 | 2.294 | 2.440 |
| $\max(\widetilde{\varpi}_{2i}^{(\beta)})$ | 1.921 | 1.839 | 1.857 |
| $\text{ESS}_{1,d}^{(\beta)}$ | 0.861 | 0.864 | 0.853 |
| $\text{ESS}_{2,d}^{(\beta)}$ | 0.933 | 0.932 | 0.935 |
| Admissible runs | 10 | 10 | 10 |
| Failed runs | 0 | 0 | 0 |

Note: Diagnostics are computed for the representative run $(r^\star)$.

Table E.15 shows that the structural estimates are insensitive to the observation-weighting scheme. Equal-store and equal-UPC weighting produce median estimates, seedwise interquartile ranges, and representative-run estimates that are nearly identical to their baseline counterparts. Equal-store weighting also leaves the residual-dependence, variance-calibration, and residual-gradient-leverage diagnostics essentially unchanged. Equal-UPC weighting reduces several variance-calibration discrepancies, but it yields a substantially higher VNQLL and weaker, more anisotropic fitted curvature. The residual-dependence and leverage diagnostics otherwise remain broadly comparable across specifications. Because the diagnostic quantities retain their equal-observation definitions, these results isolate the effect of changing the estimation weights. Overall, the baseline findings do not appear to be driven by stores carrying unusually many UPCs or by UPCs distributed across unusually many stores; the equal-observation specification is retained because it provides the most favorable validation performance and strongest fitted curvature.

**Pair-random benchmark.** As a deliberately less stringent predictive benchmark, store–UPC pairs are assigned directly to training, validation, and diagnostic partitions using the deterministic ranked-hash procedure and the same approximate 60%, 20%, and 20% proportions. Under this split, a store or UPC may appear in more than one partition. The resulting validation and diagnostic observations may therefore share store-level or product-level information with the training data. All training-derived objects are re-estimated using the pair-random training partition. The pair-random results are not treated as an alternative primary design because cross-partition dependence can make predictive performance and residual diagnostics appear more favorable. A UPC-blocked specification

Table E.16: Robustness check: Pair-random benchmark

| Quantity | Reported value | |
|---|---|---|
| | Baseline | Pair-random |
| Median $\widehat{\gamma}_1$ | 0.347 | 0.329 |
| Interquartile range of $\widehat{\gamma}_1$ | [-0.123, 0.573] | [-0.119, 0.572] |
| Median $\widehat{\gamma}_2$ | -0.238 | -0.234 |
| Interquartile range of $\widehat{\gamma}_2$ | [-0.560, 0.586] | [-0.565, 0.588] |
| Representative-run $\widehat{\gamma}_1^{(r^\star)}$ | 0.719 | 0.724 |
| Representative-run $\widehat{\gamma}_2^{(r^\star)}$ | -0.651 | -0.658 |
| $1 - \widehat{\gamma}_1^{(r^\star)}\widehat{\gamma}_2^{(r^\star)}$ | 1.468 | 1.476 |
| $\mathcal{L}_{\text{val}}^{(0)}(\widehat{\theta}^{(r^\star)})$ | -2.578 | -2.607 |
| $\lambda_{\min}(\widehat{H}_\rho)$ | 0.056 | 0.020 |
| $\lambda_{\max}(\widehat{H}_\rho)/\lambda_{\min}(\widehat{H}_\rho)$ | 48.611 | 133.724 |
| $\det(\widehat{H}_\rho)$ | 0.155 | 0.056 |
| $J_u$ | 621.365 | 650.387 |
| $T_{u,\max}$ | 22.722 | 22.069 |
| $J_{z,1}$ | 59.309 | 22.175 |
| $J_{z,2}$ | 43.694 | 20.041 |
| $T_{z,1,\max}$ | 7.000 | 3.805 |
| $T_{z,2,\max}$ | 5.354 | 2.493 |
| $\widehat{\mu}_{z,1}$ | 1.428 | 1.202 |
| $\widehat{\mu}_{z,2}$ | 1.289 | 1.123 |
| $\text{IQR}_{\widetilde{\varpi}_1^{(\beta)}}$ | 0.529 | 0.437 |
| $\text{IQR}_{\widetilde{\varpi}_2^{(\beta)}}$ | 0.373 | 0.378 |
| $Q_{0.95}(\widetilde{\varpi}_{1i}^{(\beta)})$ | 1.793 | 1.615 |
| $Q_{0.95}(\widetilde{\varpi}_{2i}^{(\beta)})$ | 1.320 | 1.288 |
| $\max(\widetilde{\varpi}_{1i}^{(\beta)})$ | 2.469 | 2.227 |
| $\max(\widetilde{\varpi}_{2i}^{(\beta)})$ | 1.921 | 1.635 |
| $\text{ESS}_{1,d}^{(\beta)}$ | 0.861 | 0.896 |
| $\text{ESS}_{2,d}^{(\beta)}$ | 0.933 | 0.940 |
| Admissible runs | 10 | 10 |
| Failed runs | 0 | 0 |

Note: Diagnostics are computed for the representative run $(r^\star)$.

is not used. It is incompatible with the baseline requirement that every retained UPC satisfy a support condition defined from training stores, because UPCs assigned exclusively to held-out partitions would necessarily fail that requirement.

Table E.16 shows that replacing the store-blocked partition with a pair-random split leaves the structural estimates essentially unchanged. The median estimates, their seedwise interquartile ranges, and the representative-run estimates are all close to the baseline values. The pair-random split yields a slightly lower VNQLL and substantially smaller variance-calibration discrepancies, with $\widehat{\mu}_{z,1}$ and $\widehat{\mu}_{z,2}$ moving closer to one; residual-gradient leverage is also somewhat less concentrated. These improvements should be interpreted cautiously, however, because stores and UPCs may occur in both the training and held-out partitions, allowing shared store- and product-level information to make out-of-sample diagnostics appear more favorable. Moreover, the fitted curvature is markedly weaker and more anisotropic under the pair-random split, as reflected in the lower minimum eigenvalue, larger condition number, and smaller determinant. Thus, the benchmark supports the stability of the coefficient estimates across partitioning schemes, but it does not constitute a preferable validation design. The store-blocked baseline is retained because it more stringently evaluates generalization to previously unseen stores.

**Alternative focal windows.** To assess sensitivity to the selected market period, the complete empirical pipeline is repeated for prespecified alternative starting weeks in the admissible candidate set $\mathcal{W}_{\text{cand}}$. The alternative windows are selected by a calendar rule fixed before model estimation. The robustness analysis uses the first two nonoverlapping admissible focal windows after the baseline window. A window is nonoverlapping when its focal

Table E.17: Final number of observations: Alternative focal windows

| | Baseline | Alternative 1 | Alternative 2 |
|---|---|---|---|
| Stores | 56 | 46 | 48 |
| UPCs | 164 | 161 | 163 |
| Observations | 6878 | 5618 | 5982 |

period does not share a week with a previously selected focal period. For each alternative starting week $t$, the historical, gap, and focal windows are redefined as

$$\mathcal{W}_H(t) = \{t-26,\ldots,t-5\}, \qquad \mathcal{W}_G(t) = \{t-4,\ldots,t-1\}, \qquad \mathcal{W}_0(t) = \{t,t+1,t+2,t+3\}.$$

The complete construction and estimation pipeline is repeated, including preliminary eligibility, store assignment, training-derived price screening, UPC-support construction, outcome construction, preprocessing, hyperparameter selection, seedwise estimation, and diagnostic evaluation. The alternative windows are not chosen according to estimated coefficients, validation losses, or diagnostic results. The estimates from these windows are not treated as independent because historical periods, stores, and UPCs may overlap even when the focal windows do not.

Tables E.17 and E.18 show that the estimated bidirectional effects are stable across the two subsequent nonoverlapping focal windows. Although the alternative windows yield smaller samples, with $5618$ and $5982$ observations compared with $6878$ in the baseline, the median estimates, seedwise interquartile ranges, and representative-run estimates remain nearly unchanged. The stability quantity $1-\widehat{\gamma}_1\widehat{\gamma}_2$ is likewise essentially constant. Both alternative windows yield higher VNQLLs and somewhat weaker, more anisotropic fitted curvature. Their residual-dependence discrepancies are smaller, whereas the variance-calibration diagnostics generally deteriorate, particularly for the second outcome. Residual-gradient leverage remains broadly comparable and, by several measures, becomes slightly less concentrated. Because the windows are selected by a prespecified calendar rule rather than by their estimates or diagnostics, these results indicate that the main coefficient findings are not specific to the baseline market period. The diagnostic differences should not be interpreted as independent temporal replications, however, because the specifications may share historical weeks, stores, and UPCs.

Table E.18: Robustness check: Alternative focal windows

| Quantity | Reported value | | |
|---|---|---|---|
| | Baseline | Alternative 1 | Alternative 2 |
| Median $\widehat{\gamma}_1$ | 0.347 | 0.356 | 0.354 |
| Interquartile range of $\widehat{\gamma}_1$ | [-0.123, 0.573] | [-0.087, 0.563] | [-0.109, 0.570] |
| Median $\widehat{\gamma}_2$ | -0.238 | -0.248 | -0.230 |
| Interquartile range of $\widehat{\gamma}_2$ | [-0.560, 0.586] | [-0.590, 0.604] | [-0.571, 0.589] |
| Representative-run $\widehat{\gamma}_1^{(r^\star)}$ | 0.719 | 0.707 | 0.720 |
| Representative-run $\widehat{\gamma}_2^{(r^\star)}$ | -0.651 | -0.667 | -0.650 |
| $1-\widehat{\gamma}_1^{(r^\star)}\widehat{\gamma}_2^{(r^\star)}$ | 1.468 | 1.471 | 1.468 |
| $\mathcal{L}_{\mathrm{val}}^{(0)}(\widehat{\theta}^{(r^\star)})$ | -2.578 | -2.271 | -2.471 |
| $\lambda_{\min}(\widehat{H}_\rho)$ | 0.056 | 0.045 | 0.044 |
| $\lambda_{\max}(\widehat{H}_\rho)/\lambda_{\min}(\widehat{H}_\rho)$ | 48.611 | 61.335 | 65.368 |
| $\det(\widehat{H}_\rho)$ | 0.155 | 0.126 | 0.125 |
| $J_u$ | 621.365 | 481.505 | 499.956 |
| $T_{u,\max}$ | 22.722 | 17.916 | 19.351 |
| $J_{z,1}$ | 59.309 | 75.715 | 69.364 |
| $J_{z,2}$ | 43.694 | 78.229 | 72.067 |
| $T_{z,1,\max}$ | 7.000 | 6.846 | 7.132 |
| $T_{z,2,\max}$ | 5.354 | 7.532 | 7.530 |
| $\widehat{\mu}_{z,1}$ | 1.428 | 1.385 | 1.631 |
| $\widehat{\mu}_{z,2}$ | 1.289 | 1.373 | 1.552 |
| $\mathrm{IQR}_{\widetilde{\varpi}_1^{(\beta)}}$ | 0.529 | 0.539 | 0.461 |
| $\mathrm{IQR}_{\widetilde{\varpi}_2^{(\beta)}}$ | 0.373 | 0.415 | 0.316 |
| $Q_{0.95}(\widetilde{\varpi}_{1i}^{(\beta)})$ | 1.793 | 1.737 | 1.752 |
| $Q_{0.95}(\widetilde{\varpi}_{2i}^{(\beta)})$ | 1.320 | 1.343 | 1.297 |
| $\max(\widetilde{\varpi}_{1i}^{(\beta)})$ | 2.469 | 2.269 | 2.201 |
| $\max(\widetilde{\varpi}_{2i}^{(\beta)})$ | 1.921 | 1.722 | 1.512 |
| $\mathrm{ESS}_{1,d}^{(\beta)}$ | 0.861 | 0.863 | 0.872 |
| $\mathrm{ESS}_{2,d}^{(\beta)}$ | 0.933 | 0.933 | 0.937 |
| Admissible runs | 10 | 10 | 10 |
| Failed runs | 0 | 0 | 0 |

Note: Diagnostics are computed for the representative run $(r^\star)$.


\end{document}